\documentclass{article}

\PassOptionsToPackage{square,numbers,sort&compress}{natbib}
\usepackage[final]{neurips_2018}

\usepackage[utf8]{inputenc} %
\usepackage[T1]{fontenc}    %
\usepackage{hyperref}       %
\usepackage{url}            %
\usepackage{booktabs}       %
\usepackage{amsfonts}       %
\usepackage{nicefrac}       %
\usepackage{microtype}      %

\usepackage{standalone}
\usepackage{color}
\usepackage{subcaption}
\usepackage{amsmath}

\usepackage{tabularx}

\graphicspath{{figures/}}

\definecolor{human.100}{RGB}{165, 30, 55}
\definecolor{vgg.100}{RGB}{0, 105, 170}
\definecolor{googlenet.100}{RGB}{80, 170, 200}
\definecolor{resnet.100}{RGB}{65, 90, 140}
\definecolor{resnet50.100}{RGB}{43, 140, 190}
\definecolor{all.distortions.net.100}{RGB}{123, 204, 196}
\definecolor{specialised.net.100}{RGB}{186, 228, 188}

\title{Generalisation in humans and deep neural networks}

\author{
  Robert Geirhos\textsuperscript{1-3$\ast\S$}
  \And
  Carlos R. Medina Temme\textsuperscript{1$\ast$}
  \And
  Jonas Rauber\textsuperscript{2,3$\ast$}
  \AND
  Heiko H. Sch\"utt\textsuperscript{1,4,5}
  \And
  Matthias Bethge\textsuperscript{2,6,7$\ast$}
  \And
  Felix A. Wichmann\textsuperscript{1,2,6,8$\ast$}
  \AND
  \\ \multicolumn{3}{c}{\textsuperscript{1}{Neural Information Processing Group, University of T\"{u}bingen}}
  \\ \multicolumn{3}{c}{\textsuperscript{2}{Centre for Integrative Neuroscience, University of T\"{u}bingen}}
  \\ \multicolumn{3}{c}{\textsuperscript{3}{International Max Planck Research School for Intelligent Systems}}
  \\ \multicolumn{3}{c}{\textsuperscript{4}{Graduate School of Neural and Behavioural Sciences, University of T\"{u}bingen}}
  \\ \multicolumn{3}{c}{\textsuperscript{5}{Department of Psychology, University of Potsdam}}
  \\ \multicolumn{3}{c}{\textsuperscript{6}{Bernstein Center for Computational Neuroscience T\"{u}bingen}}
  \\ \multicolumn{3}{c}{\textsuperscript{7}{Max Planck Institute for Biological Cybernetics}}
  \\ \multicolumn{3}{c}{\textsuperscript{8}{Max Planck Institute for Intelligent Systems}}
  \\ \multicolumn{3}{c}{\textsuperscript{$\ast$}{Joint first / joint senior authors}}
  \\ \multicolumn{3}{c}{\textsuperscript{$\S$}{To whom correspondence should be addressed: \texttt{robert.geirhos@bethgelab.org}}}
}

\begin{document}

\newcommand{\figwidth}{0.24\textwidth}
\newcommand{\captionspace}{-2\baselineskip}
\newcommand{\captionspaceII}{-1.3\baselineskip}

\maketitle

\begin{abstract}
We compare the robustness of humans and current convolutional deep neural networks (DNNs) on object recognition under twelve different types of image degradations.
First, using three well known DNNs (ResNet-152, VGG-19, GoogLeNet) we find the human visual system to be more robust to nearly all of the tested image manipulations, and we observe progressively diverging classification error-patterns between humans and DNNs when the signal gets weaker. Secondly, we show that DNNs trained directly on distorted images consistently surpass human performance on the exact distortion types they were trained on, yet they display extremely poor generalisation abilities when tested on other distortion types. For example, training on salt-and-pepper noise does not imply robustness on uniform white noise and vice versa. Thus, changes in the noise distribution between training and testing constitutes a crucial challenge to deep learning vision systems that can be systematically addressed in a lifelong machine learning approach. Our new dataset consisting of 83K carefully measured human psychophysical trials provide a useful reference for lifelong robustness against image degradations set by the human visual system.
\end{abstract}

\begin{figure}
    \begin{subfigure}{0.32\linewidth}
        \centering
        \textbf{Train\hspace{45pt}Test}
        \includegraphics[width=\linewidth]{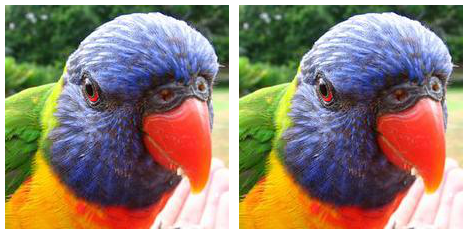}
        \vspace{\captionspaceII}
      \caption{Super-human performance}
    \end{subfigure}\hfill
    \begin{subfigure}{0.32\linewidth}
        \centering
        \textbf{Train\hspace{45pt}Test}
        \includegraphics[width=\linewidth]{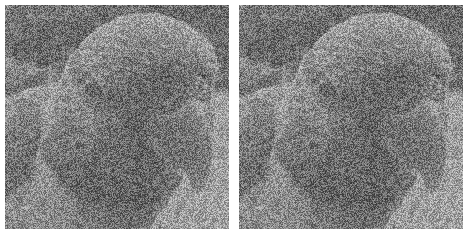}
        \vspace{\captionspaceII}
      \caption{Super-human performance}
    \end{subfigure}\hfill
    \begin{subfigure}{0.32\linewidth}
        \centering
        \textbf{Train\hspace{45pt}Test}
        \includegraphics[width=\linewidth]{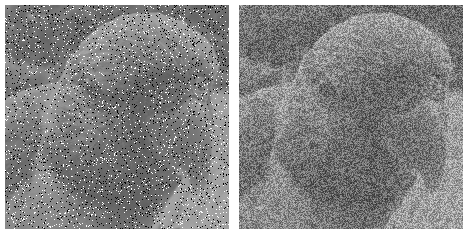}
        \vspace{\captionspaceII}
      \caption{Chance level performance}
      \label{fig:introduction_non_iid}
    \end{subfigure}\hfill
    \caption{Classification performance of ResNet-50 trained from scratch on (potentially distorted) ImageNet images. \textbf{(a)}~Classification performance when trained on standard colour images and tested on colour images is close to perfect (better than human observers).
    \textbf{(b)}~Likewise, when trained and tested on images with additive uniform noise, performance is super-human.
    \textbf{(c)}~Striking generalisation failure: When trained on images with salt-and-pepper noise and tested on images with uniform noise, performance is at chance level---even though both noise types do not seem much different to human observers.}
    \label{fig:introduction}
\end{figure}

\section{Introduction}

\subsection{Deep neural networks as models of human object recognition}

The visual recognition of objects by humans in everyday life is rapid and seemingly effortless, as well as largely independent of viewpoint and object orientation \cite{Biederman1987}. The rapid and primarily foveal recognition during a single fixation has been termed \textit{core object recognition} (see \cite{DiCarlo2012} for a review). We know, for example, that it is possible to reliably identify objects in the central visual field within a single fixation in less than 200 ms when viewing ``standard'' images \cite{DiCarlo2012, Potter1976, Thorpe1996}. Based on the rapidness of object recognition, core object recognition is often thought to be achieved with mainly feedforward processing although feedback connections are ubiquitous in the primate brain. Object recognition in the primate brain is believed to be realised by the ventral visual pathway, a hierarchical structure consisting of areas V1-V2-V4-IT, with information from the retina reaching the cortex in V1 (e.g. \cite{goodale1992}).
 
Until a few years ago, animate visual systems were the only ones known to be capable of broad-ranging visual object recognition. This has changed, however, with the advent of brain-inspired deep neural networks (DNNs) which, after having been trained on millions of labeled images, achieve human-level performance when classifying objects in images of natural scenes \cite{Krizhevsky2012}. DNNs are now employed on a variety of tasks and set the new state-of-the-art, sometimes even surpassing human performance on tasks which only a few years ago were thought to be beyond an algorithmic solution for decades to come \cite{He2015, Silver2016}. Since DNNs and humans achieve similar accuracy, a number of studies have started investigating similarities and differences between DNNs and human vision \cite{Cadieu2014, Yamins2014, Cichy2016, Kheradpisheh2016, Dodge2017a, Dodge2017b, Dekel2017, Pramod2016, Karimi2017, Rosenfeld2018, Flachot2018, Wallis2017, Berardino2017, Jozwik2017, Lake2017, Kietzmann2017}. On the one hand, the network units are an enormous simplification given the sophisticated nature and diversity of neurons in the brain \cite{Douglas1991}. On the other hand, often the strength of a model lies not in replicating the original system but rather in its ability to capture the important aspects while abstracting from details of the implementation (e.g. \cite{Box1976, Kriegeskorte2015}).

One of the most remarkable properties of the human visual system is its ability to generalise robustly. Humans generalise across a wide variety of changes in the input distribution, such as across different illumination conditions and weather types. For instance, human object recognition is largely unimpaired even if there are rain drops or snow flakes in front of an object. While humans are certainly exposed to a large number of such changes during their preceding lifetime  (i.e., at ``training time'', as we would say for DNNs), there seems to be something very generic about the way the human visual system is able to generalise that is not limited to the same distribution one was exposed to previously. Otherwise we would not be able to make sense of a scene if there was some sort of ``new'', previously unseen noise. Even if one never had a shower of confetti before, one is still able to effortlessly recognise objects at a carnival parade. Naturally, such generic, robust mechanisms are not only desirable for animate visual systems but also for solving virtually any visual task that goes beyond a well-confined setting where one knows the exact test distribution already at training time. Deep learning for autonomous driving may be one prominent example: one would like to achieve robust classification performance in the presence of confetti, despite not having had any confetti exposure during training time. Thus, from a machine learning perspective, general noise robustness can be used as a highly relevant example of lifelong machine learning \cite{Chen_Liu2016} requiring generalisation that does not rely on the standard assumption of independent, identically distributed (i.i.d.) samples at test time.

\subsection{Comparing generalisation abilities}
Generalisation in DNNs usually works surprisingly well: First of all, DNNs are able to learn sufficiently general features on the training distribution to achieve a high accuracy on the i.i.d. test distribution despite having sufficient capacity to completely memorise the training data \cite{Zhang2016}, and considerable effort has been devoted to understand this phenomenon (e.g. \cite{Kawaguchi2017, Neyshabur2017, Shwartz2017}).\footnote{Still, DNNs usually need orders of magnitude more training data in comparison to humans, as explored by the literature on one-shot or few-shot learning (see e.g. \cite{Lake2017} for an overview).} Secondly, features learned on one task often transfer to only loosely related tasks, such as from classification to saliency prediction \cite{Kuemmerer2016}, emotion recognition \cite{Ng2015}, medical imaging \cite{Greenspan2016} and a large number of other transfer learning tasks \cite{Donahue2014}. However, transfer learning still requires a substantial amount of training before it works on the new task. Here, we focus on a third setting that adopts the lifelong machine learning point of view of generalisation \cite{Thrun1996}: How well can a visual learning system cope with a new image degradation after it has learned to cope with a certain set of image distortions before. As a measure of object recognition robustness we can test the ability of a classifier or visual system to tolerate changes in the input distribution up to a certain degree, i.e., to achieve high recognition performance despite being evaluated on a test distribution that differs to some degree from the training distribution (testing under realistic, non-i.i.d. conditions). Using this approach we measure how well DNNs and human observers cope with parametric image manipulations that gradually distort the original image.

\begin{figure}
    \includegraphics[width=\linewidth]{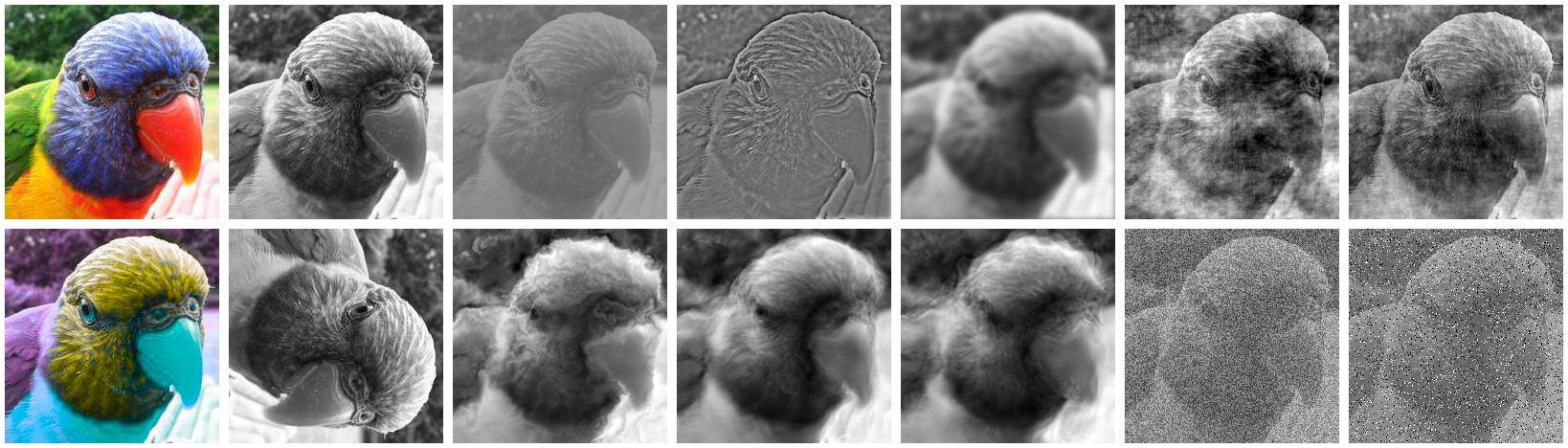}
    \vspace{\captionspaceII}
     \caption{Example stimulus image of class \texttt{bird} across all distortion types. From left to right, image manipulations are: colour (undistorted), greyscale, low contrast, high-pass, low-pass (blurring), phase noise, power equalisation. Bottom row: opponent colour, rotation, Eidolon I, II and III, additive uniform noise, salt-and-pepper noise. Example stimulus images across all used distortion levels are available in the supplementary material.}
    \label{fig:all_stimuli}
\end{figure}

First, we assess how top-performing DNNs that are trained on ImageNet,  GoogLeNet \cite{Szegedy2015}, VGG-19 \cite{Simonyan2015} and ResNet-152 \cite{He2016}, compare against human observers when tested on twelve different distortions such as additive noise or phase noise (see Figure~\ref{fig:all_stimuli} for an overview)---in other words, how well do they generalise towards previously unseen distortions.\footnote{We have reported a subset of these experiments on arXiv in an earlier version of this paper \cite{Geirhos2017}.} In a second set of experiments, we train networks directly on distorted images to see how well they can in general cope with noisy input, and how much training on distortions as a form of data augmentation helps in dealing with other distortions.
Psychophysical investigations of human behaviour on object recognition tasks, measuring accuracies depending on image colour (greyscale vs. colour), image contrast  and the amount of additive visual noise have been powerful means of exploring the human visual system, revealing much about the internal computations and mechanisms at work \cite{Nachmias1974,Pelli1999,Wichmann1999,Henning2002,Carandini2012,Carandini1997,Delorme2000}. As a consequence, similar experiments might yield equally interesting insights into the functioning of DNNs, especially as a comparison to high-quality measurements of human behaviour. In particular, human data for our experiments were obtained using a controlled lab environment (instead of e.g. Amazon Mechanical Turk without sufficient control about presentation times, display calibration, viewing angles, and sustained attention of participants). Our carefully measured behavioural datasets---twelve experiments encompassing a total number of 82,880 psychophysical trials---as well as materials and code are available online at \url{https://github.com/rgeirhos/generalisation-humans-DNNs}.

\section{Methods}
We here report the core elements of employed paradigm, procedure, image manipulations, observers and DNNs; this is aimed at giving the reader just enough information to understand experiments and results. For in-depth explanations we kindly refer to the comprehensive supplementary material, which seeks to provide exhaustive and reproducible experimental details.

\subsection{Paradigm, procedure \& 16-class-ImageNet}
\label{meth:paradigm_procedure_images}
For this study, we developed an experimental paradigm aimed at comparing human observers and DNNs as fair as possible by using a forced-choice image categorisation task.\footnote{This is the same paradigm as reported in \cite{Wichmann2017}.}
Achieving a fair psychophysical comparison comes with a number of challenges: First of all, many high-performing DNNs are trained on the ILSRVR 2012 database \cite{Russakovsky2015} with 1,000 fine-grained categories (e.g., over a hundred different \texttt{dog} breeds). If humans are asked to name objects, however, they most naturally categorise them into so-called entry-level categories (e.g. \texttt{dog} rather than \texttt{German shepherd}). We thus developed a mapping from 16 entry-level categories such as \texttt{dog}, \texttt{car} or \texttt{chair} to their corresponding ImageNet categories using the WordNet hierarchy \cite{Miller1995}. We term this dataset ``16-class-ImageNet'' since it groups a subset of ImageNet classes into 16 entry-level categories (\texttt{airplane, bicycle, boat, car, chair, dog, keyboard, oven, bear, bird, bottle, cat, clock, elephant, knife, truck}). In every experiment, then, an image was presented on a computer screen and observers had to choose the correct category by clicking on one of these 16 categories. For pre-trained DNNs, the sum of all softmax values mapping to a certain entry-level category was computed. The entry-level category with the highest sum was then taken as the network's decision.\footnote{This aggregation method is suboptimal. We later derived the optimal method (see appendix).}
A second challenge is the fact that standard DNNs only use feedforward computations at inference time, while recurrent connections are ubiquitous in the human brain \cite{Lamme1998, Sporns2004}.\footnote{But see e.g. \cite{Gerstner2005} for a critical assessment of this argument.} In order to prevent this discrepancy from playing a major confounding role in our experimental comparison, presentation time for human observers was limited to 200 ms. An image was immediately followed by a 200 ms presentation of a noise mask with 1/\textit{f} spectrum, known to minimise, as much as psychophysically possible, feedback influence in the brain.

\subsection{Observers \& pre-trained deep neural networks}
Data from human observers were compared against classification performance of three pre-trained DNNs: VGG-19 \cite{Simonyan2015}, GoogLeNet \cite{Szegedy2015} and ResNet-152 \cite{He2016}.  For each of the twelve experiments that were conducted, either five or six observers participated (with the exception of the colour experiment, for which only three observers participated since similar experiments had already been performed by a number of studies \cite{Delorme2000, Kubilius2016, Wichmann2006}). Observers reported normal or corrected-to-normal vision.

\begin{figure}
    \begin{subfigure}{\figwidth}
        \centering
        \textbf{Accuracy}\\
        \includegraphics[width=\linewidth]{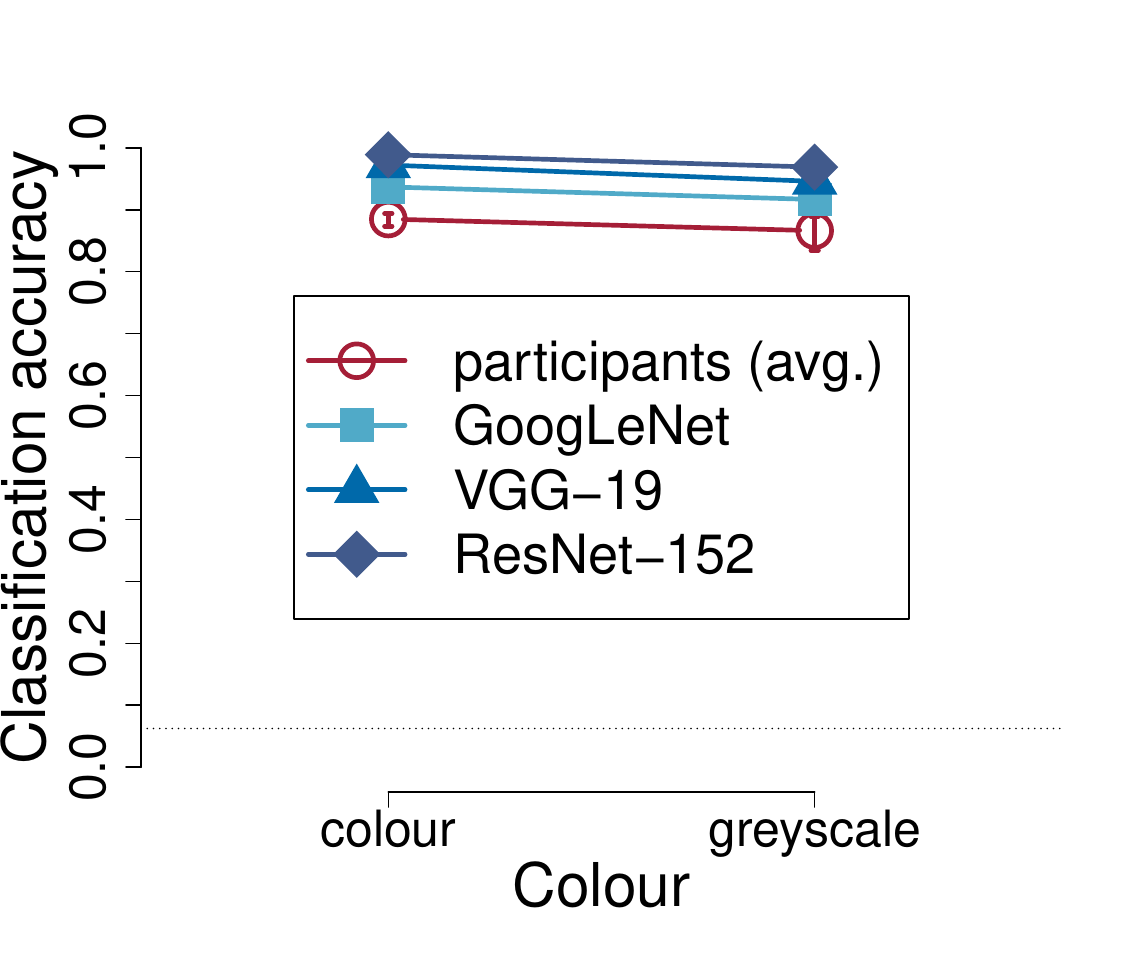}
        \vspace{\captionspace}
        \caption{Colour vs. greyscale}
    \end{subfigure}\hfill
    \begin{subfigure}{\figwidth}
        \centering
        \textbf{Entropy}\\
        \includegraphics[width=\linewidth]{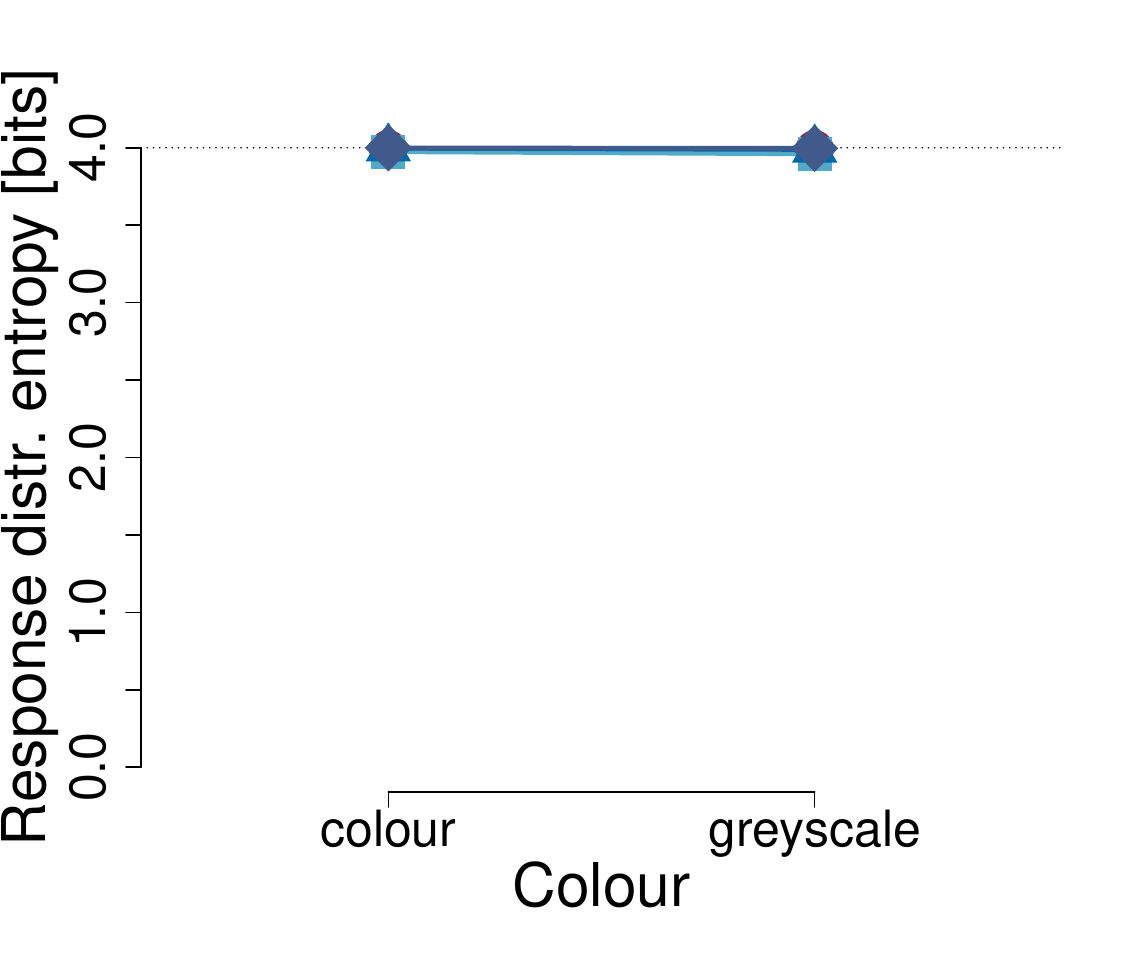}
        \vspace{\captionspace}
        \caption*{}
    \end{subfigure}\hfill
    \begin{subfigure}{\figwidth}
        \centering
        \textbf{Accuracy}\\
        \includegraphics[width=\linewidth]{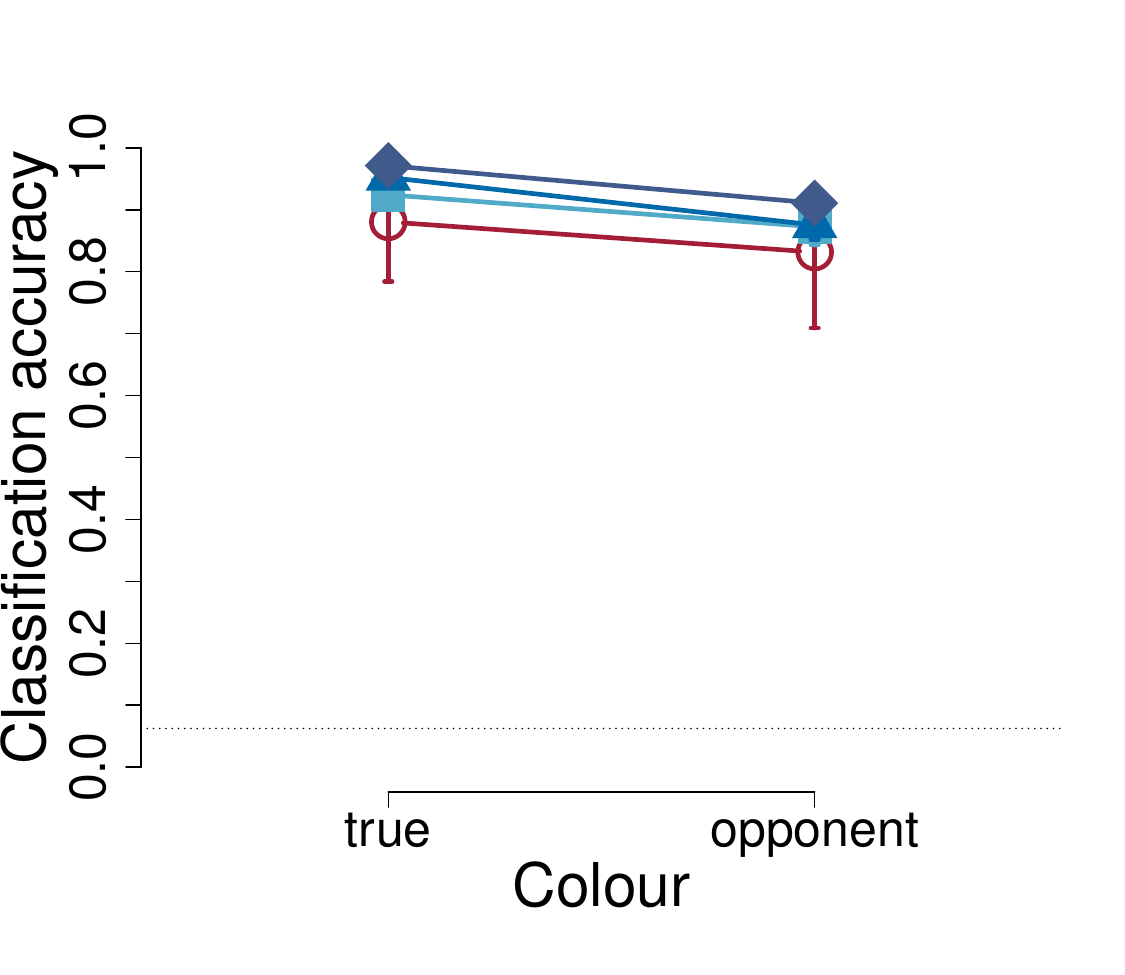}
        \vspace{\captionspace}
        \caption{True vs. false colour}
    \end{subfigure}\hfill
    \begin{subfigure}{\figwidth}
        \centering
        \textbf{Entropy}\\
        \includegraphics[width=\linewidth]{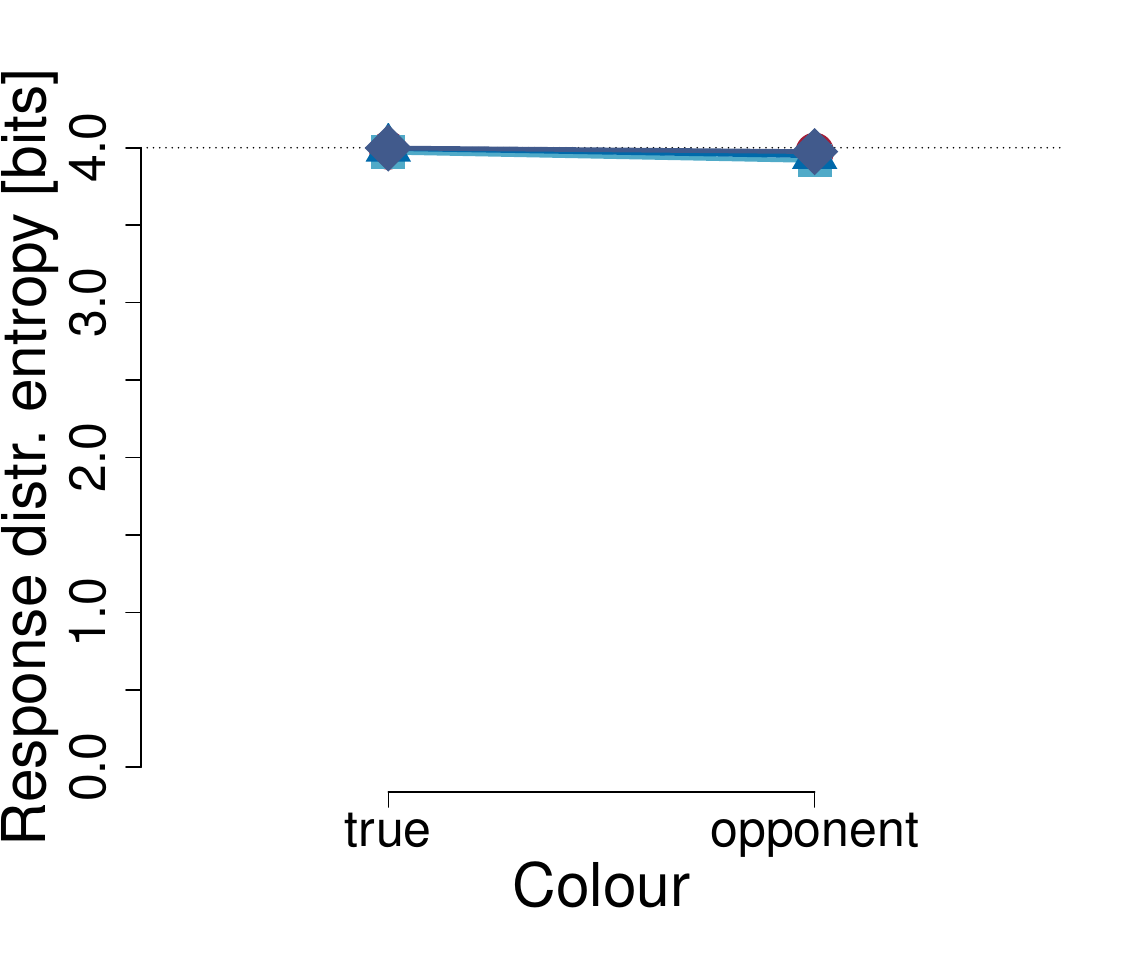}
        \vspace{\captionspace}
        \caption*{}
    \end{subfigure}\hfill

    \begin{subfigure}{\figwidth}
        \centering
        \includegraphics[width=\linewidth]{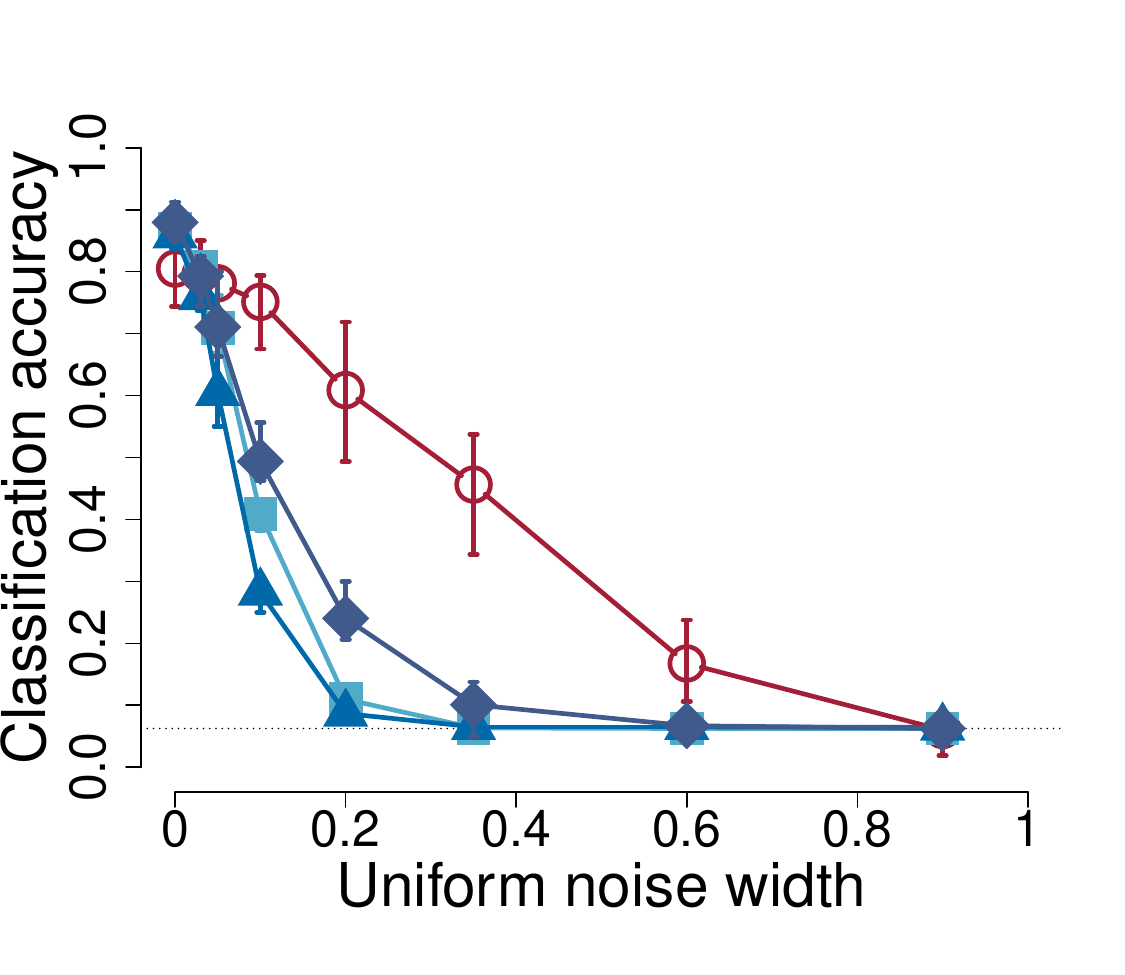}
        \vspace{\captionspace}
        \caption{Uniform noise}
    \end{subfigure}\hfill
    \begin{subfigure}{\figwidth}
        \centering
        \includegraphics[width=\linewidth]{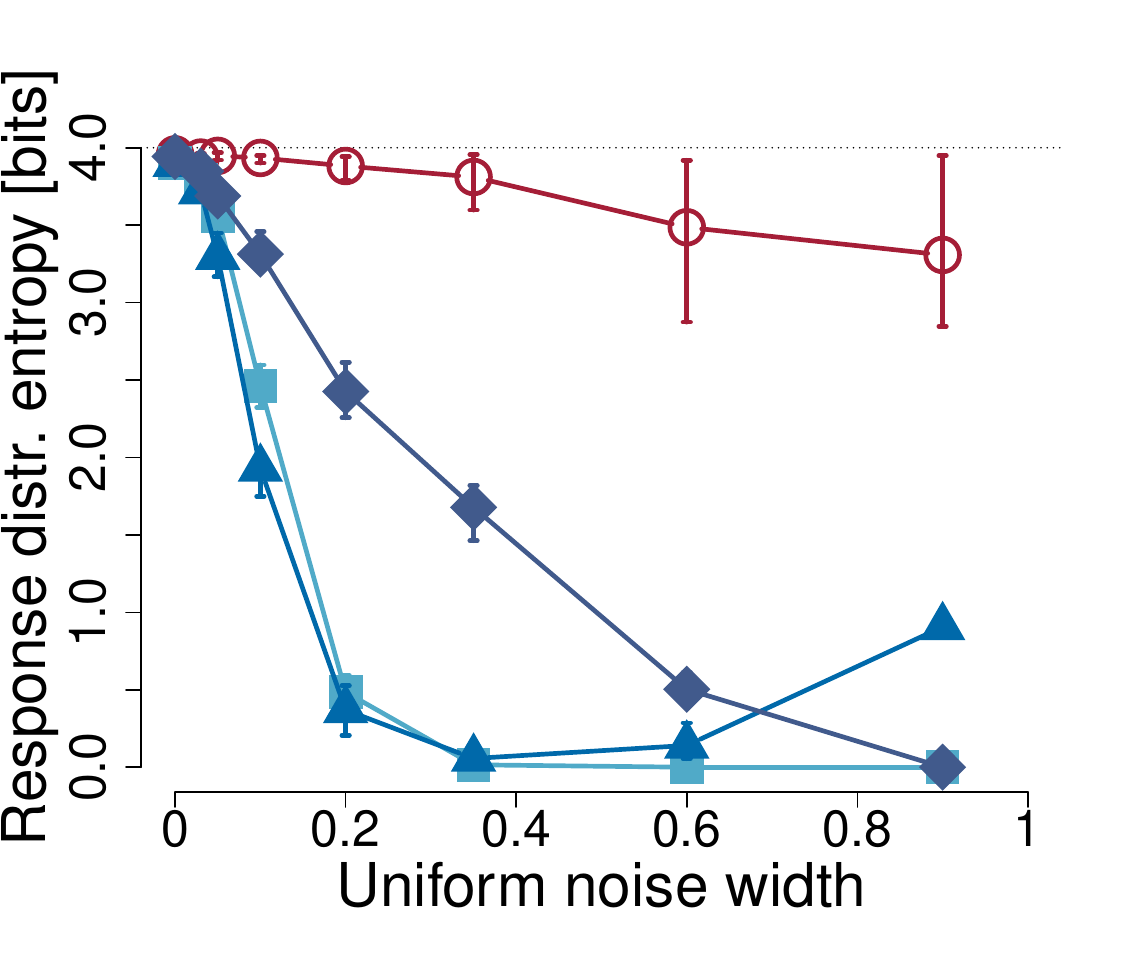}
        \vspace{\captionspace}
        \caption*{}
    \end{subfigure}\hfill
    \begin{subfigure}{\figwidth}
        \centering
        \includegraphics[width=\linewidth]{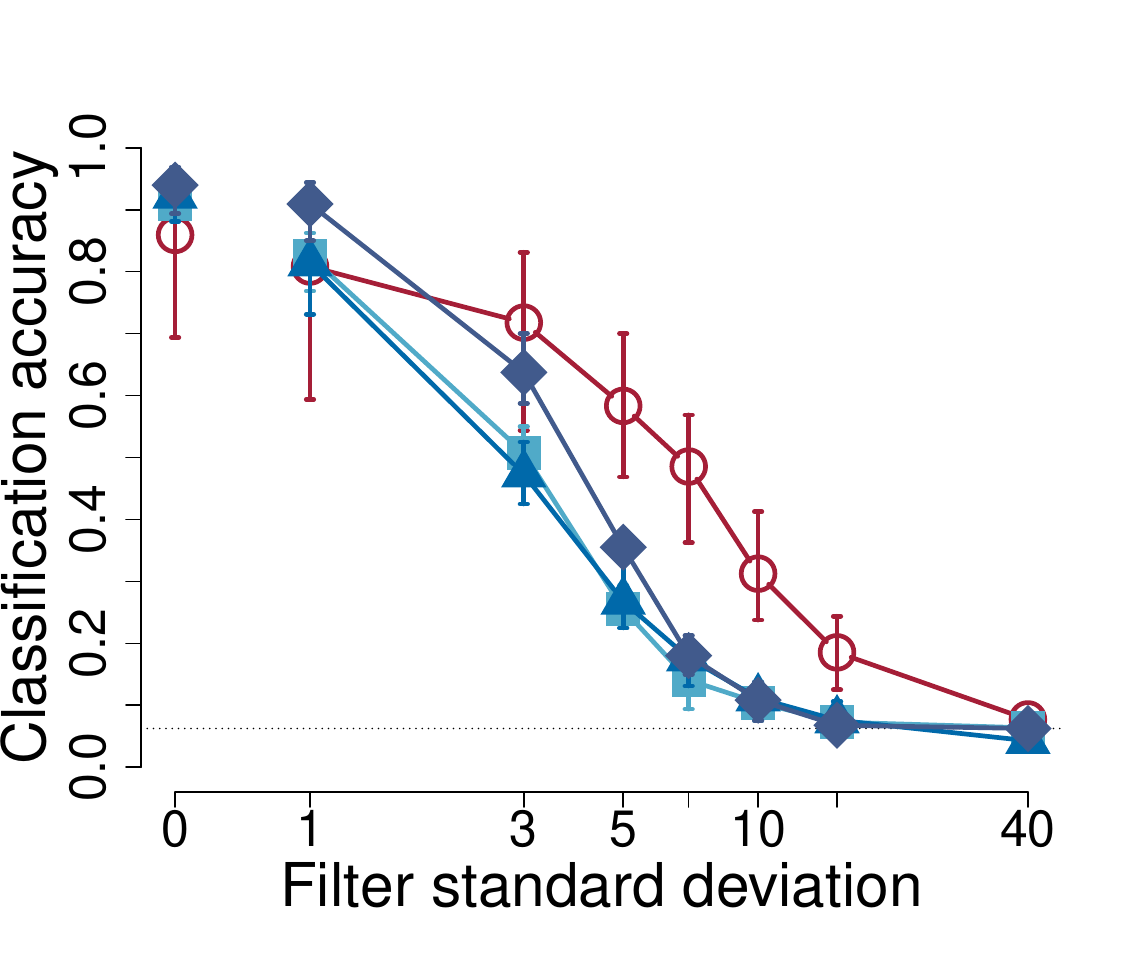}
        \vspace{\captionspace}
        \caption{Low-pass}
    \end{subfigure}\hfill
    \begin{subfigure}{\figwidth}
        \centering
        \includegraphics[width=\linewidth]{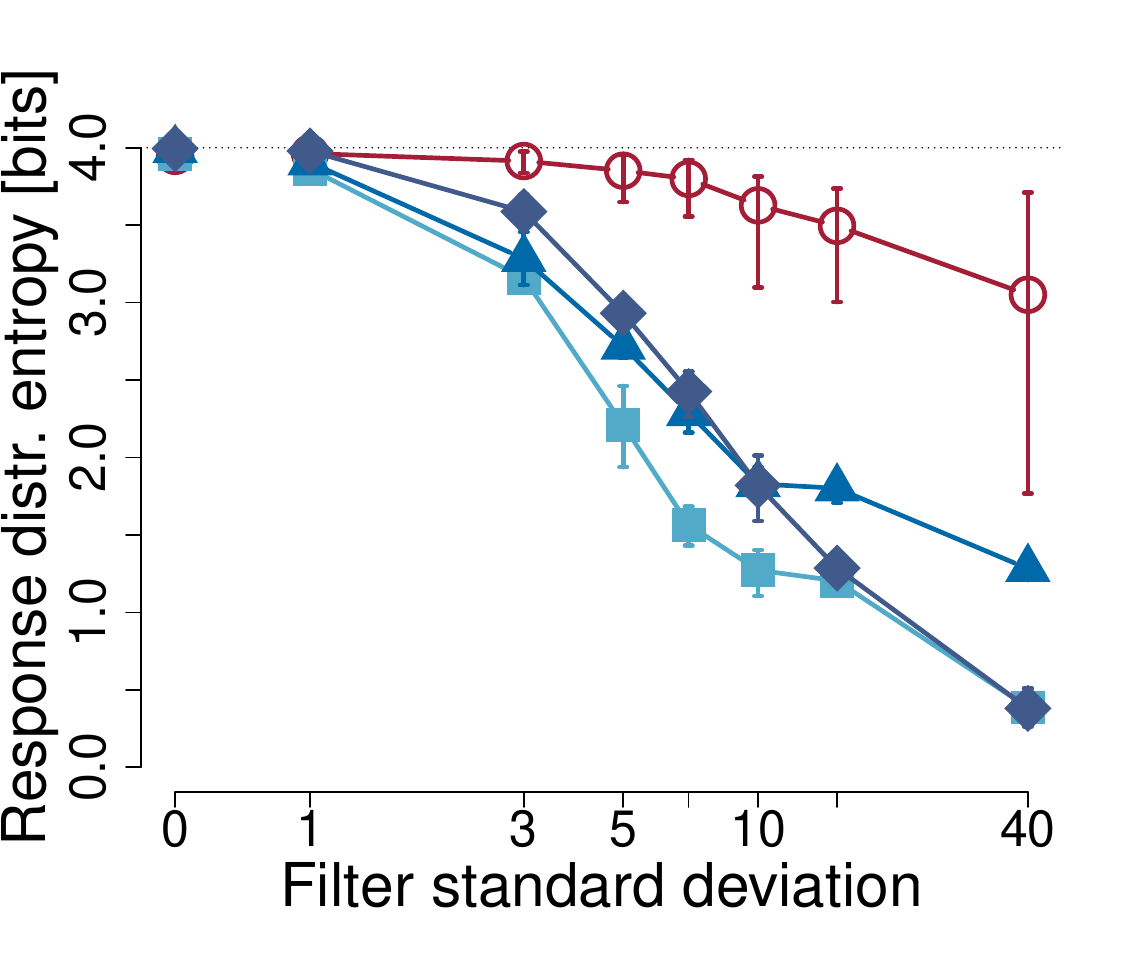}
        \vspace{\captionspace}
        \caption*{}
    \end{subfigure}\hfill

    \begin{subfigure}{\figwidth}
        \centering
        \includegraphics[width=\linewidth]{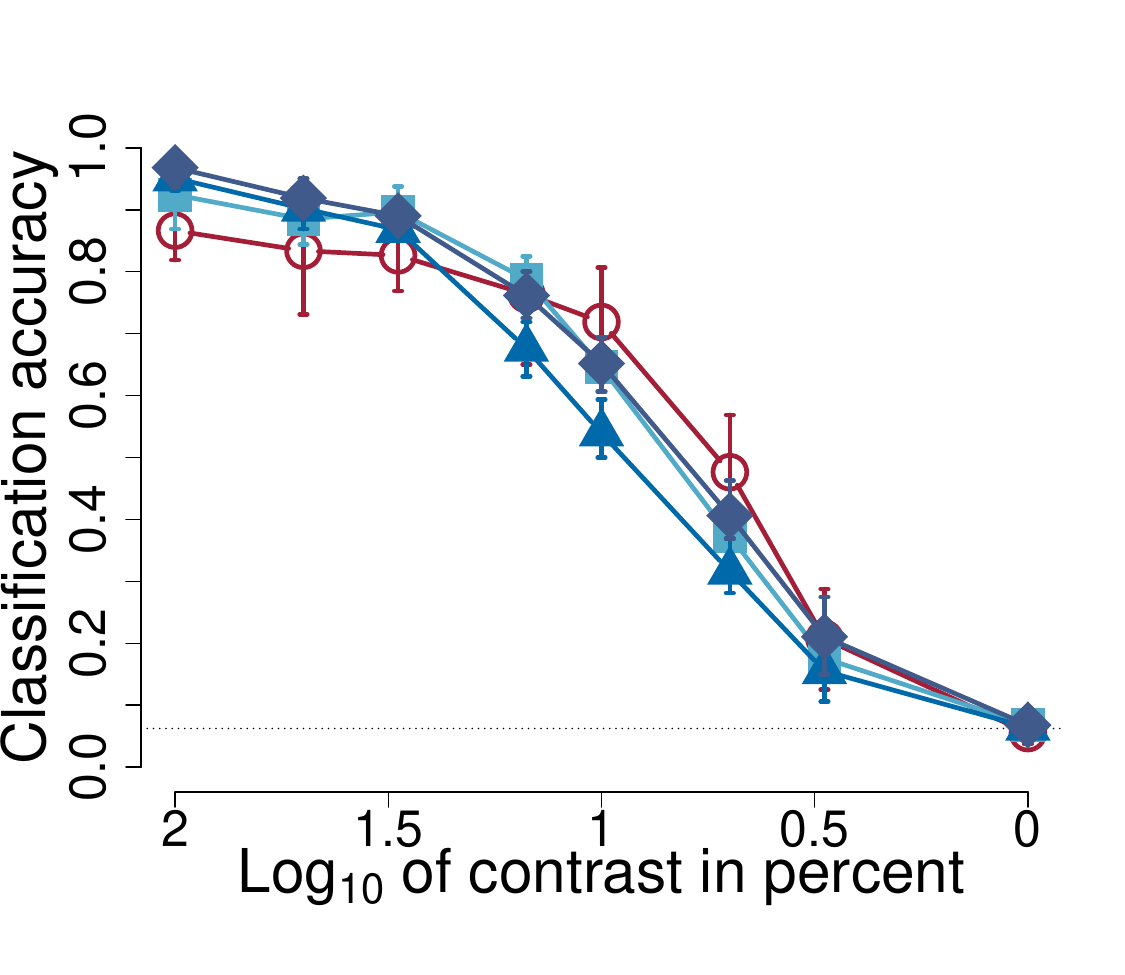}
        \vspace{\captionspace}
        \caption{Contrast}
    \end{subfigure}\hfill
    \begin{subfigure}{\figwidth}
        \centering
        \includegraphics[width=\linewidth]{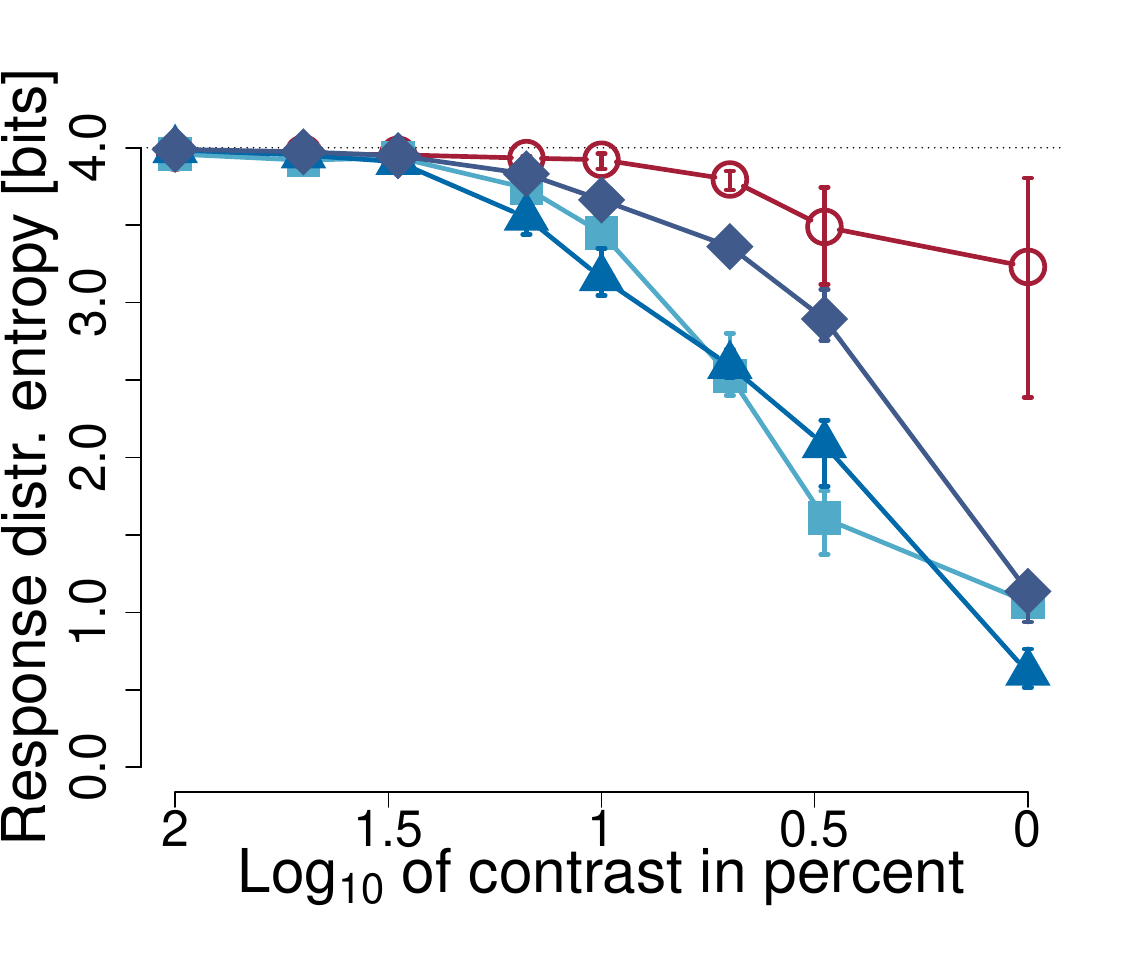}
        \vspace{\captionspace}
        \caption*{}
    \end{subfigure}\hfill
    \begin{subfigure}{\figwidth}
        \centering
        \includegraphics[width=\linewidth]{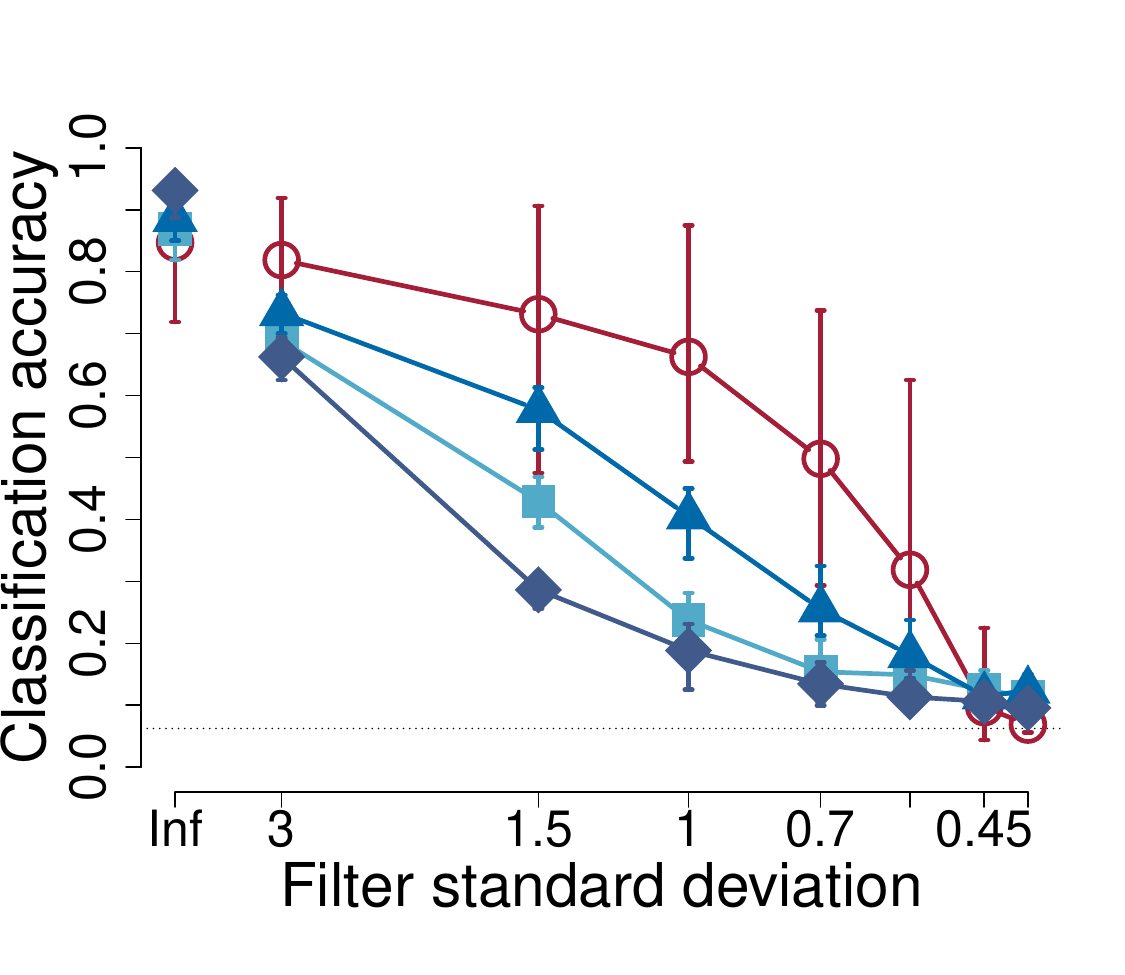}
        \vspace{\captionspace}
        \caption{High-pass}
    \end{subfigure}\hfill
    \begin{subfigure}{\figwidth}
        \centering
        \includegraphics[width=\linewidth]{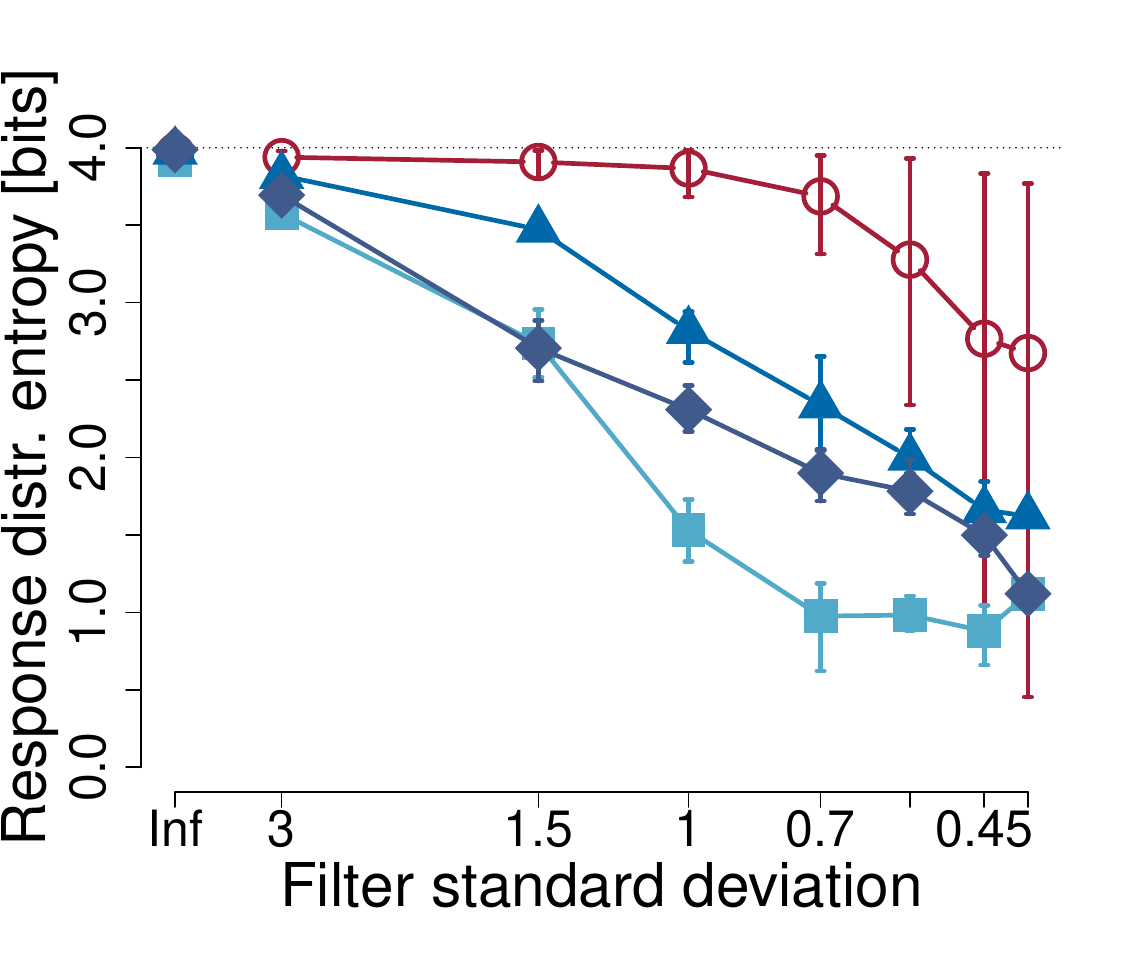}
        \vspace{\captionspace}
        \caption*{}
    \end{subfigure}\hfill

    \begin{subfigure}{\figwidth}
        \centering
        \includegraphics[width=\linewidth]{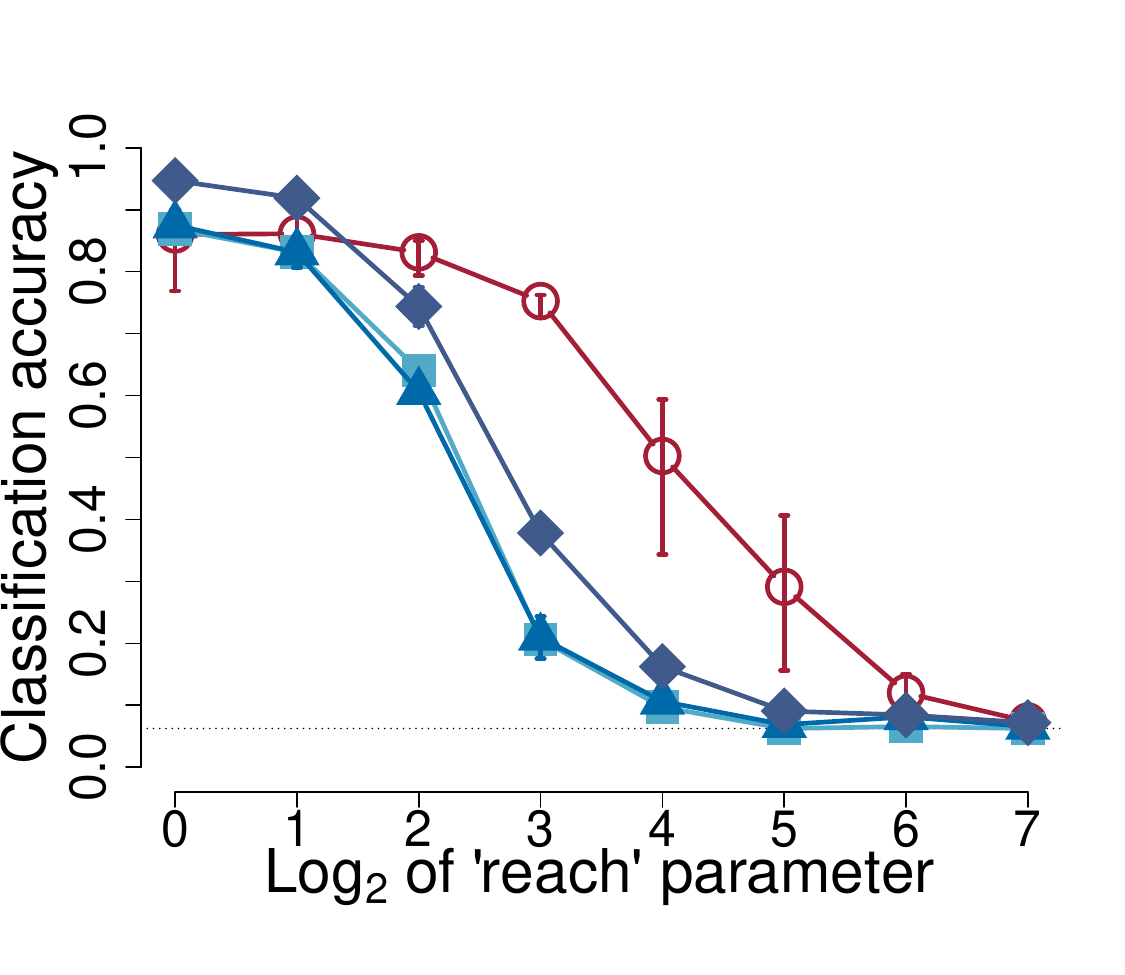}
        \vspace{\captionspace}
        \caption{Eidolon I}
    \end{subfigure}\hfill
    \begin{subfigure}{\figwidth}
        \centering        \includegraphics[width=\linewidth]{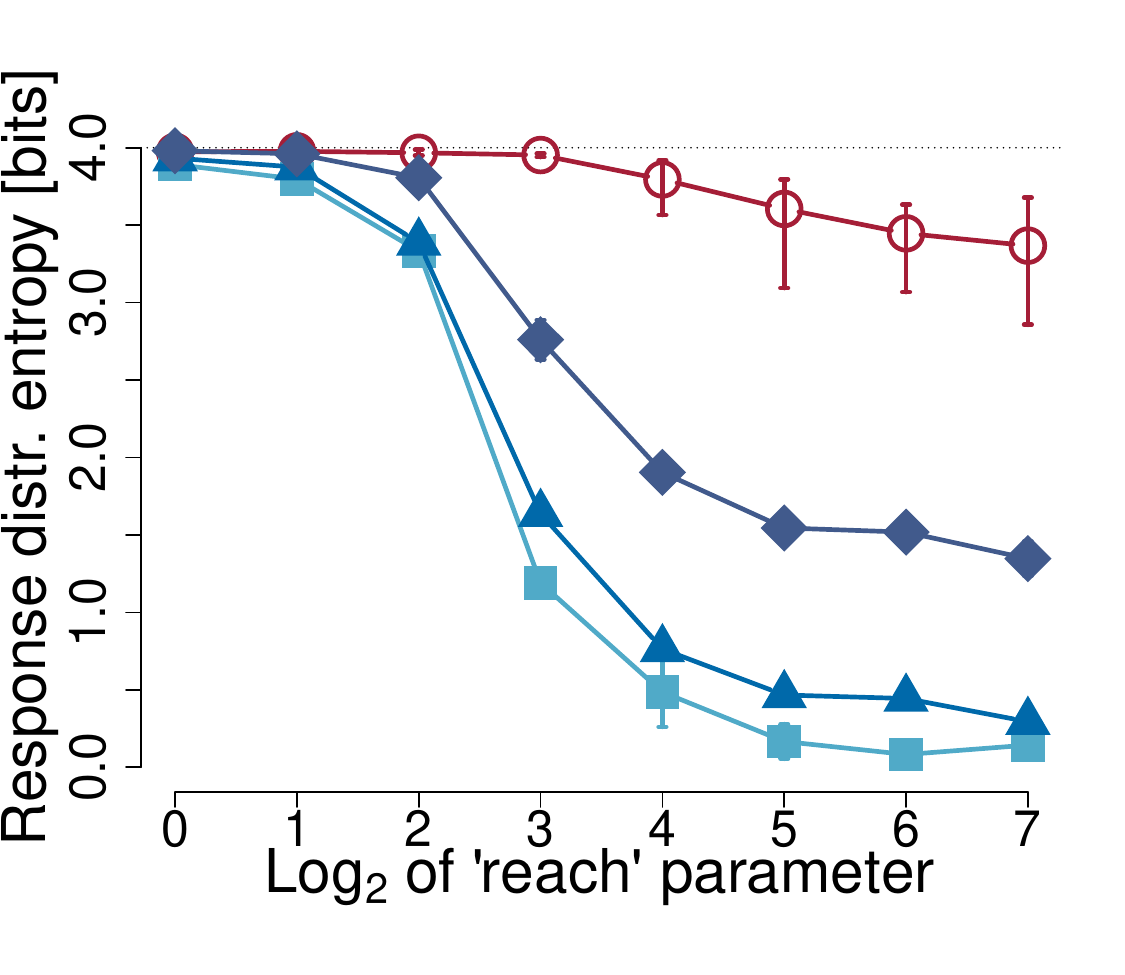}
        \vspace{\captionspace}
        \caption*{}
    \end{subfigure}\hfill
    \begin{subfigure}{\figwidth}
        \centering
        \includegraphics[width=\linewidth]{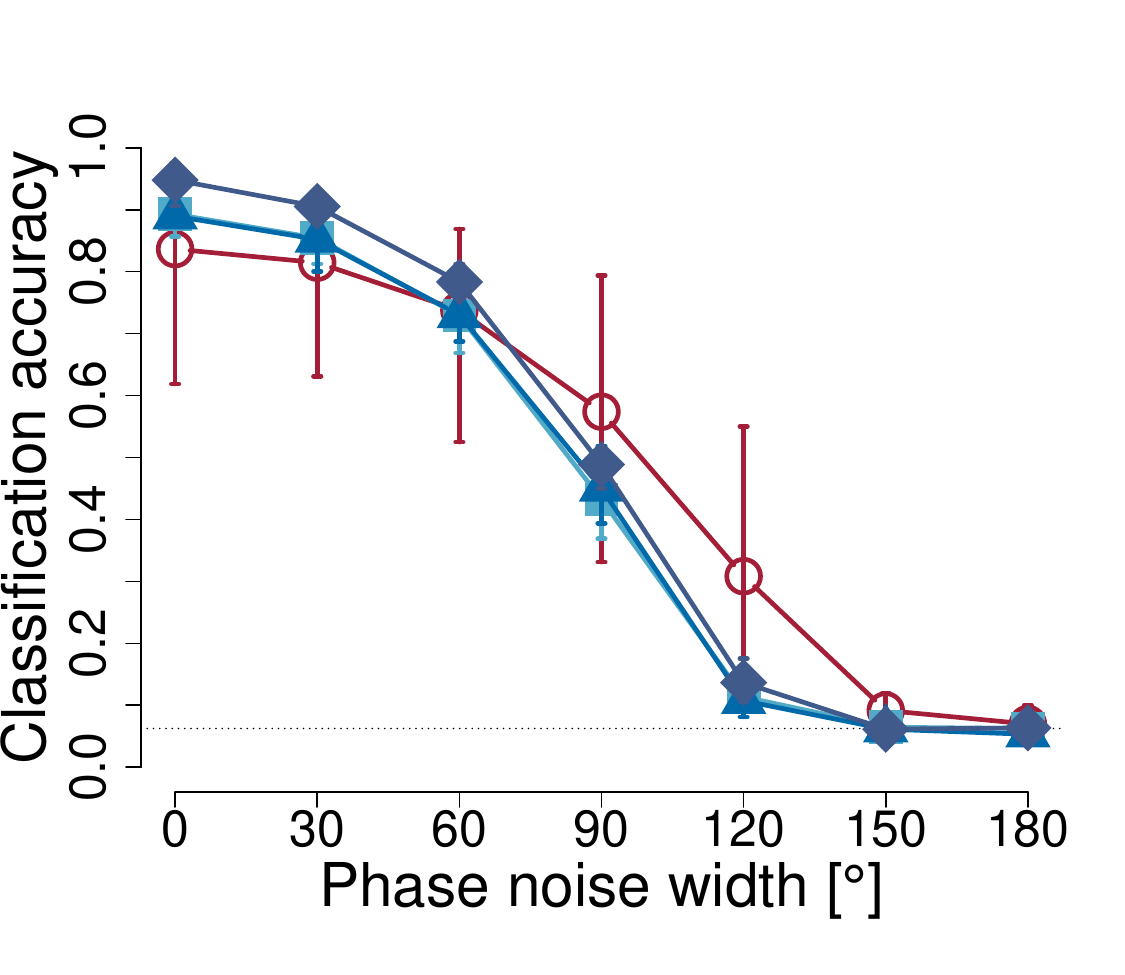}
        \vspace{\captionspace}
        \caption{Phase noise}
    \end{subfigure}\hfill
    \begin{subfigure}{\figwidth}
        \centering
        \includegraphics[width=\linewidth]{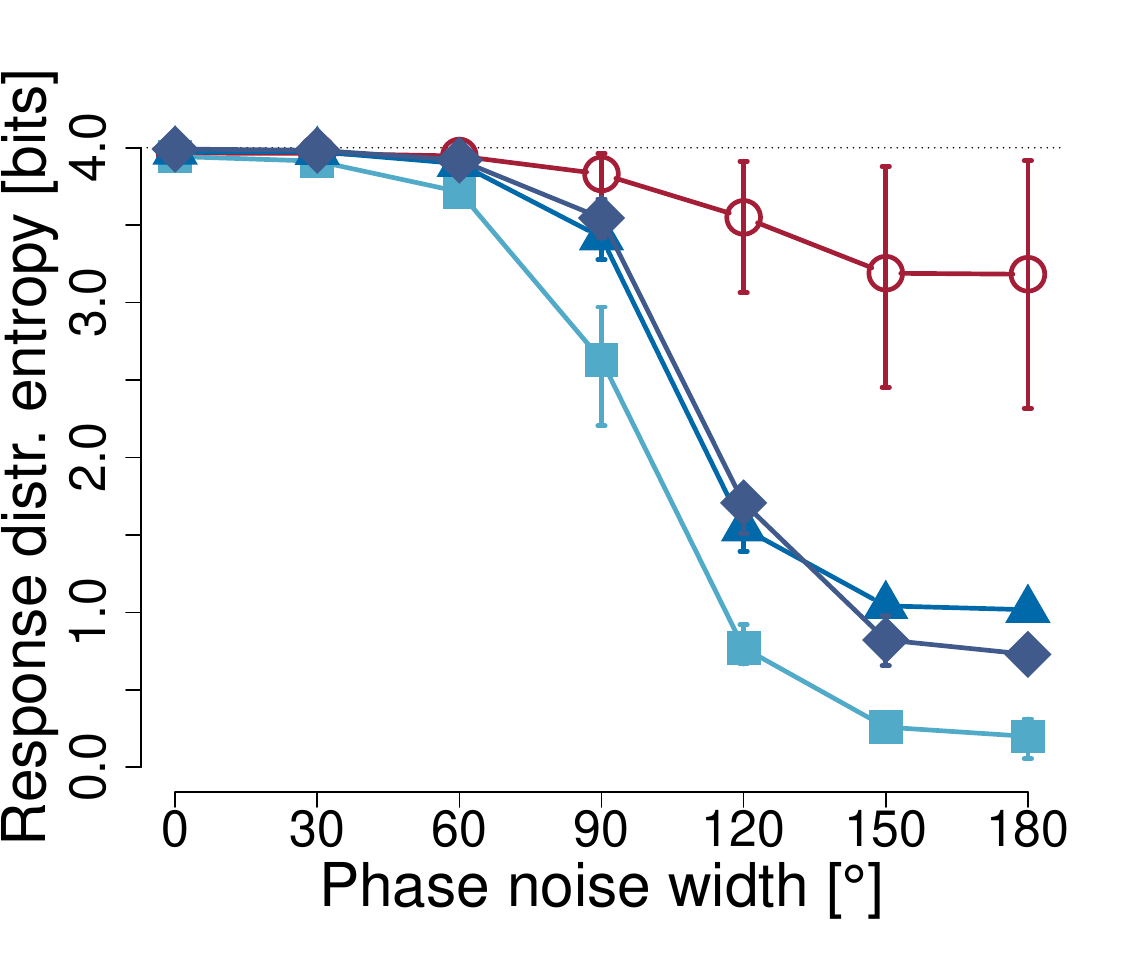}
        \vspace{\captionspace}
        \caption*{}
    \end{subfigure}\hfill

    \begin{subfigure}{\figwidth}
        \centering
        \includegraphics[width=\linewidth]{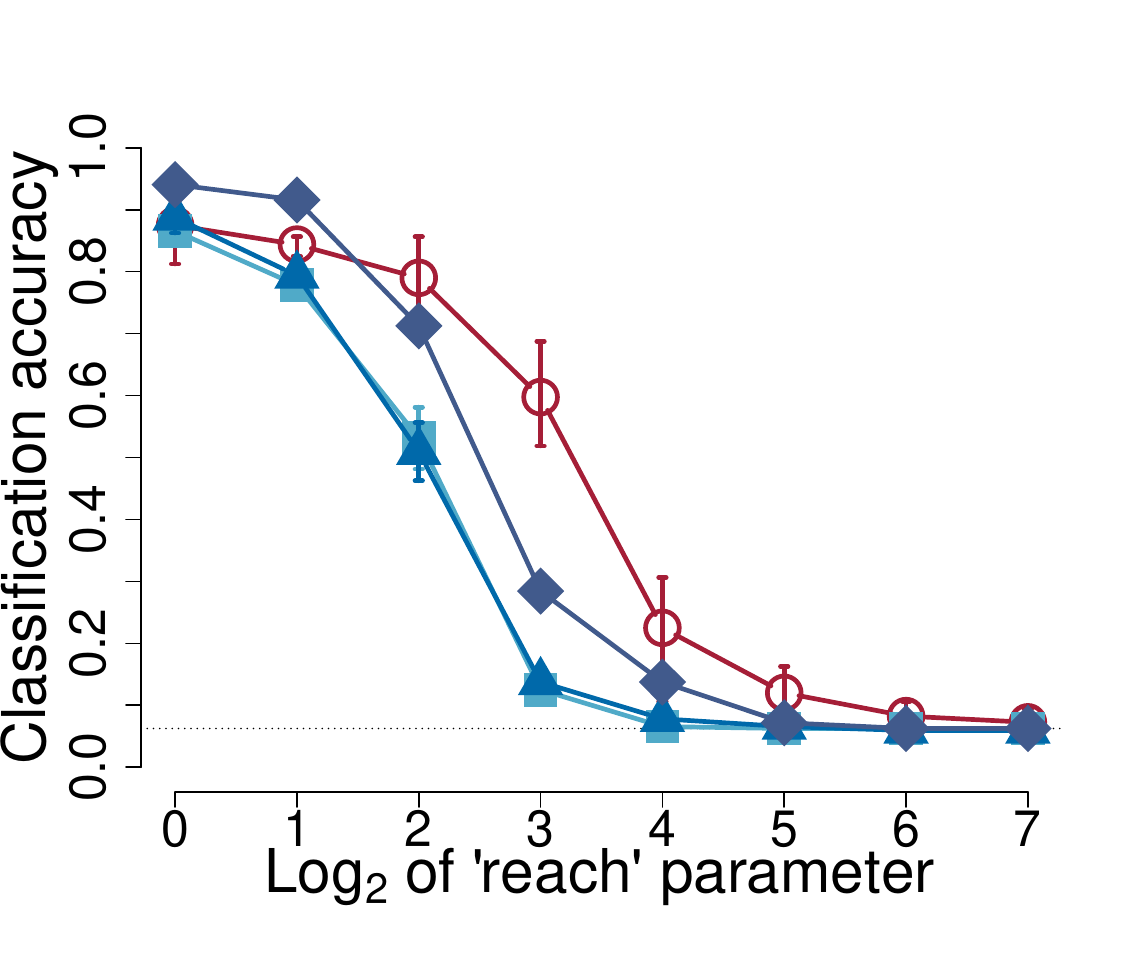}
        \vspace{\captionspace}
        \caption{Eidolon II}
    \end{subfigure}\hfill
    \begin{subfigure}{\figwidth}
        \centering
        \includegraphics[width=\linewidth]{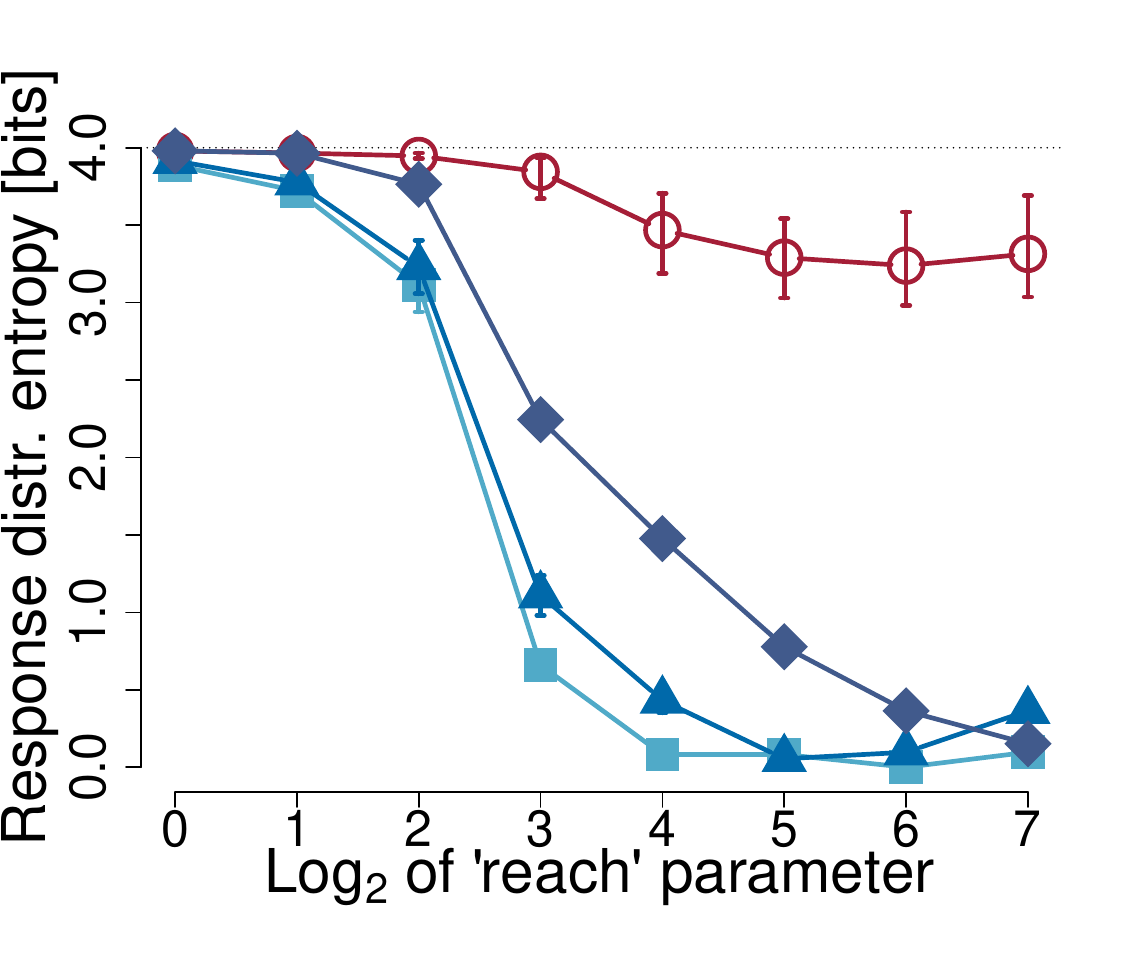}
        \vspace{\captionspace}
        \caption*{}
    \end{subfigure}\hfill
    \begin{subfigure}{\figwidth}
        \centering
        \includegraphics[width=\linewidth]{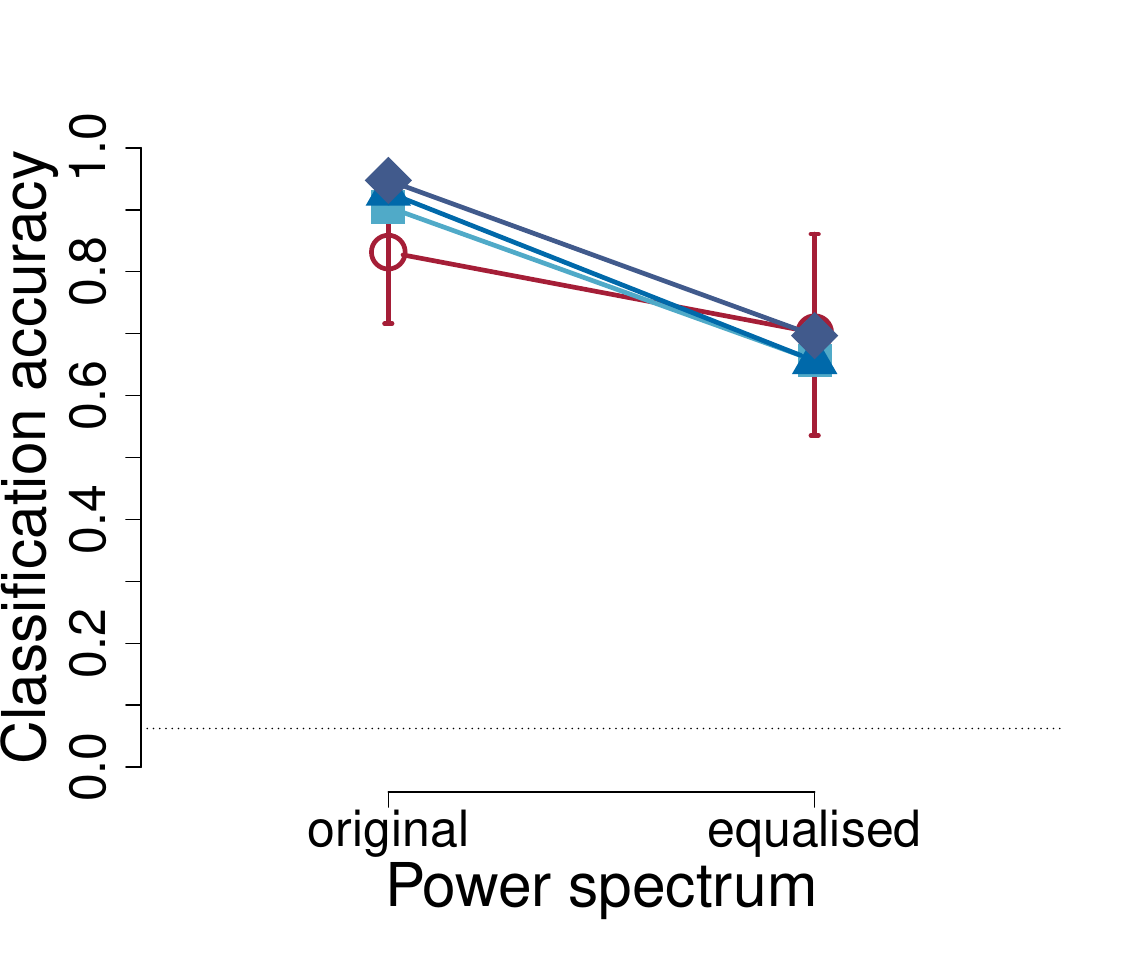}
        \vspace{\captionspace}
        \caption{Power equalisation}
    \end{subfigure}\hfill
    \begin{subfigure}{\figwidth}
        \centering
        \includegraphics[width=\linewidth]{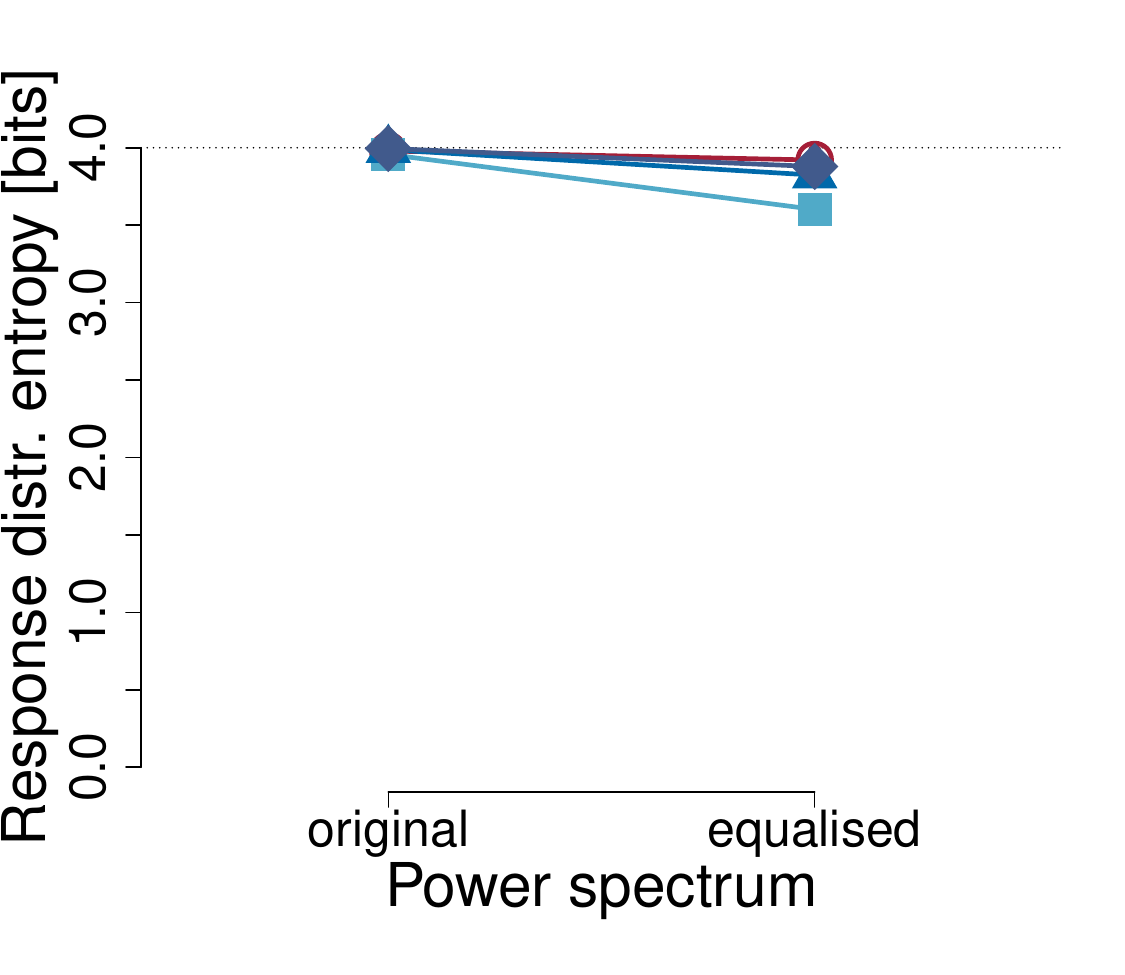}
        \vspace{\captionspace}
        \caption*{}
    \end{subfigure}\hfill

    \begin{subfigure}{\figwidth}
        \centering
        \includegraphics[width=\linewidth]{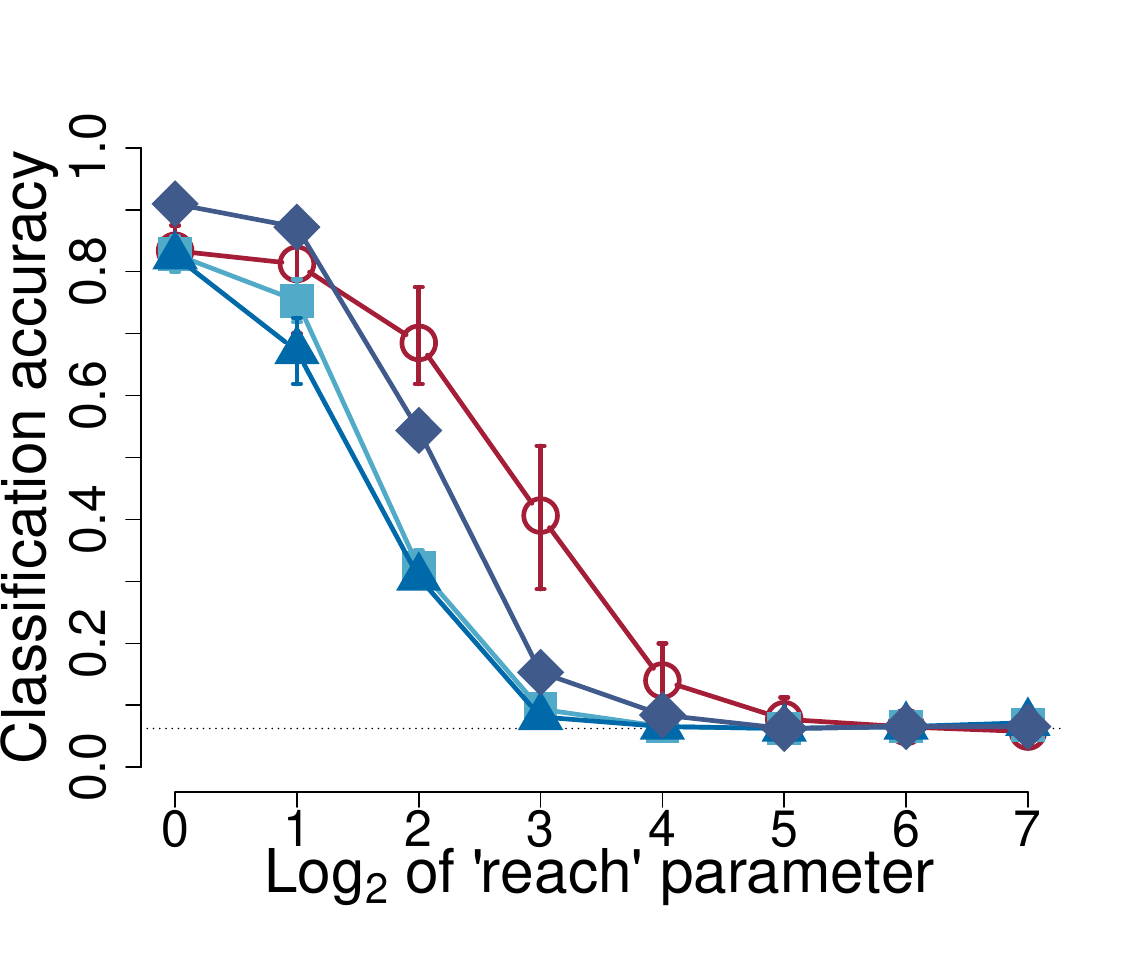}
        \vspace{\captionspace}
        \caption{Eidolon III}
    \end{subfigure}\hfill
    \begin{subfigure}{\figwidth}
        \centering        \includegraphics[width=\linewidth]{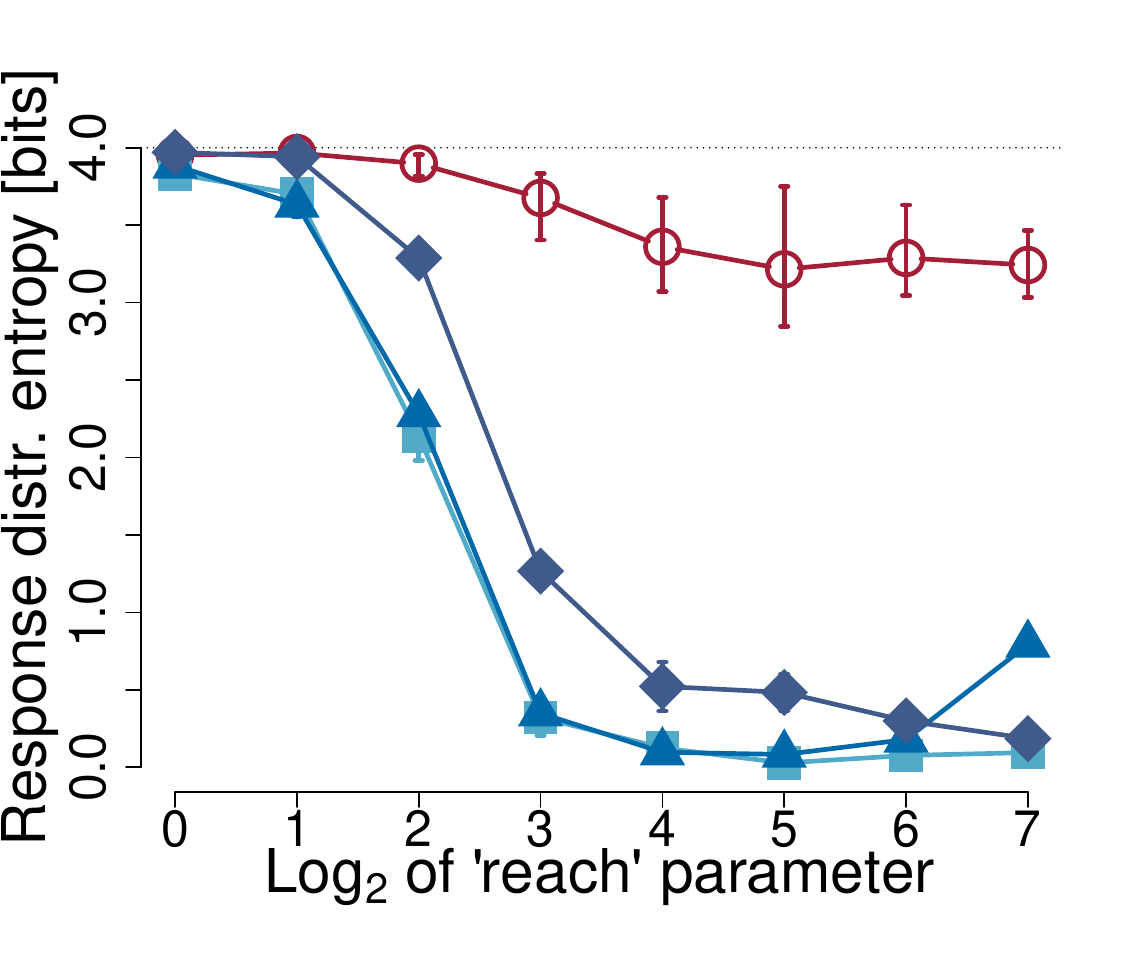}
        \vspace{\captionspace}
        \caption*{}
    \end{subfigure}\hfill
    \begin{subfigure}{\figwidth}
        \centering
        \includegraphics[width=\linewidth]{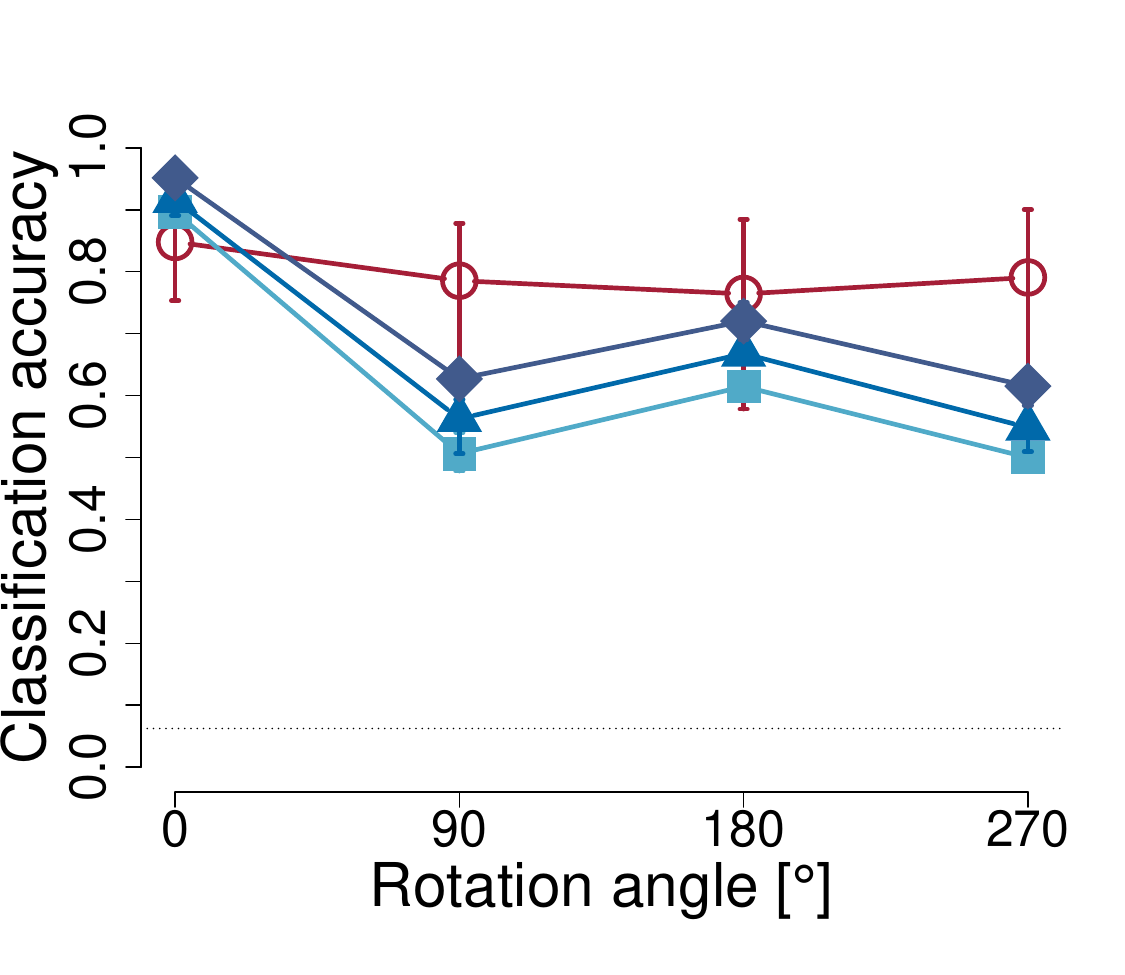}
        \vspace{\captionspace}
        \caption{Rotation}
    \end{subfigure}\hfill
    \begin{subfigure}{\figwidth}
        \centering
        \includegraphics[width=\linewidth]{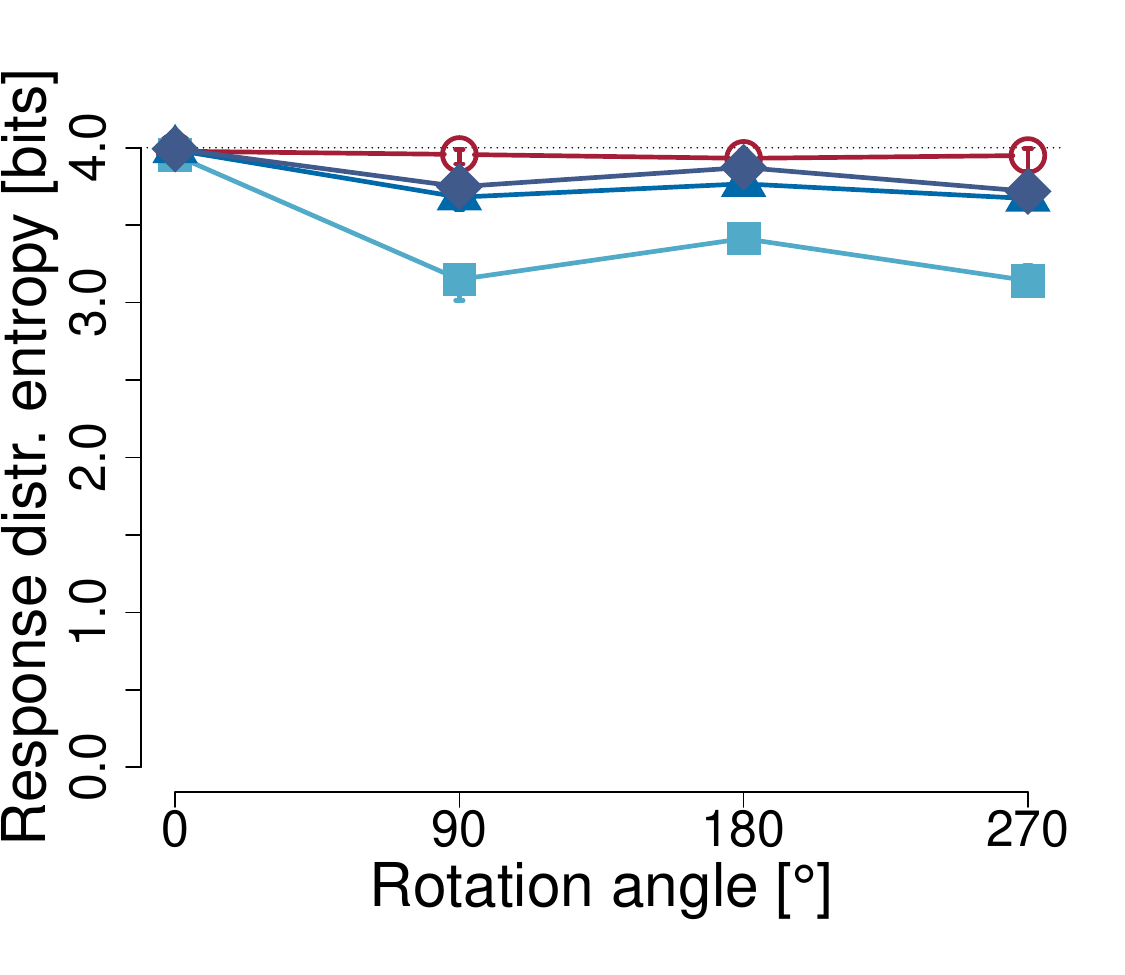}
        \vspace{\captionspace}
        \caption*{}
    \end{subfigure}\hfill
    \caption{Classification accuracy and response distribution entropy for  \textcolor{googlenet.100}{GoogLeNet}, \textcolor{vgg.100}{VGG-19} and \textcolor{resnet.100}{ResNet-152} as well as for \textcolor{human.100}{human observers}. `Entropy' indicates the Shannon entropy of the response/decision distribution (16 classes). It here is a measure of bias towards certain categories: using a test dataset that is balanced with respect to the number of images per category, responding equally frequently with all 16 categories elicits the maximum possible entropy of four bits. If a network or observer responds prefers some categories over others, entropy decreases (down to zero bits in the extreme case of responding with one particular category all the time, irrespective of the ground truth category). Human `error bars' indicate the full range of results across participants. Image manipulations are explained in Section~\ref{meth:image_manipulations} and visualised in Figures~\ref{fig:stimuli_noise_contrast}, \ref{fig:stimuli_lowpass_highpass}, \ref{fig:stimuli_eidolon_I_II}, \ref{fig:stimuli_eidolon_III_phase_scrambling_false_colour_power_equalisation} and \ref{fig:stimuli_salt_and_pepper_noise}.}
    \label{fig:results_accuracy_entropy}
\end{figure}

\subsection{Image manipulations}
\label{meth:image_manipulations}
A total of twelve experiments were performed in a well-controlled psychophysical lab setting. In every experiment, a (possibly parametric) distortion was applied to a large number of images, such that the signal strength ranged from `no distortion / full signal' to `distorted / weak(er) signal'. We then measured how classification accuracy changed as a function of signal strength. Three of the employed image manipulations were dichotomous (colour vs. greyscale, true vs. opponent colour, original vs. equalised power spectrum); one manipulation had four different levels (0, 90, 180 and 270 degrees of rotation); one had seven levels (0, 30, ..., 180 degrees of phase noise) and the other distortions had eight different levels. Those manipulations were: uniform noise, controlled by the `width' parameter indicating the bounds of pixel-wise additive uniform noise; low-pass filtering and high-pass filtering (with different standard deviations of a Gaussian filter); contrast reduction (contrast levels from 100\% to 1\%) as well as three different manipulations from the eidolon toolbox \cite{Koenderink2017}). The three eidolon experiments correspond to different versions of a parametric image manipulation, with the `reach' parameter controlling the strength of the distortion. Additionally, for experiments with training on distortions, we also evaluated performance on stimuli with salt-and-pepper noise (controlled by parameter $p$ indicating probability of setting a pixel to either black or white; $p \in [0, 10, 20, 35, 50, 65, 80, 95]\%$). More information about the different image manipulations is provided in the supplementary material (Section \nameref{methods:image_preprocessing}), where we also show example images across all manipulations and distortion levels (Figures~\ref{fig:stimuli_noise_contrast}, \ref{fig:stimuli_lowpass_highpass}, \ref{fig:stimuli_eidolon_I_II}, \ref{fig:stimuli_eidolon_III_phase_scrambling_false_colour_power_equalisation}, \ref{fig:stimuli_salt_and_pepper_noise}). For a brief overview, Figure~\ref{fig:all_stimuli} depicts one exemplary manipulation per distortion. Overall, the manipulations we used were chosen to reflect a large variety of possible distortions. 

\subsection{Training on distortions}
Beyond evaluating standard pre-trained DNNs on distortions (results reported in Figure~\ref{fig:results_accuracy_entropy}), we also trained networks directly on distortions (Figure~\ref{fig:results_training}). These networks were trained on 16-class-ImageNet, a subset of the standard ImageNet dataset as described in Section~\ref{meth:paradigm_procedure_images}. This reduced the size of the unperturbed training set to approximately one fifth. To correct for the highly imbalanced number of samples per class, we weighted each sample in the loss function with a weight proportional to one over the number of samples of the corresponding class. All networks trained in these experiments had a ResNet-like architecture that differed from a standard ResNet-50 only in the number of output neurons that we reduced from 1000 to 16 to match the 16 entry-level classes of the dataset. Weights were initialised with a truncated normal distribution with zero mean and a standard deviation of \( \frac{1}{\sqrt{n}} \) where \( n \) is the number of output neurons in a layer. While training from scratch, we performed on-the-fly data augmentation using different combinations of the image manipulations. When training a network on multiple types of image manipulations (models B1 to B9 as well as C1 and C2 of Figure~\ref{fig:results_training}), the type of manipulation (including \emph{unperturbed}, i.e. standard colour images if applicable) was drawn uniformly and we only applied one manipulation at a time (i.e., the network never saw a single image perturbed with multiple image manipulations simultaneously, except that some image manipulations did include other manipulations per construction: uniform noise, for example, was always added after conversion to greyscale and contrast reduction to 30\%). For a given image manipulation, the amount of perturbation was drawn uniformly from the levels used during test time (cf. Figure~\ref{fig:results_accuracy_entropy}). The remaining aspects of the training followed standard training procedures for training a ResNet on ImageNet: we used SGD with a momentum of 0.997, a batch size of 64, and an initial learning rate of 0.025. The learning rate was multiplied with 0.1 after 30, 60, 80 and 90 epochs (when training for 100 epochs) or 60, 120, 160 and 180 epochs (when training for 200 epochs). Training was done using TensorFlow 1.6.0 \cite{Abadi2016}. In the training experiments, all manipulations with more than two levels were included except for the eidolon stimuli, since the generation of those stimuli is computationally too slow for ImageNet training. For comparison purposes, we additionally included colour vs. greyscale as well as salt-and-pepper noise (for which there is no human data, but informal comparisons between uniform noise and salt-and-pepper noise strongly suggest that human performance will be similar, see Figure~\ref{fig:introduction_non_iid}).

\section{Generalisation of humans and pre-trained DNNs towards distortions}
\label{results_pre-trained}
In order to assess generalisation performance when the signal gets weaker, we tested twelve different ways of degrading images. These images at various levels of signal strength were then shown to both human observers in a lab and to pre-trained DNNs (ResNet-152, GoogLeNet and VGG-19) for classification. The results of this comparison are visualised in Figure~\ref{fig:results_accuracy_entropy}. While human and DNN performance was similar for comparatively minor colour-related distortions such as conversion to greyscale or opponent colours, we find human observers to be more robust for all of the other distortions: by a small margin for low contrast, power equalisation and phase noise images and by a larger margin for uniform noise, low-pass, high-pass, rotation and all three eidolon experiments. Furthermore, there are strong differences in the error patterns as measured by the response distribution entropy (indicating biases towards certain categories). Human participants' responses were distributed more or less equally amongst the 16 classes, whereas all three DNNs show increasing biases towards certain categories when the signal gets weaker. These biases are not completely explained by the prior class probabilities, and deviate from distortion to distortion. For instance, ResNet-152 almost solely predicts class \texttt{bottle} for images with strong uniform noise (irrespective of the ground truth category),\footnote{A category-level analysis of decision biases for the uniform noise experiment is provided in the supplementary material, Figure~\ref{fig:confusion_noise}.} and classes \texttt{dog} or \texttt{bird} for images distorted by phase noise. One might think of simple tricks to reduce the discrepancy between the response distribution entropy of DNNs and humans. One possible way would be increasing the softmax temperature parameter and assuming that model decisions are sampled from the softmax distribution rather than taking the argmax. However, increasing the response DNN distribution entropy in this way dramatically decreases classification accuracy and thus comes with a trade-off (cf. Figure~\ref{fig:temperature_tradeoff} in the supplementary material).

These results are in line with previous findings reporting human-like processing of chromatic information in DNNs \cite{Flachot2018} but strong decreases in DNN recognition accuracy for image degradations like noise and blur \cite{Vasiljevic2016, Dodge2016, Dodge2017a, Dodge2017b, Zhou2017}. Overall,  DNNs seem to have much more problems generalising to weaker signals than humans, across a wide variety of image distortions. While the human visual system has been exposed to a number of distortions during evolution and lifetime, we clearly had no exposure whatsoever to many of the exact image manipulations that we tested here. Thus, our human data show that a high level of generalisation is, in principle, possible. There may be many different reasons for the discrepancy between human and DNN generalisation performance that we find: Are there limitations in terms of the currently used network architectures (as hypothesised by \cite{Dodge2016}), which may be inferior to the human brain's intricate computations? Is it a problem of the training data (as suggested by e.g. \cite{Zhou2017}), or are today's training methods / optimisers not sufficient to solve robust and general object recognition? In order to shed light on the dissimilarities we found, we performed a second batch of experiments by training networks directly on distorted images.

\section{Training DNNs directly on distorted images}
\label{results_training}

\begin{figure*}
\centering
\includegraphics[width=\linewidth]{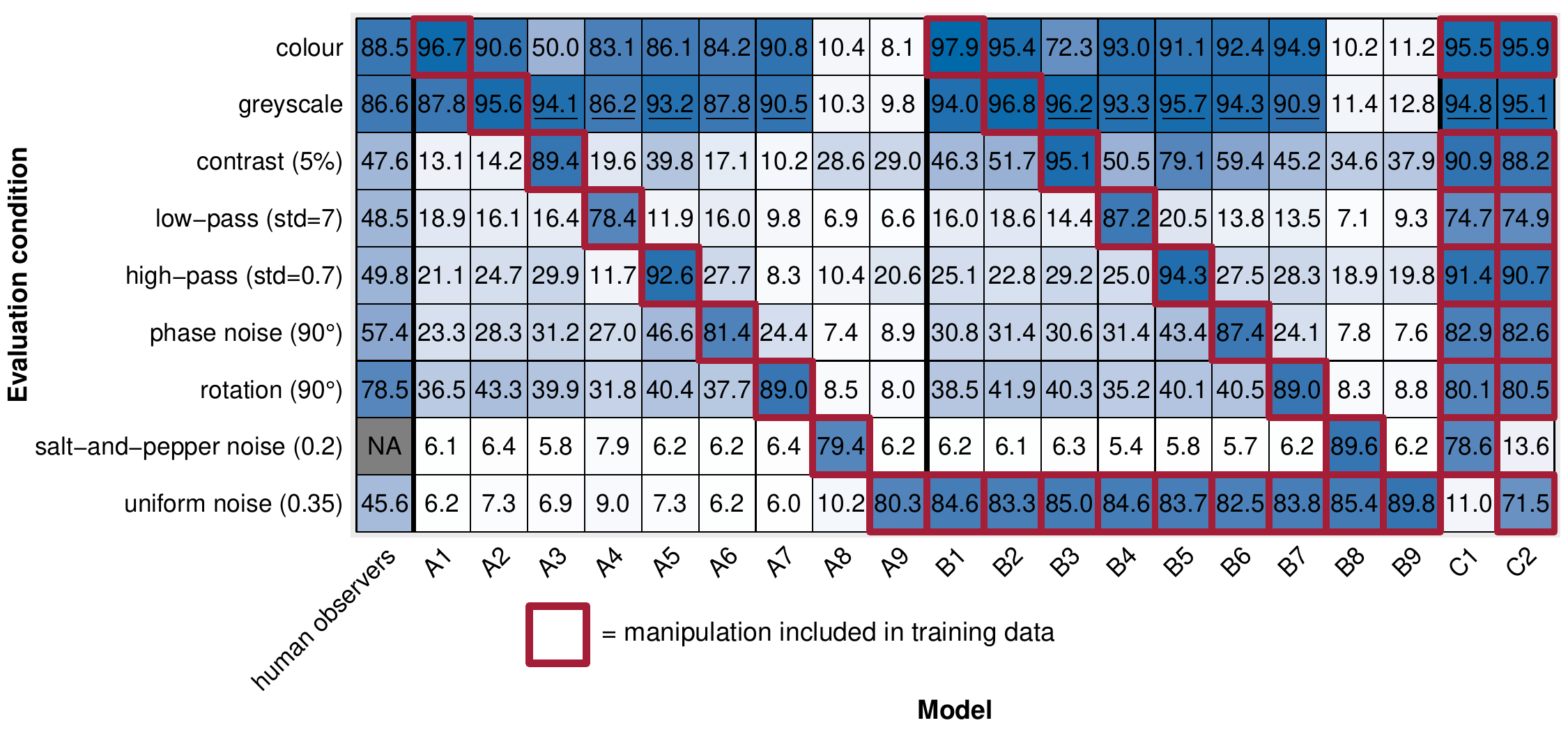}
\caption{Classification accuracy (in percent) for networks with potentially distorted training data. Rows show different test conditions at an intermediate difficulty (exact condition indicated in brackets, units as in Figure~\ref{fig:results_accuracy_entropy}). Columns correspond to differently trained networks (leftmost column: human observers for comparison; no human data available for salt-and-pepper noise). All of the networks were trained from scratch on (a potentially manipulated version of) 16-class-ImageNet. Manipulations included in the training data are indicated by a \textcolor{human.100}{red} rectangle; additionally `greyscale' is underlined if it was part of the training data because a certain distortion encompasses greyscale images at full contrast. Models~\textbf{A1~to~A9}: ResNet-50 trained on a single distortion (100 epochs). Models~\textbf{B1~to~B9}: ResNet-50 trained on uniform noise plus one other distortion (200 epochs). Models~\textbf{C1~\&~C2}: ResNet-50 trained on all but one distortion (200 epochs). Chance performance is at $\frac{1}{16}=6.25\%$ accuracy.}
\label{fig:results_training}
\end{figure*}

We trained one network per distortion directly and from scratch on (potentially manipulated) 16-class-ImageNet images. The results of this training are visualised in Figure~\ref{fig:results_training} (models A1 to A9). We find that these specialised networks consistently outperformed human observers, by a large margin, on the image manipulation they were trained on (as indicated by strong network performance on the diagonal).
This is a strong indication that currently employed architectures (such as ResNet-50) and training methods (standard optimiser and training procedure) are sufficient to `solve' distortions under i.i.d. train/test conditions. We were able to not only close the human-DNN performance gap that was observed by \cite{Dodge2017a} (who fine-tuned networks on distortions, reporting improved but not human-level DNN performance) but to surpass human performance in this respect. While the human visual system certainly has a much more complicated structure \cite{Kietzmann2017}, this does not seem to be necessary to deal with even strong image manipulations of the type employed here.

However, as noted earlier, robust generalisation is primarily not about solving a specific problem known exactly in advance. We therefore tested how networks trained on a certain distortion type perform when tested on other distortions. These results are visualised in Figure~\ref{fig:results_training} by the off-diagonal cells of models A1 to A9. Overall, we find that training on a certain distortion slightly improves performance on other distortions in a few instances, but is detrimental in other cases (when compared to a vanilla ResNet-50 trained on colour images, model A1 in the figure).\footnote{The no free lunch theorem \cite{Wolpert1997} states that better performance on some input is necessarily accompanied by worse performance on other input; however we here are only interested in a narrow subset of the possible input space (natural images corrupted by distortions). The high accuracies of human observers across distortions indicate that it is, in principle, possible to achieve good performance on many distortions simultaneously.} Performance on salt-and-pepper noise as well as uniform noise was close to chance level for all networks, even for a network trained directly on the respectively other noise model. This may be surprising given that these two types of noise do not seem very different to a human eye (as indicated in Figure~\ref{fig:introduction_non_iid}). Hence, training a network on one distortion does not generally lead to improvements on other distortions.

Since training on a single distortion alone does not seem to be sufficient to evoke robust generalisation performance in DNNs, we also trained the same architecture (ResNet-50) on two additional settings. Models B1 to B9 in Figure~\ref{fig:results_training} show performance for training on one particular distortion in combination with uniform noise (training consisted of 50\% images from each manipulation). Uniform noise was chosen since it seemed to be one of the hardest distortions for all networks, and hence they might benefit from including this particular distortion in the training data. Furthermore, we trained models C1 and C2 on all but one distortion (either uniform or salt-and-pepper noise was left out).

We find that object recognition performance of models B1 to B9 is improved compared to models A1 to A9, both on the distortions they were actually trained on (diagonal entries with red rectangles in Figure~\ref{fig:results_training}) as well as on a few of the distortions that were not part of the training data. However, this improvement may be largely due to the fact that models B1 to B9 were trained on 200 epochs instead of 100 epochs as for models A1 to A9, since the accuracy of model B9 (trained \& tested on uniform noise, 200 epochs) also shows an improvement towards model A9 (trained \& tested on uniform noise, 100 epochs). Hence, in the presence of heavy distortions, training longer may go a long way but incorporating other distortions in the training does not seem to be generally beneficial to model performance. Furthermore, we find that it is possible even for a single model to reach high accuracies on all of the eight distortions it was trained on (models C1 \& C2), however for both left-out uniform and salt-and-pepper noise, object recognition accuracy stayed around 11 to 14\%, which is by far closer to chance level (approx. 6\%) than to the accuracy reached by a specialised network trained on this exact distortion (above 70\%, serving as a lower bound on the achievable performance).

Taken together, these findings indicate that data augmentation with distortions alone may be insufficient to overcome the generalisation problem that we find. It may be necessary to move from asking ``why are DNNs generalising so well (under i.i.d. settings)?'' \cite{Zhang2016} to ``why are DNNs generalising so poorly (under non-i.i.d. settings)?''. It is up to future investigations to determine how DNNs that are currently being handled as computational models of human object recognition can solve this challenge. At the exciting interface between cognitive science / visual perception and deep learning, inspiration and ideas may come from both fields: While the computer vision sub-area of domain adaptation (see \cite{Patel2015} for a review) is working on robust machine inference in spite of shifts in the input distribution, the human vision community is accumulating strong evidence for the benefits of local gain control mechanisms. These normalisation processes seem to be crucial for many aspects of robust animal and human vision \cite{Carandini2012}, are predictive for human vision data \cite{Berardino2017, Schuett2017} and have proven useful in the context of computer vision \cite{Jarrett2009, Ren2016}. It could be an interesting avenue for future research to determine whether there is a connection between neural normalisation processes and DNN generalisation performance. Furthermore, incorporating a shape bias in DNNs seems to be a very promising avenue towards general noise robustness, strongly improving performance on many distortions \cite{Geirhos2018}.

\section{Conclusion}
  \label{conclusion}
We conducted a behavioural comparison of human and DNN object recognition robustness against twelve different image distortions. In comparison to human observers, we find the classification performance of three well-known DNNs trained on ImageNet---ResNet-152, GoogLeNet and VGG-19---to decline rapidly with decreasing signal-to-noise ratio under image distortions. Additionally, we find progressively diverging patterns of classification errors between humans and DNNs with weaker signals.Our results, based on 82,880 psychophysical trials under well-controlled lab conditions, demonstrate that there are still marked differences in the way humans and current DNNs process object information. These differences, in our setting, cannot be overcome by training on distorted images (i.e., data augmentation): While DNNs cope perfectly well with the exact distortion they were trained on, they still show a strong generalisation failure towards previously unseen distortions. Since the space of possible distortions is literally unlimited (both theoretically and in real-world applications), it is not feasible to train on all of them. DNNs have a generalisation problem when it comes to settings that go beyond the usual (yet often unrealistic) i.i.d. assumption. We believe that solving this generalisation problem will be crucial both for robust machine inference and towards better models of human object recognition, and we envision that our findings as well as our carefully measured and freely available behavioural data\footnote{\url{https://github.com/rgeirhos/generalisation-humans-DNNs}} may provide a new useful benchmark for improving DNN robustness and a motivation for neuroscientists to identify mechanisms in the brain that may be responsible for this remarkable robustness.

\subsubsection*{Author contributions}
The initial project idea of comparing humans against DNNs was developed by F.A.W. and R.G. All authors jointly contributed towards designing the study and interpreting the data. R.G. and C.R.M.T. developed the image manipulations and acquired the behavioural data with input from H.H.S. and F.A.W.; J.R. trained networks on distortions and derived the optimal aggregation method; experimental data and networks were evaluated by C.R.M.T., R.G. and J.R. with input from H.H.S, M.B. and F.A.W.; R.G. and C.R.M.T. worked on making our work reproducible (data, code and materials openly accessible; writing supplementary material); R.G. wrote the paper with significant input from all other authors.

\subsubsection*{Acknowledgments}

This work has been funded, in part, by the German Federal Ministry of Education and Research (BMBF) through the Bernstein Computational Neuroscience Program T\"ubingen (FKZ: 01GQ1002) as well as the German Research Foundation (DFG; Sachbeihilfe Wi 2103/4-1 and SFB 1233 on ``Robust Vision''). The authors thank the International Max Planck Research School for Intelligent Systems (IMPRS-IS) for supporting R.G. and J.R.; J.R. acknowledges support by the Bosch Forschungsstiftung (Stifterverband, T113/30057/17); M.B. acknowledges support by the Centre for Integrative Neuroscience T\"ubingen (EXC 307) and by the Intelligence Advanced Research Projects Activity (IARPA) via Department of Interior/Interior Business Center (DoI/IBC) contract number D16PC00003.

We would like to thank David Janssen for his invaluable contributions in shaping the early stage of this project. Furthermore, we are very grateful to Tom Wallis for providing the MATLAB source code of one of his experiments, and for allowing us to use and modify it; Silke Gramer for administrative and Uli Wannek for technical support, as well as Britta Lewke for the method of creating response icons and Patricia Rubisch for help with testing human observers. Moreover, we would like to thank Nikolaus Kriegeskorte, Jakob Macke and Tom Wallis for helpful feedback, and three anonymous reviewers for constructive suggestions.

\medskip
\small
\bibliographystyle{unsrtnat}
\bibliography{refs.bib}

\newpage
\section*{Supplementary material}
While the main aspects of employed paradigm, procedure, observers and DNNs were already mentioned earlier, this section aims at providing exhaustive and reproducible experimental details. Furthermore, Figure~\ref{fig:uncertainty} examines how network uncertainty develops as a function of signal strength, and Figure~\ref{fig:training_accuracy_entropy} shows the classification accuracy of networks trained on distortions across all conditions.
All data, if not stated otherwise, were analysed using R version 3.2.3 \cite{RCoreteam}.

\subsection*{Paradigm \& procedure}
\begin{figure}
\centering
\includegraphics[width=0.9\linewidth]{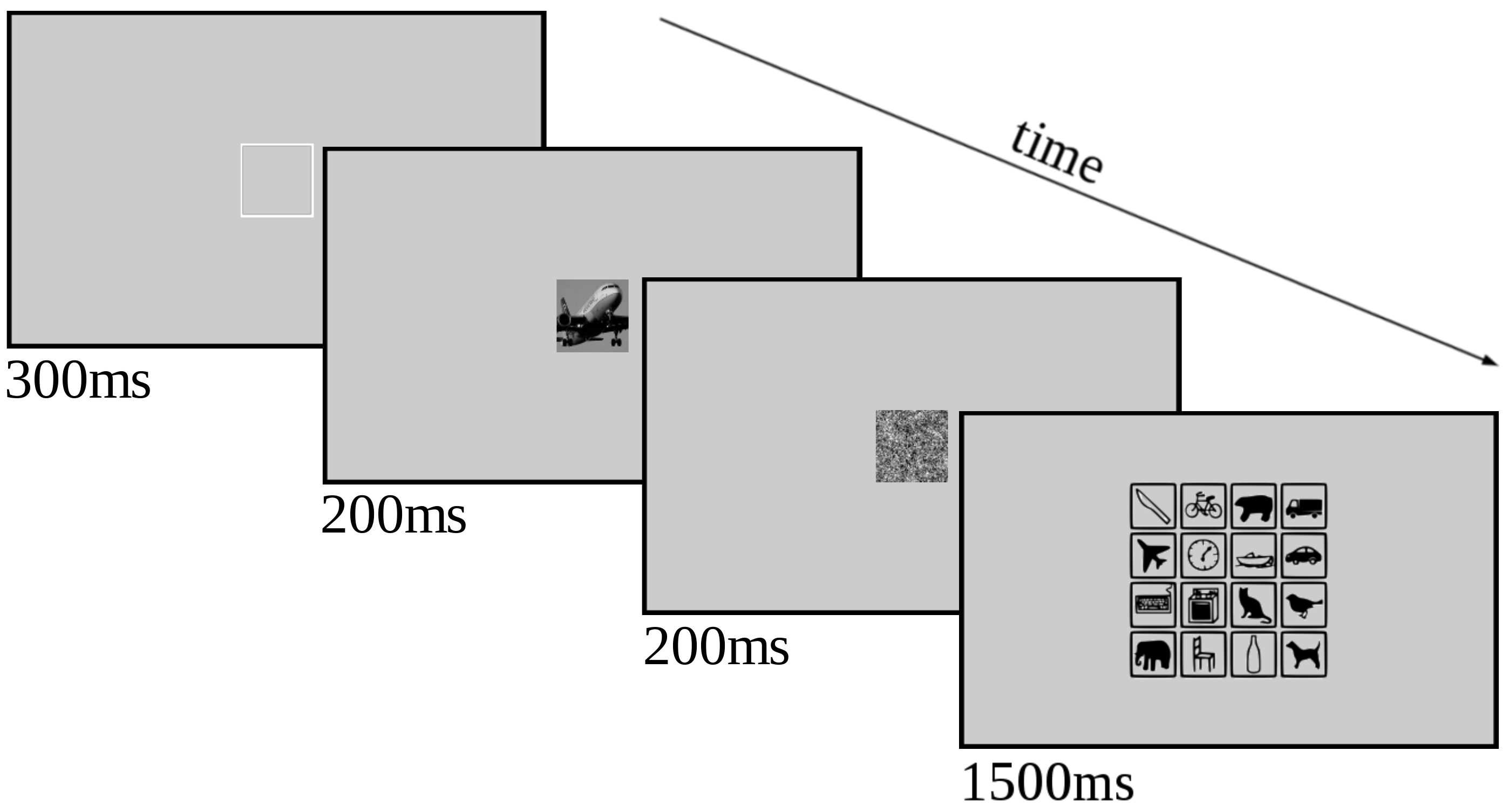}
\caption{Schematic of a trial. After the presentation of a central fixation square (300 ms), the image was visible for 200 ms, followed immediately by a noise mask with 1/\textit{f} spectrum (200 ms). Then, a response screen appeared for 1500 ms, during which the observer clicked on a category. Note that we increased the contrast of the noise mask in this figure for better visibility when printed.
Categories row-wise from top to bottom: \texttt{knife, bicycle, bear, truck, airplane, clock, boat, car, keyboard, oven, cat, bird, elephant, chair, bottle, dog}. The icons are a modified version of the ones from the MS COCO website (\url{http://mscoco.org/explore/}).}
\label{fig:typical_trial}
\end{figure}

A schematic of a typical trial is shown in Figure~\ref{fig:typical_trial}. Prior to starting the experiment, all participants were shown the response screen and asked to name all categories to ensure that the task was fully clear. They were instructed to click on the category that they thought resembles the image best, and to guess if they were unsure. They were allowed to change their choice within the 1500 ms response interval; the last click on a category icon of the response screen was counted as the answer. The experiment was not self-paced, i.e. the response screen was always visible for 1500 ms and thus, each experimental trial lasted exactly 2200 ms (300 ms + 200 ms + 200 ms + 1500 ms). During the whole experiment, the screen background was set to a grey value of 0.454 in the [0, 1] range, corresponding to the mean greyscale value of all images in the dataset (41.17 cd/m\textsuperscript{2}).

On separate days we conducted twelve different experiments. The number of trials per experiment is reported in Table \ref{tab:TrialN}. For each experiment, we randomly chose between 70 and 80 images per category from the pool of images without replacement (i.e., no observer ever saw an image more than once throughout the entire experiment). Within each category, all conditions were counterbalanced. Random stimulus selection was done individually for each participant to reduce the influence of any accidental bias in the image selection. Images within the experiments were presented in randomised order. After 256 trials (colour, uniform noise and eidolon experiments), 128 trials (contrast experiment) and 160 trials (remaining experiments), the mean performance of the last block was displayed on the screen, and observers were free to take a short break. Ahead of each experiment, all observers conducted approximately 10 minutes of practice trials to gain familiarity with the task and the position of the categories on the response screen. Trials in which human observers failed to click on any category were recorded as an incorrect answer in the data analysis, and are shown as a separate category (top row) in the confusion matrices (DNNs, obviously, never fail to respond). Such a failure to respond occurred, on average, in only 1.91\% of trials per experiment---one of the advantages of controlled laboratory studies (\emph{SD} = 0.69\%).

\subsection*{Apparatus}
\label{methods:apparatus}
All stimuli were presented on a VIEWPixx LCD monitor (VPixx Technologies, Saint-Bruno, Canada) in a dark chamber. The 22'' monitor ($ 484\times 302$ mm) had a spatial resolution of  $1920 \times 1200$  pixels at a refresh rate of 120 Hz. Stimuli were presented at the center of the screen with $256 \times 256$ pixels, corresponding, at a viewing distance of 123 cm, to $3 \times 3$ degrees of visual angle. A chin rest was used in order to keep the position of the head constant over the course of an experiment. Stimulus presentation and response recording were controlled using MATLAB (Release 2016a, The MathWorks, Inc., Natick, Massachusetts, U.S.) and the Psychophysics Toolbox extensions version 3.0.12 \cite{Brainard1997, Kleiner2007} along with the iShow library (\url{http://dx.doi.org/10.5281/zenodo.34217}) on a desktop computer (12 core CPU i7-3930K, AMD HD7970 graphics card “Tahiti” by AMD, Sunnyvale, California, United States) running Kubuntu 14.04 LTS. Responses were collected with a standard computer mouse.

\subsection*{Observers \& pre-trained networks}
Three observers participated in the colour experiment (all male; 22 to 28 years; mean: 25 years) and in the contrast experiment. Six observers participated in the opponent colour, high-pass filter, low-pass filter, phase noise and power equalisation experiments (three female, three male; 20 to 25 years; mean: 22 years). In the other two experiments, five observers took part (uniform noise experiment: one female, four male; 20 to 28 years; mean: 23 years; eidolon experiments: three female, two male; 19 to 28 years; mean: 22 years). Subject-01 is an author and participated in all but the eidolon experiments. All other participants were either paid 10 Euros per hour for their participation or gained course credit. All observers were students and reported normal or corrected-to-normal vision.

We used GoogLeNet \cite{Szegedy2015}, VGG-19 \cite{Simonyan2015} and ResNet-152 \cite{He2016} for our analyses. For all three networks, we used the pre-trained implementations as provided by the TensorFlow-Slim framework (\url{https://github.com/tensorflow/models/tree/master/research/slim} cloned on May 2, 2017) and programmed in the TensorFlow library for machine learning \cite{Abadi2016}. The individual pre-trained weights were also downloaded from the latter GitHub repository. We validated that our installation reproduced the classification accuracies provided on the website. The networks' input were $224 \times 224$ pixel RGB images. For greyscale images, we set all three channels to be equal to the greyscale image's single channel. Images were fed through the networks using a single feedforward pass.

\begin{table}
  \caption{Numbers of trials in the respective experiments. C. = conditions; P. = practice trials \& blocks; M.= main experiment trials and blocks. The per condition column reports the number of trials per category and distortion level. The duration is reported without breaks.}
  \label{tab:TrialN}
  \centering
  \begin{tabular}{llllllll}
    \toprule
    Distortion type & C. & P. blocks & P. total & M. blocks & M. total & Per C. & Duration\\
    \midrule
    Colour              & 2  & 2 & 320  & 5 & 1280 & 40 & 47 min\\
    Uniform noise       & 8  & 2 & 256  & 5 & 1280 & 10 & 47 min\\
    Contrast            & 8  & 2 & 256  & 10 & 1280 & 10 & 47 min\\
    Eidolon I           & 8  & 4 & 384  & 5 & 1280 & 10 & 47 min\\
    Eidolon II          & 8  & 4 & 384  & 5 & 1280 & 10 & 47 min\\
    Eidolon III         & 8  & 4 & 384  & 5 & 1280 & 10 & 47 min\\
    Opponent colours    & 2  & 2 & 224  & 7 & 1120 & 35 & 41 min\\
    Low-pass filtering  & 8  & 2 & 256  & 8 & 1280 & 10 & 47 min\\
    High-pass filtering & 8  & 2 & 256  & 8 & 1280 & 10 & 47 min\\
    Phase noise         & 7  & 2 & 224  & 7 & 1120 & 10 & 41 min\\
    Power-equalisation  & 2  & 2 & 224  & 7 & 1120 & 35 & 41 min\\ 
    Rotation            & 4  & 2 & 256  & 8 & 1280 & 20 & 47 min\\
    \bottomrule
  \end{tabular}
\end{table}

\subsection*{Categories and image database}
\label{methods:categories}
The images serving as psychophysical stimuli were images extracted from the training set of the ImageNet Large Scale Visual Recognition Challenge 2012 database \cite{Russakovsky2015}. This database contains millions of labeled images grouped into 1,000 very fine-grained categories (e.g., over a hundred different dog breeds). If human observers are asked to name objects, however, they most naturally categorise them into so-called basic or entry-level categories, e.g. \texttt{dog} rather than \texttt{German shepherd} \cite{Rosch1999}. The Microsoft COCO (MS COCO) database \cite{Lin2015} is an image database structured according to 91 such entry-level categories, making it an excellent source of categories for an object recognition task. Thus for our experiments we fused the carefully selected entry-level categories in the MS COCO database with the large quantity of images in ImageNet. Using WordNet's \emph{hypernym} relationship (\emph{x} is a hypernym of \emph{y} if \emph{y} is a ``kind of'' \emph{x}, e.g., \texttt{dog} is a hypernym of \texttt{German shepherd}), we mapped every ImageNet label to an entry-level category of MS COCO in case such a relationship exists, retaining 16 clearly non-ambiguous categories with sufficiently many images within each category (see Figure~\ref{fig:typical_trial} for a iconic representation of the 16 categories). A complete list of ImageNet labels used for the experiments can be found in our online repository.\footnote{\url{https://github.com/rgeirhos/generalisation-humans-DNNs}} Since all investigated DNNs, when shown an image, output classification predictions for all 1,000 ImageNet categories, we disregarded all predictions for categories that were not mapped to any of the 16 entry-level categories. For each of those 16 categories we summed over the predictions of all ImageNet categories mapping to that particular entry-level category. Then the entry-level category with the highest summed prediction was selected as the network's response. This way, the DNN response selection corresponds directly to the forced-choice paradigm for our human observers.

\subsection*{Image preprocessing and distortions}
\label{methods:image_preprocessing}
We used Python for all image preprocessing (Version 2.7.11) and for running experiments through pre-trained networks (Version 3.5). From the pool of ImageNet images of the 16 entry-level categories, we excluded all greyscale images (1\%) as well as all images not at least $ 256 \times 256 $ pixels in size (11\% of non-greyscale images). We then cropped all images to a center patch of $ 256 \times 256 $ pixels as follows: First, every image was cropped to the largest possible center square. This center square was then downsampled to the desired size with \texttt{PIL.Image.thumbnail((256, 256), Image.ANTIALIAS)}. Human observers get adapted to the mean luminance of the display during experiments, and thus images which are either very bright or very dark may be harder to recognise due to their very different perceived brightness. We therefore excluded all images which had a mean deviating more than two standard deviations from that of other images (5\% of correct-sized colour-images excluded). In total we retained 213,555 images from ImageNet.

\begin{figure*}[h]
    \begin{subfigure}{0.5\textwidth}
        \includegraphics[width=\linewidth]{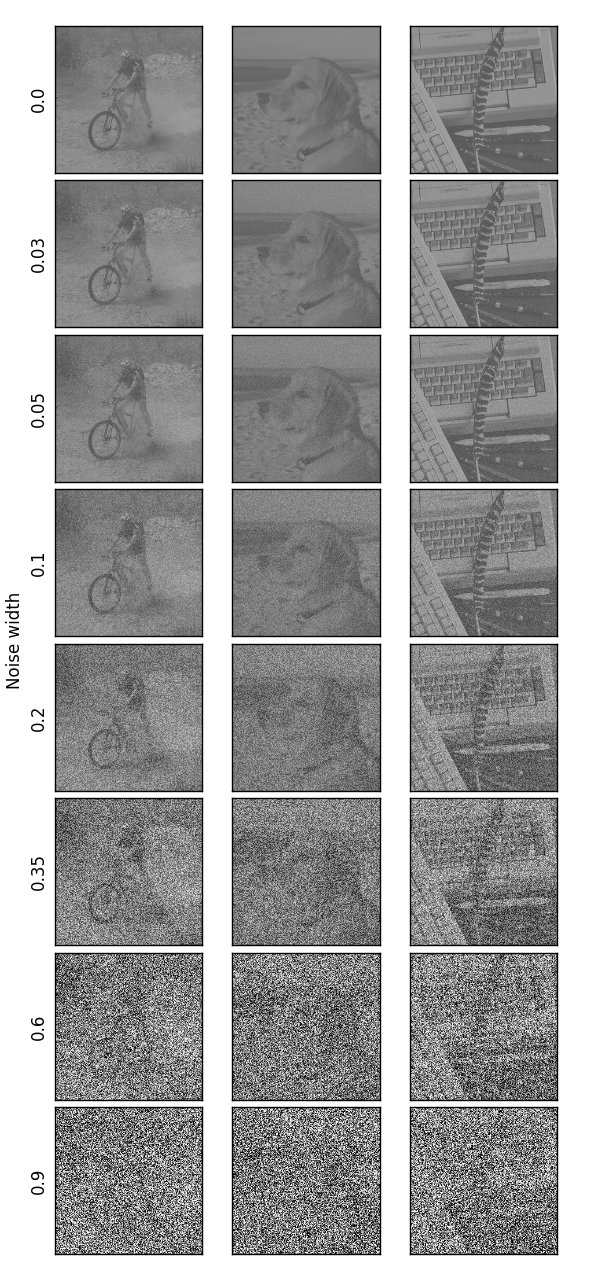}
        \vspace{\captionspace}
        \caption{Uniform noise}
    \end{subfigure}\hfill
    \begin{subfigure}{0.5\textwidth}
        \includegraphics[width=\linewidth]{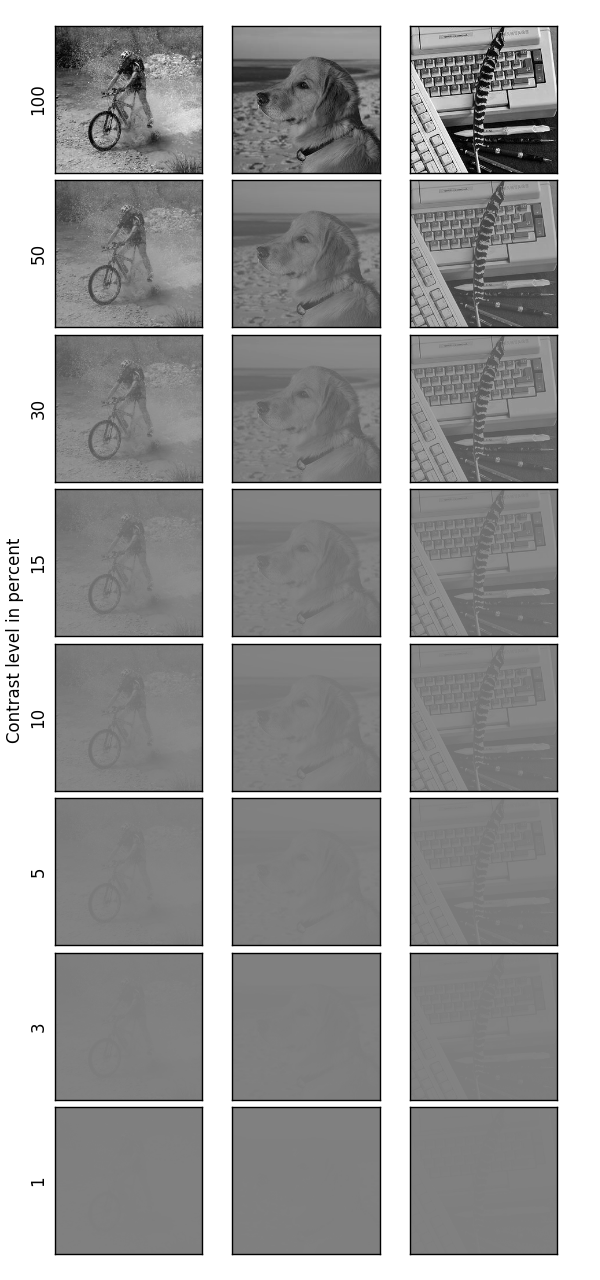}
        \vspace{\captionspace}
        \caption{Contrast}
    \end{subfigure}\hfill
\caption{Three example stimuli for different conditions of uniform noise and contrast experiments. The three images (categories \texttt{bicycle}, \texttt{dog} and \texttt{keyboard}) were drawn randomly from the pool of images used in the experiments. Best viewed on screen.}
\label{fig:stimuli_noise_contrast}
\end{figure*}

\begin{figure*}
    \begin{subfigure}{0.5\textwidth}
        \includegraphics[width=\linewidth]{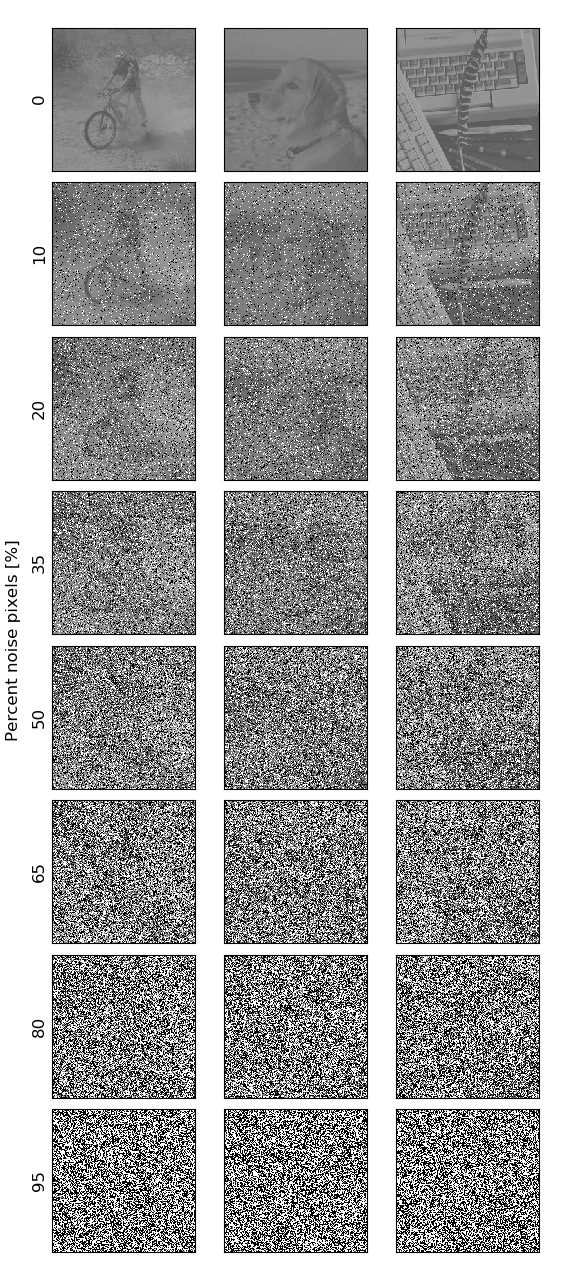}
        \vspace{\captionspace}
        \caption{Salt and pepper noise}
    \end{subfigure}
\caption{Three example stimuli for different conditions of salt and pepper noise. The three images (categories \texttt{bicycle}, \texttt{dog} and \texttt{keyboard}) were drawn randomly from the pool of images used in the experiments. Best viewed on screen. Salt and pepper noise was used in DNN training experiments with the conditions depicted in the figure above.}
\label{fig:stimuli_salt_and_pepper_noise}
\end{figure*}

For the experiments using greyscale images the stimuli were converted using the \texttt{rgb2gray} method \cite{VanderWalt2014} in Python. This was the case for all experiments and conditions except for the `colour' condition of the \textbf{colour experiment}, as well as for the opponent colour experiment. For the \textbf{contrast experiment}, we employed eight different contrast levels $ c \in \{1, 3, 5, 10, 15, 30, 50, 100\%\}$. For an image in the [0, 1] range, scaling the image to a new contrast level $ c $ was achieved by computing $$ new\_value = \frac{c}{100\%} \cdot original\_value + \frac{1-\frac{c}{100\%}}{2} $$ for each pixel. For the \textbf{uniform noise experiment}, we first scaled all images to a contrast level of $ c=30\% $. Subsequently, white uniform noise of range $ [-w, w] $ was added pixelwise, $ w \in \{0.0, 0.03, 0.05, 0.1, 0.2, 0.35, 0.6, 0.9\} $. In case this resulted in a value out of the [0, 1] range, this value was clipped to either 0 or 1. By design, this never occurred for a noise range less or equal to 0.35 due to the reduced contrast (see above). For $ w = 0.6 $, clipping occurred in $ 17.2 \% $ of all pixels and for $ w = 0.9 $ in $ 44.4 \% $ of all pixels. See Figure~\ref{fig:stimuli_noise_contrast} for example stimuli. For the \textbf{salt and pepper noise experiment}, used in DNN training experiments, we also scaled the greyscale image to a contrast level of 30\% prior to adding noise in order to ensure maximal comparability with the uniform noise experiment. Salt and pepper noise, i.e. setting pixels to either black or white, was drawn pixelwise with a certain probability $p$, $p\in \{0, 10, 20, 35, 50, 65, 80, 95\}\%$. See Figure~\ref{fig:stimuli_salt_and_pepper_noise} for example salt-and-pepper stimuli at all conditions.

For the \textbf{opponent colours experiment}, our aim was to produce images that would be perceived by human observers as having exactly the opposite colours of the original, while retaining the same luminance. Therefore, we converted images to a colour space in which we could invert the colours without affecting luminance values. One such colour space is the Derrington-Krauskopf-Lennie (DKL) colour space \citep{Derrington1984}. In order to account for the nonlinearity of our experimental display monitor, we measured the emitted luminance for RGB grey values between 0 and 255. From this we built a lookup table from RGB grey values to actual emitted luminance values $f_{monitor}$. To evaluate how much the human retina's long-, middle-, and short-wave receptors would be excited by the colours presented on the monitor, we measured the intensity of all emitted wave lengths between 390-780 nm for the RGB values (255 0 0), (0 255 0), (0 0 255), respectively. We then multiplied the respective emitted spectra between 390-780 nm with the corresponding cone sensitivities taken from the 2-deg LMS fundamentals proposed by \cite{Stockman2000} and summed over them. This resulted in a matrix $C$ from RGB to cone activities (LMS space). Then we calculated a conversion matrix $D$ of cone activities into the DKL colour space  following the conversion example in \cite{Brainard2002}. An image was, consequently, converted from RGB to DKL by applying $f_{monitor}$ to it and subsequent multiplication with first $C$ and then $D$. The DKL space has three channels reminiscent of the opponent colour process of the human visual system \citep{Brainard2002}. They are $DKL_{lum}$, a luminance channel, $DKL_{L-M}$, a channel representing the difference between long- and middle-wave receptor activation, as well as $DKL_{S-lum}$, a channel representing the difference between the activation of the short-wave receptor and the luminance. Since we wanted to keep the luminance unchanged, we multiplied the $DKL_{L-M}$ and $DKL_{S-lum}$ channels with the value `-1'. Subsequently, we converted the manipulated images back to RGB using the inverse matrices of $D$ and $C$ and then applied the inverse of $f_{monitor}$ to them. All resulting pixel values outside the range [0, 1] were clipped to 0 or 1. This only happened for 0.34\% of pixels with a mean clipped away value of 0.004. This corresponds to the minimal colour intensity step as $0.004 \approx \frac{1}{255}$. 

\begin{figure*}
    \begin{subfigure}{0.5\textwidth}
        \includegraphics[width=\linewidth]{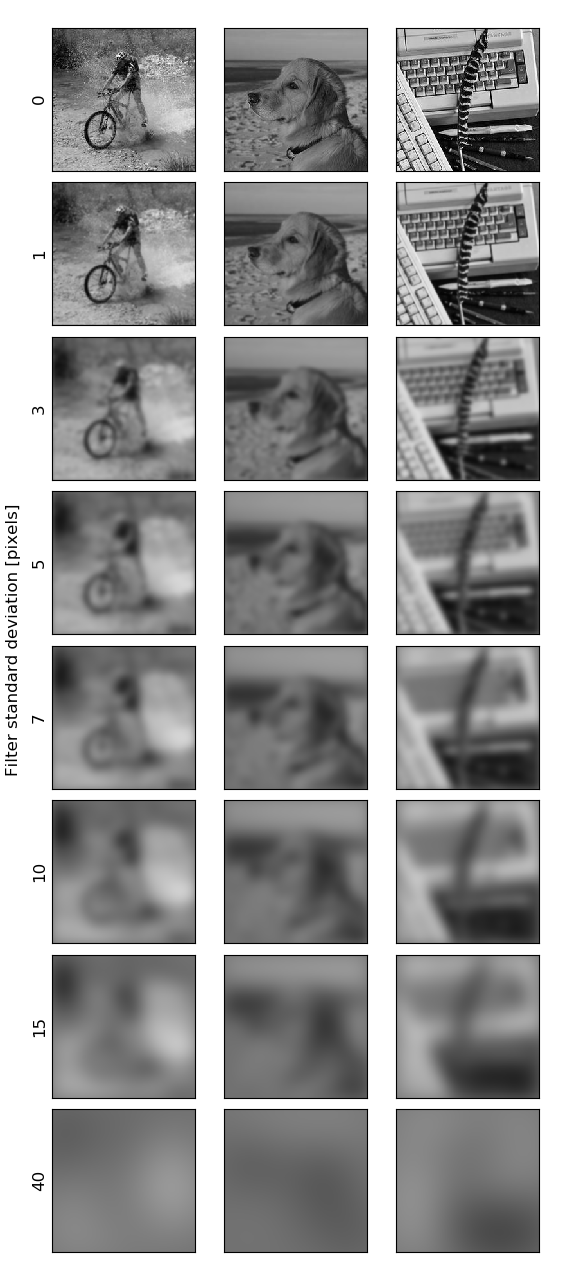}
        \vspace{\captionspace}
        \caption{Low-pass}
    \end{subfigure}\hfill
    \begin{subfigure}{0.5\textwidth}
        \includegraphics[width=\linewidth]{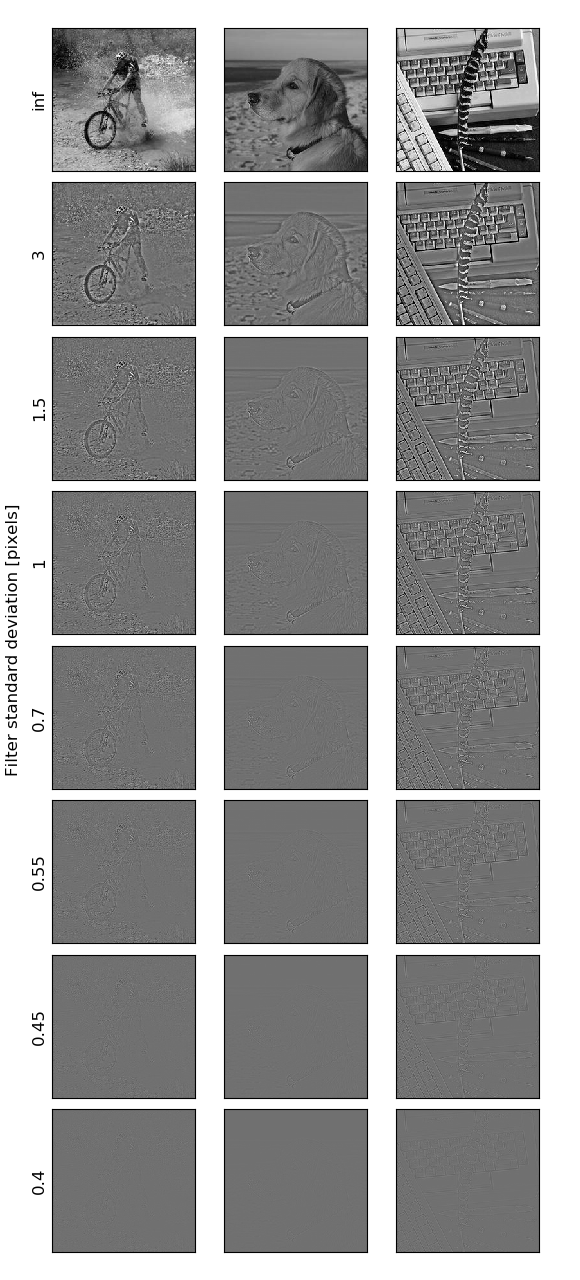}
        \vspace{\captionspace}
        \caption{High-pass}
    \end{subfigure}\hfill
\caption{Three example stimuli for different conditions of low-pass and high-pass experiments. The three images (categories \texttt{bicycle}, \texttt{dog} and \texttt{keyboard}) were drawn randomly from the pool of images used in the experiments. Best viewed on screen.}
\label{fig:stimuli_lowpass_highpass}
\end{figure*}

For the low-pass and high-pass experiments we used the \texttt{scipy.ndimage.filters.gaussian\_filter()} function. The \textbf{low-pass experiment}'s eight conditions differed in the standard deviation of the Gaussian filter. Standard deviations were 0 (original image), 1, 3, 7, 10, 15 and 40 pixels (Figure \ref{fig:stimuli_lowpass_highpass}. We used constant padding with the mean pixel value over the testing images (0.4423) and truncation at four standard deviations.
The \textbf{high-pass experiment} also had eight conditions. Standard deviations were 0.4, 0.45, 0.55, 0.7, 1, 1.5, 3 pixels and inf (original image) (Figure \ref{fig:stimuli_lowpass_highpass}). The high-pass filtered images were produced by subtracting a low-pass filtered image as described above from the original image. However, many of the high-pass filtered images' pixels fell outside the [0, 1] range. To resolve this, we calculated the difference between the mean pixel value over all test images (0.4423) and the mean pixel value of the high-pass filtered image. That difference was added back to the image. This had the effect that images approached a uniform mean grey image of value 0.4423 for low standard deviations. 
For both experiments pixel values were clipped to the [0, 1] range, if lying outside after the filtering. This only happened for \textless 0.001\% of pixels with a mean clipped away value of \textless 0.001 for the both filtering experiments.

\begin{figure*}
    \begin{subfigure}{0.5\textwidth}
        \includegraphics[width=\linewidth]{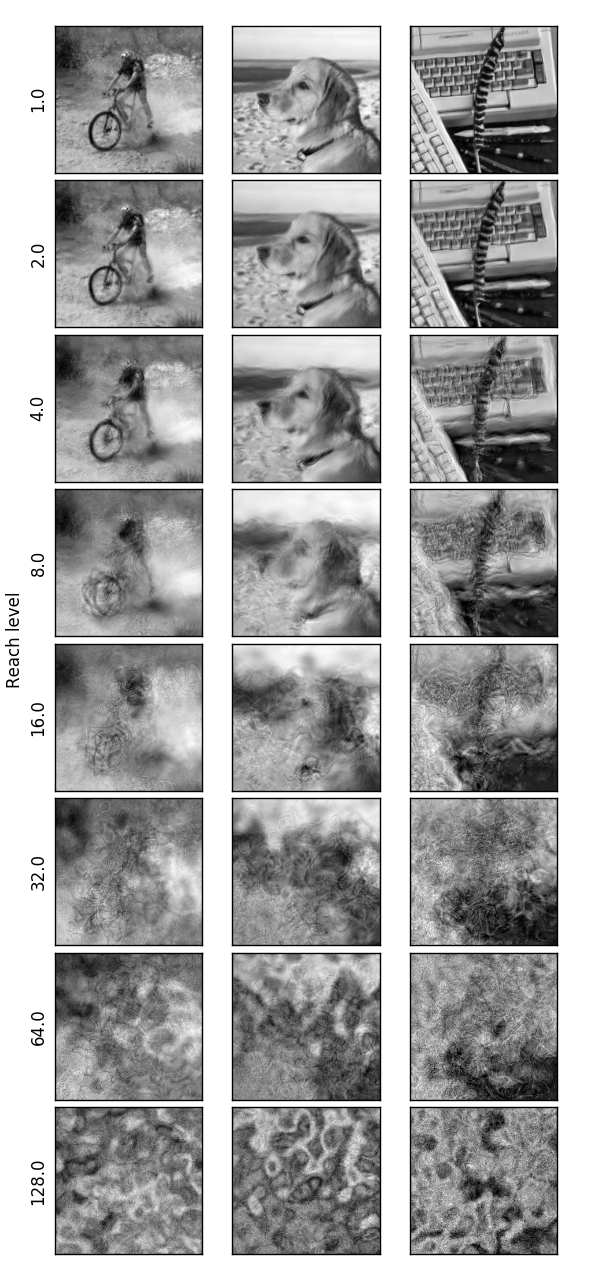}
        \vspace{\captionspace}
        \caption{Eidolon III}
        \label{fig:stimuli_eidolon_III}
    \end{subfigure}\hfill
    \begin{subfigure}{0.5\textwidth}
        \includegraphics[width=\linewidth]{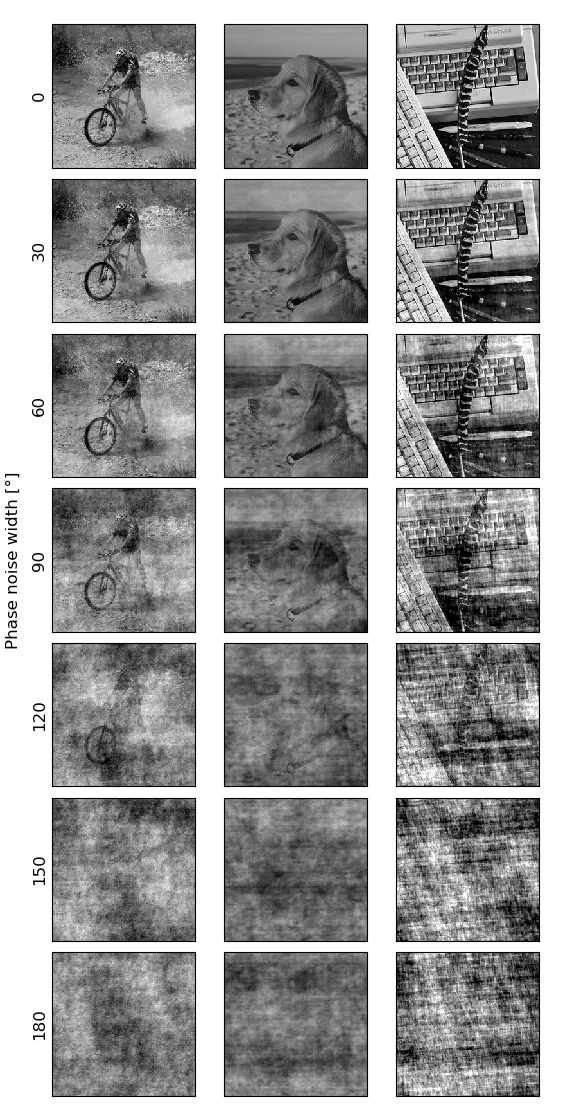}
        \vspace{\captionspace}
        \caption{Phase noise}
    \end{subfigure}\hfill
    
    \begin{subfigure}{0.5\textwidth}
        \includegraphics[width=\linewidth]{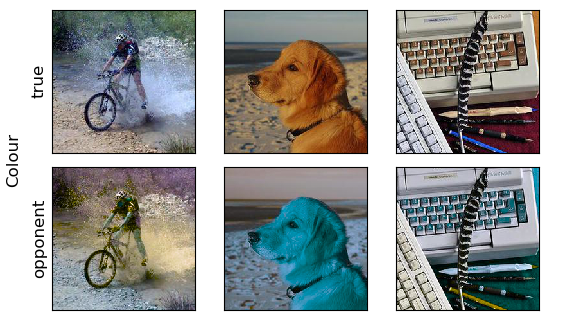}
        \caption{False colour}
    \end{subfigure}\hfill
    \begin{subfigure}{0.5\textwidth}
        \includegraphics[width=\linewidth]{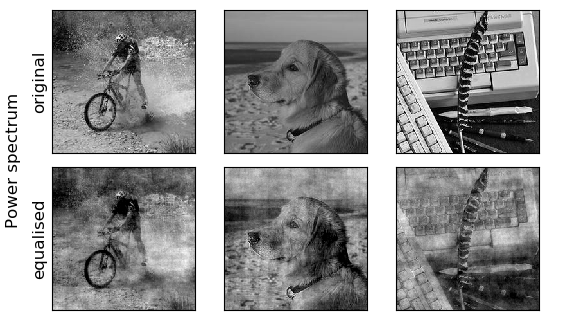}
        \caption{Power equalisation}
    \end{subfigure}\hfill
\caption{Three example stimuli for different conditions of Eidolon III, phase noise, false colour and power equalisation experiments. The three images (categories \texttt{bicycle}, \texttt{dog} and \texttt{keyboard}) were drawn randomly from the pool of images used in the experiments. Best viewed on screen.}
\label{fig:stimuli_eidolon_III_phase_scrambling_false_colour_power_equalisation}
\end{figure*}

We implemented the equalisation of the power spectra and phase noise in the Fourier domain. Conversion to frequency domain was accomplished by a fast Fourier transform through the application of the \texttt{fft2()} and then \texttt{fftshift()} functions of the Python package \texttt{scipy.fftpack}. This results in a matrix of complex numbers $F$, which represents both the phases and amplitudes of the individual frequencies in one complex number. $F$ is organised in symmetric pairs of complex numbers with just their imaginary part differing in its sign and cancelling each other out when reversing the Fourier transform again. When transforming $F$ to polar coordinates, the angle represents the respective frequency's phase and the distance from the origin represents its amplitude. Hence, we extracted the phases and amplitudes of the individual frequencies with the functions \texttt{numpy.angle(F)} and \texttt{numpy.abs(F)}, respectively. 
The \textbf{power equalisation experiment} had two conditions: original and power-equalised (Figure \ref{fig:stimuli_eidolon_III_phase_scrambling_false_colour_power_equalisation}). For the power-equalised images, we first calculated the mean amplitude spectrum over all test images, which showed the typical $\frac{1}{f}$ shape \cite[e.g.][]{VanderSchaaf1996, Wichmann2010}. Thereafter, we set all images amplitudes to the mean amplitude spectrum. Since the power spectrum is the square of the amplitude spectrum, the images were essentially power-equalised. 
There were seven conditions in the \textbf{phase noise experiment}. These were 0, 30, 60, 90, 120, 150 and 180 degrees noise width $w$ (Figure \ref{fig:stimuli_eidolon_III_phase_scrambling_false_colour_power_equalisation}). To each frequency's phase a phase shift randomly drawn from a continuous uniform distribution over the interval $[-w, w]$ was added. To ensure that the imaginary parts would later cancel out again, we added the same phase noise to both frequencies of each symmetric pair. 
After performing the respective manipulations, a $F_{new}$ was calculated by recombining the new phases and amplitudes. Then we did an inverse Fourier transform using \texttt{ifftshift()} and then \texttt{ifft2()}. Finally we clipped all pixel values to the [0, 1] range. This was the case for 0.038\% of pixels with a mean clipped value of about 0.003 for the phase noise experiment and for 0.013\% of pixels with a mean clipped value of 0.005 for the power-equalisation experiment. 

There were four conditions for the \textbf{rotation experiment}: 0 (original), 90, 180 and 270 degrees rotation angle. Rotation by 90 degrees was accomplished by first transposing the image matrix and then reversing the column order.  Rotation by 180 degrees was done by reversing both, row and column ordering. Rotation by 270 degrees was implemented by first reversing the images columns and then transposing it.

\begin{figure*}
    \begin{subfigure}{0.5\textwidth}
        \includegraphics[width=\linewidth]{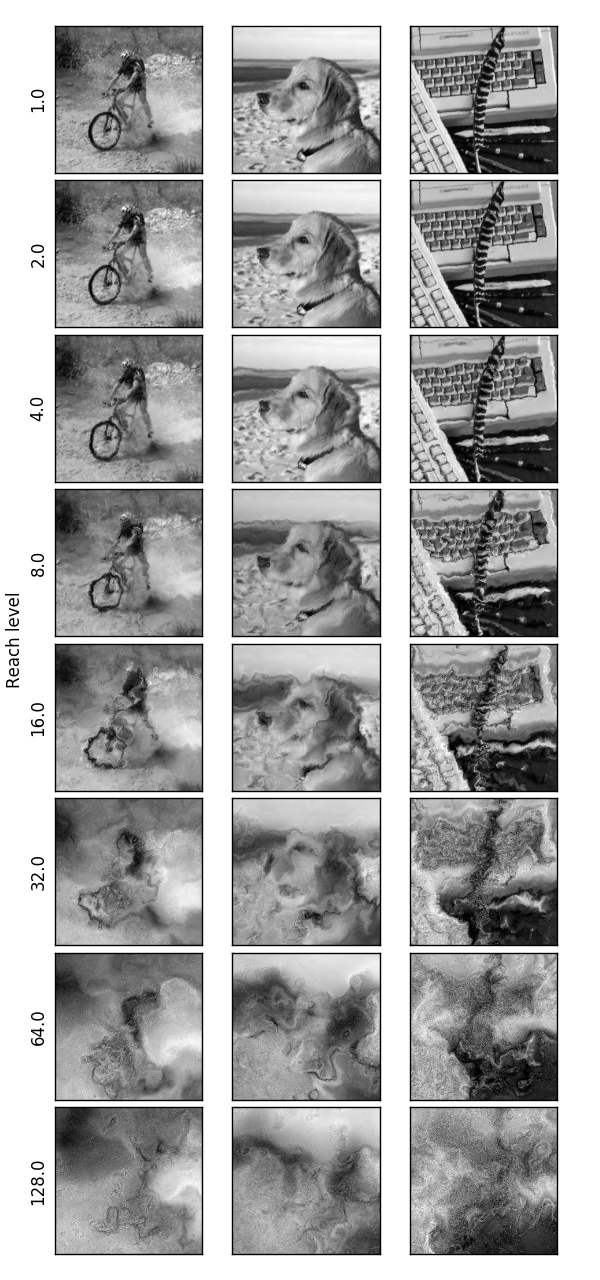}
        \vspace{\captionspace}
        \caption{Eidolon I}
    \end{subfigure}\hfill
    \begin{subfigure}{0.5\textwidth}
        \includegraphics[width=\linewidth]{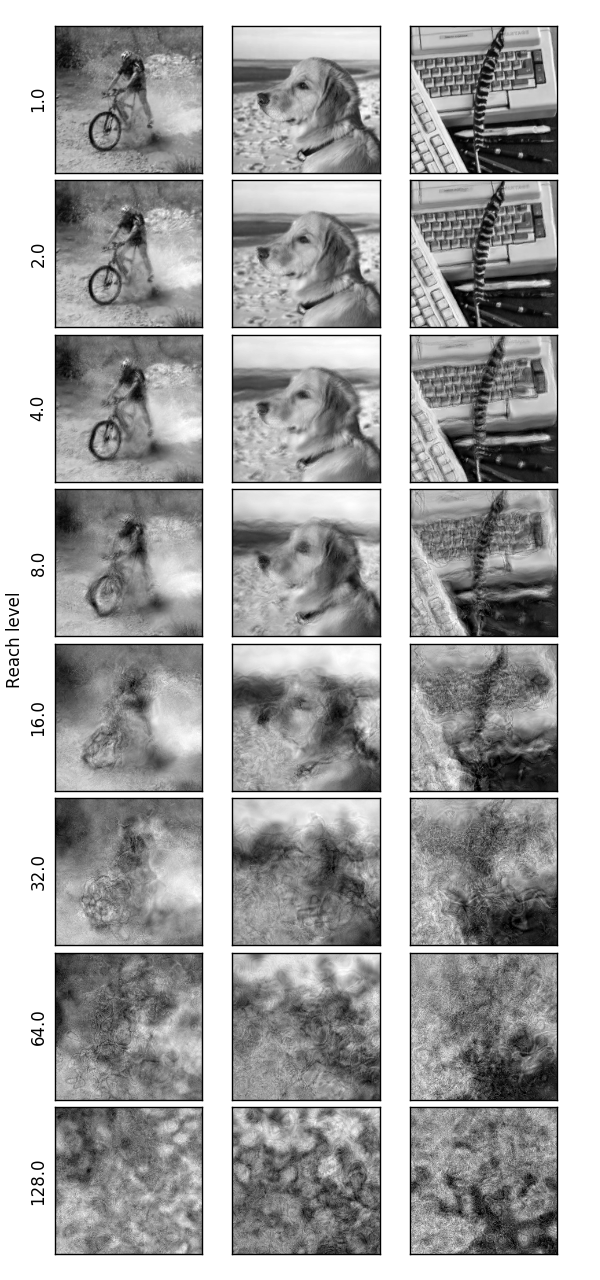}
        \vspace{\captionspace}
        \caption{Eidolon II}
    \end{subfigure}\hfill
\caption{Three example stimuli for different conditions of eidolon I and eidolon II experiments. The three images (categories \texttt{bicycle}, \texttt{dog} and \texttt{keyboard}) were drawn randomly from the pool of images used in the experiments. Best viewed on screen.}
\label{fig:stimuli_eidolon_I_II}
\end{figure*}

For the \textbf{eidolon experiments}, all stimuli were generated using the eidolon toolbox for Python\footnote{\url{https://github.com/gestaltrevision/Eidolon}}, more specifically its \texttt{PartiallyCoherentDisarray(image, reach, coherence, grain)} function.
Using a combination of the three parameters reach, coherence and grain, one obtains a distorted version of the original image (a so-called eidolon). The parameters reach and coherence were varied in the experiment; grain was held constant with a value of 10.0 throughout the experiment (grain indicates how fine-grained the distortion is; a value of 10.0 corresponds to a medium-grainy distortion). Reach $ \in \{1.0, 2.0, 4.0, 8.0, 16.0, 32.0, 64.0, 128.0\} $ is an amplitude-like parameter indicating the strength of the distortion, coherence $ \in \{0.0, 0.3, 1.0\}$ defines the relationship between local and global image structure. Those two parameters were fully crossed, resulting in $ 8 \cdot 3 = 24 $ different eidolon conditions. A high coherence value ``retains the local image structure even when the global image structure is destroyed'' \cite[p. 10]{Koenderink2017}. A coherence value of 0.0 corresponds to `completely incoherent', a value of 1.0 to `fully coherent'. The third value 0.3 was chosen because it produces images that perceptually lie---as informally determined by the authors---in the middle between those two extremes. See Figures~\ref{fig:stimuli_eidolon_I_II} and \ref{fig:stimuli_eidolon_III_phase_scrambling_false_colour_power_equalisation} for example eidolon stimuli. The coherence levels of 1.0, 0.3 and 0.0 are refered to as eidolon experiment I, II and III throughout the paper.

\subsection*{Experimental modifications}
Our psychophysical experiments were conducted in two batches and over an extended period of time. After completing the first batch of experiments (all experiments on the left half of Figure~\ref{fig:results_accuracy_entropy}, i.e. a, c, e, g, i and k), we performed a number of modifications for the second batch of experiments. We here briefly list all the changes in which the second batch of experiments differed from the previously reported methods.

\emph{Noise mask:} In the human experiments, each experimental image was immediately followed by a $\frac{1}{f}$ pink noise mask (cf. Figure~\ref{fig:typical_trial}). In the second batch of experiments this noise mask was enhanced to improve its masking effect. This was done by multiplying each pixel value by four. Values greater than $1$ or smaller than $0$ due to the multiplication were then clipped to $1$ or $0$.

\emph{Cropping vs. downsampling}: In the first experimental batch, humans saw $256 \times 256$ images. However, DNN classification was based on those images $224 \times 224$ centre crop. Thus, humans and DNNs saw slightly different images. Therefore, we used downsampling to $224 \times 224$ for the second batch of both, human and DNN experiments. As a consequence, the mean grey pixel value over all experimental images and hence the background grey value for presenting those images changed slightly from $0.454$ to $0.442$ in the [0, 1] range.

\emph{JPEG vs. PNG}: 
All images of the first batch of experiments, prior to showing them to human observers or DNNs, were saved in the JPEG format using the default settings of the \texttt{skimage.io.imsave} function. The JPEG format was chosen because the image training database for all three networks, ImageNet \cite{Russakovsky2015}, consists of JPEG images. However, as the lossy compression of JPEG may introduce artefacts, we also examined the difference in DNN results between saving to JPEG and to PNG, which is lossless up to rounding issues. We therefore ran all those DNN experiments additionally saving them in the (up to rounding issues) lossless PNG format. We did not find any noteworthy differences for the colour, noise, and eidolon experiments. However, for the contrast experiment, the networks achieved on average better results for PNG images. We therefore tested three human observers additionally on the same stimuli (PNG instead of JPEG images). In this experiment, three of the JPEG experiment's five observers participated for maximal comparability.\footnote{A time gap of approximately six months between both experiments should minimise memory effects; furthermore, human participants were not shown any feedback (correct / incorrect classification choice) during the experiments.} We found human observers to be better for PNG images as well. In absolute terms, participants were 2.68\% better on average. In order to disentangle the influence of JPEG compression and image manipulations, we used PNG images for all other experiments, that is for the false colour, phase noise, power equalisation, rotation, high-pass and low-pass experiments as well as for the DNN training experiments.

\emph{Python Version:} The second batch used Python Version 3.5 instead of Python 2.7 for image preprocessing.

\begin{figure}
	\begin{subfigure}{\figwidth}
	    \centering
	    \includegraphics[width=\linewidth]{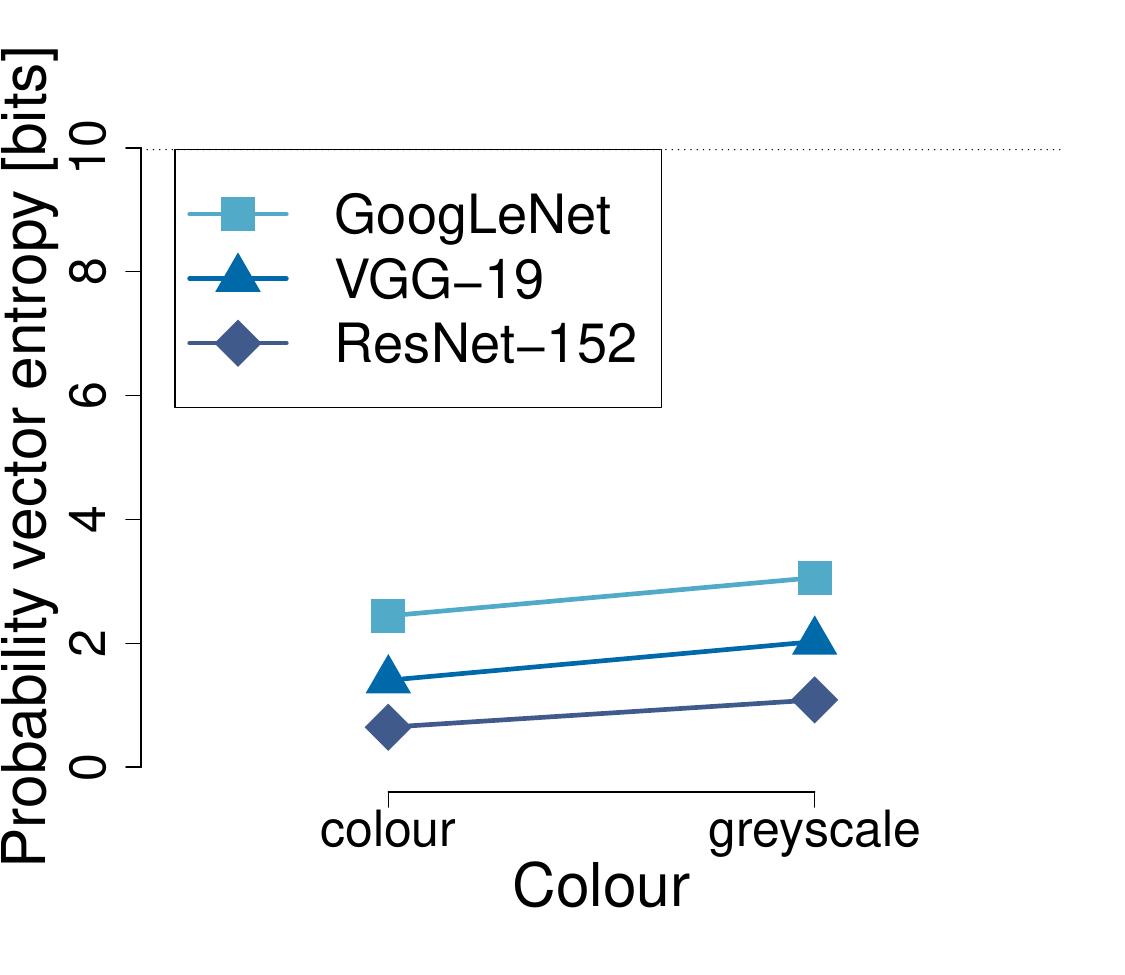}
        \vspace{\captionspace}
        \caption{Colour vs. greyscale}
	\end{subfigure}\hfill
		\begin{subfigure}{\figwidth}
	    \centering
	    \includegraphics[width=\linewidth]{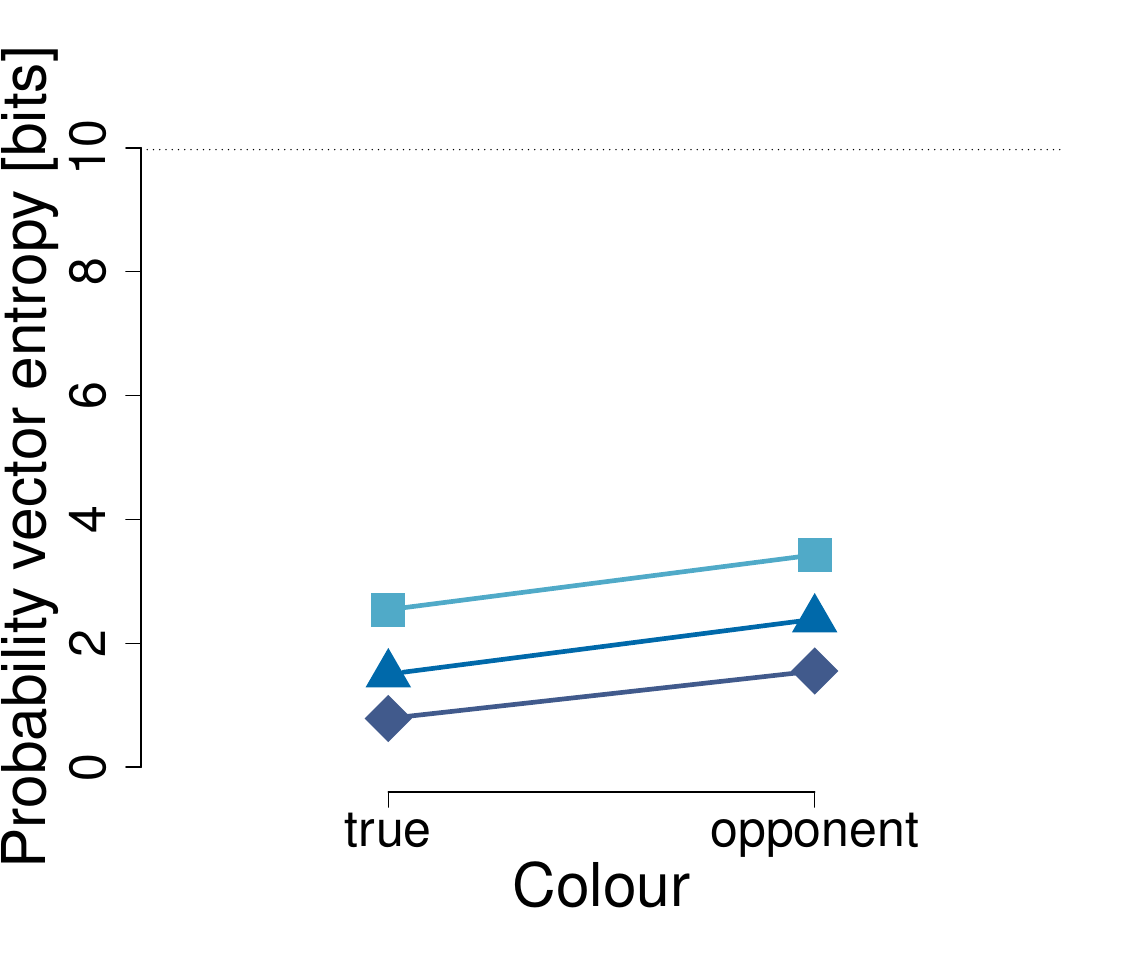}
        \vspace{\captionspace}
        \caption{True vs. false colour}
	\end{subfigure}\hfill
	\begin{subfigure}{\figwidth}
	    \centering
	    \includegraphics[width=\linewidth]{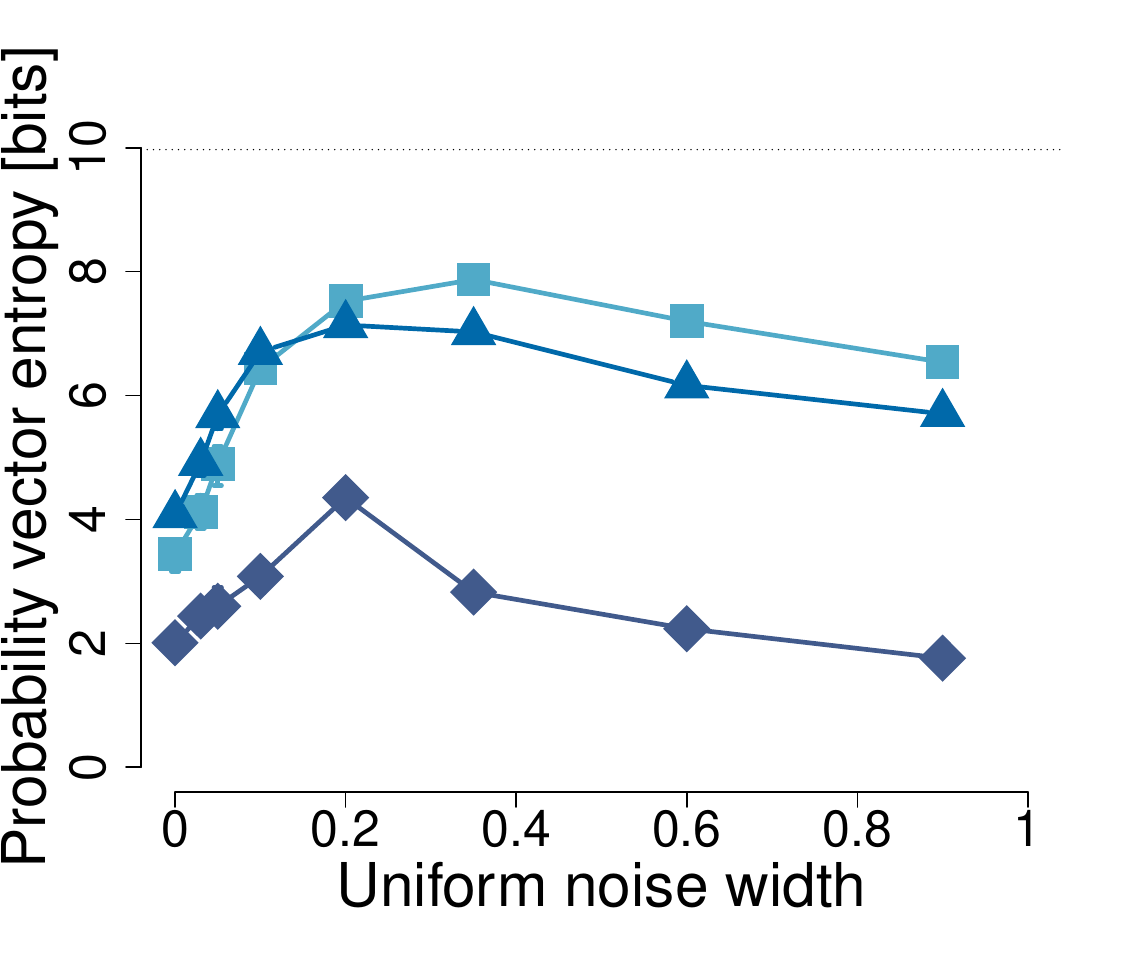}
        \vspace{\captionspace}
        \caption{Uniform noise}
	\end{subfigure}\hfill
	\begin{subfigure}{\figwidth}
	    \centering
	    \includegraphics[width=\linewidth]{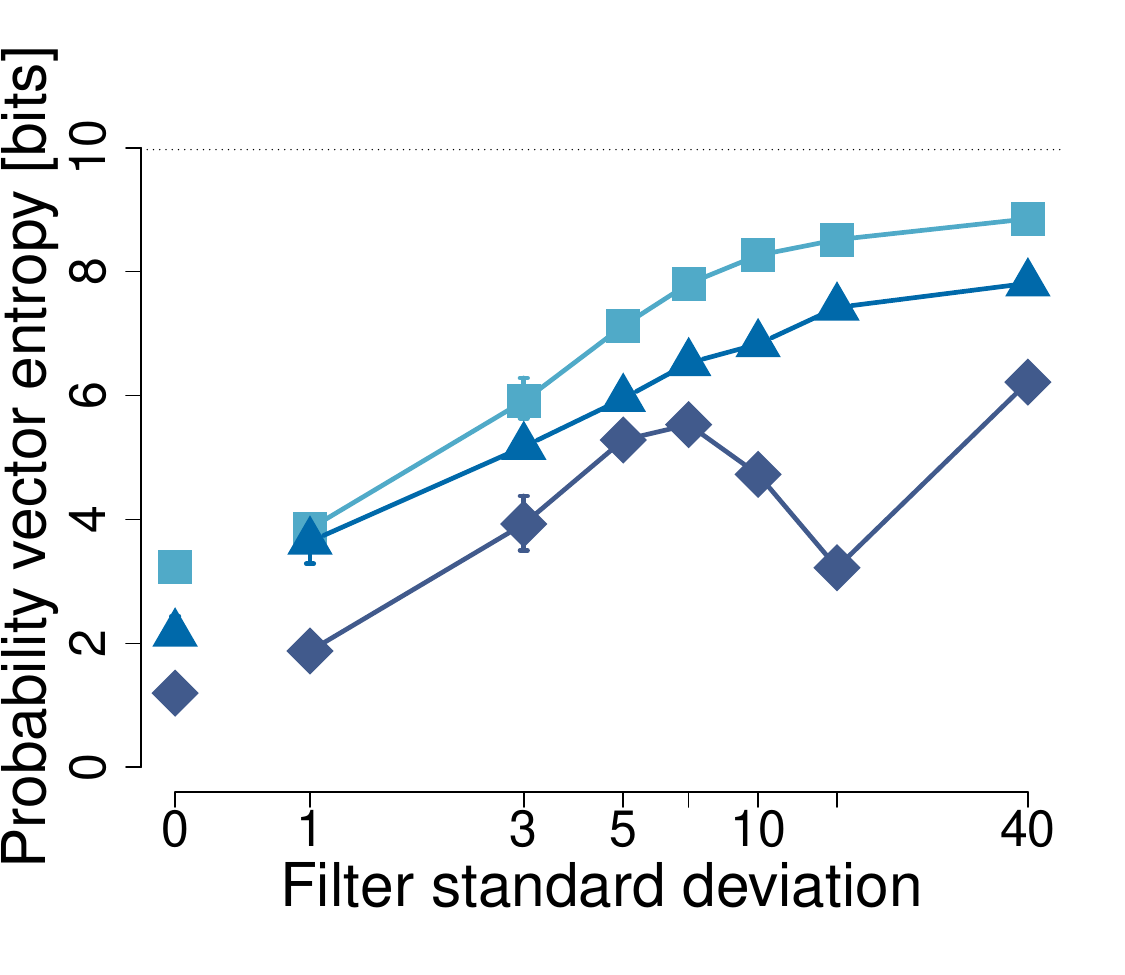}
        \vspace{\captionspace}
        \caption{Low-pass}
	\end{subfigure}\hfill
	
	\begin{subfigure}{\figwidth}
	    \centering
	    \includegraphics[width=\linewidth]{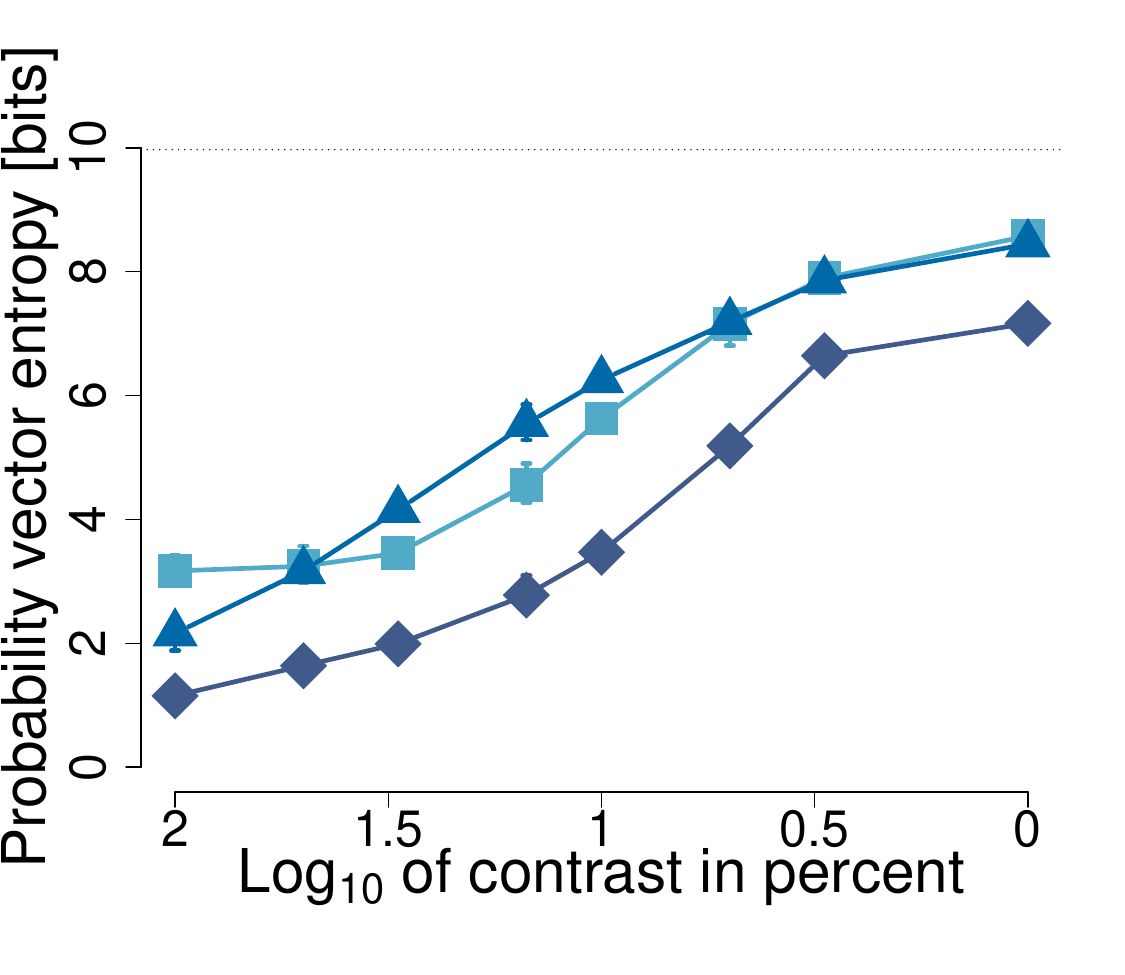}
        \vspace{\captionspace}
        \caption{Contrast}
	\end{subfigure}\hfill
	\begin{subfigure}{\figwidth}
	    \centering
	    \includegraphics[width=\linewidth]{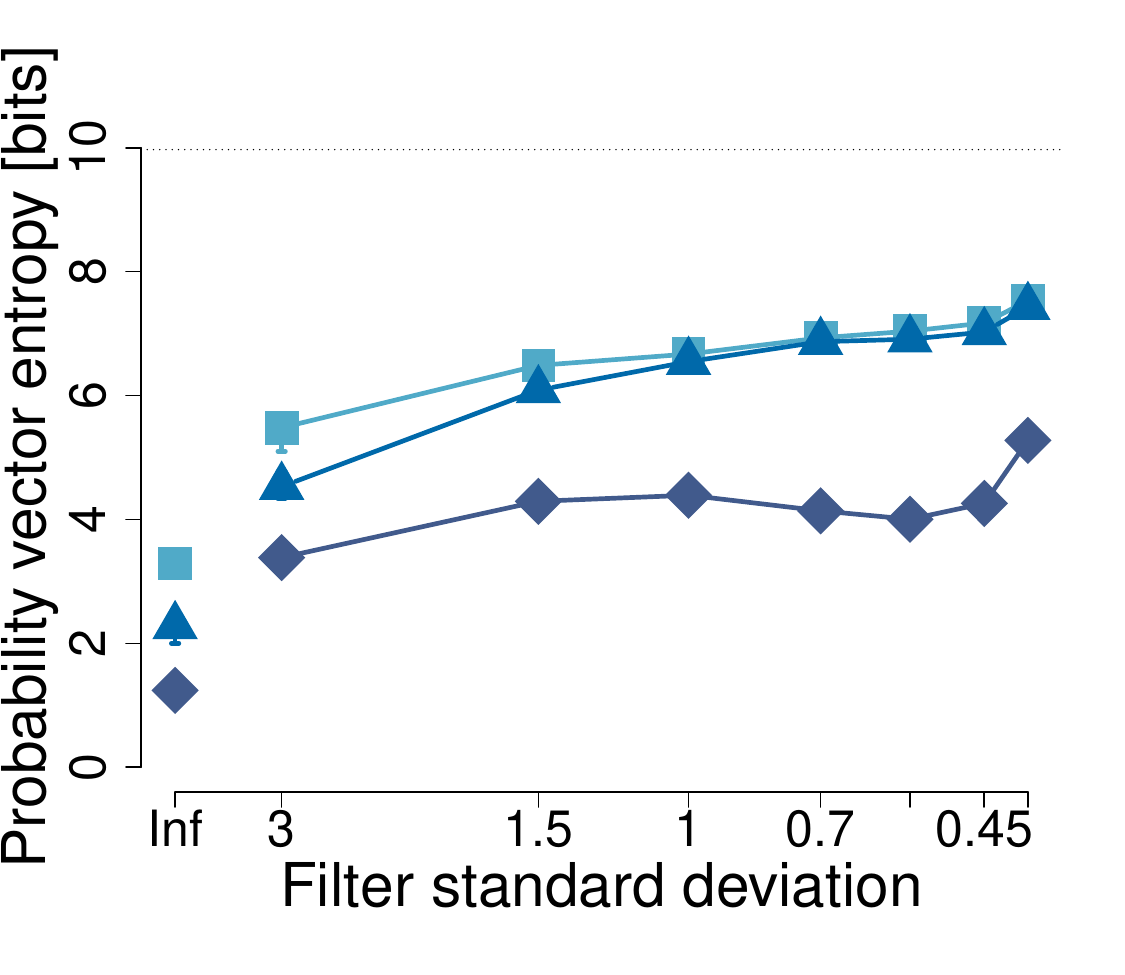}
        \vspace{\captionspace}
        \caption{High-pass}
	\end{subfigure}\hfill
	\begin{subfigure}{\figwidth}
	    \centering
	    \includegraphics[width=\linewidth]{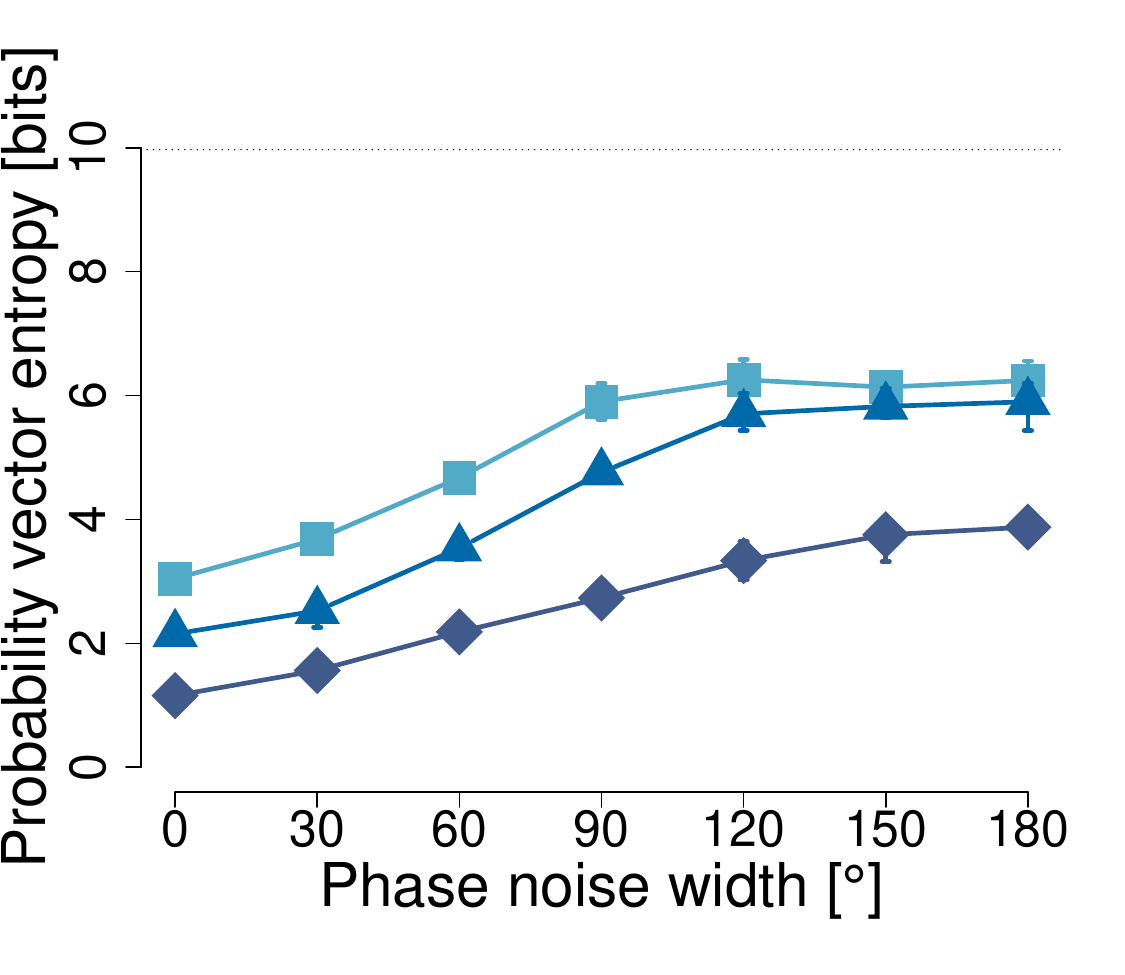}
        \vspace{\captionspace}
        \caption{Phase noise}
	\end{subfigure}\hfill
	\begin{subfigure}{\figwidth}
	    \centering
	    \includegraphics[width=\linewidth]{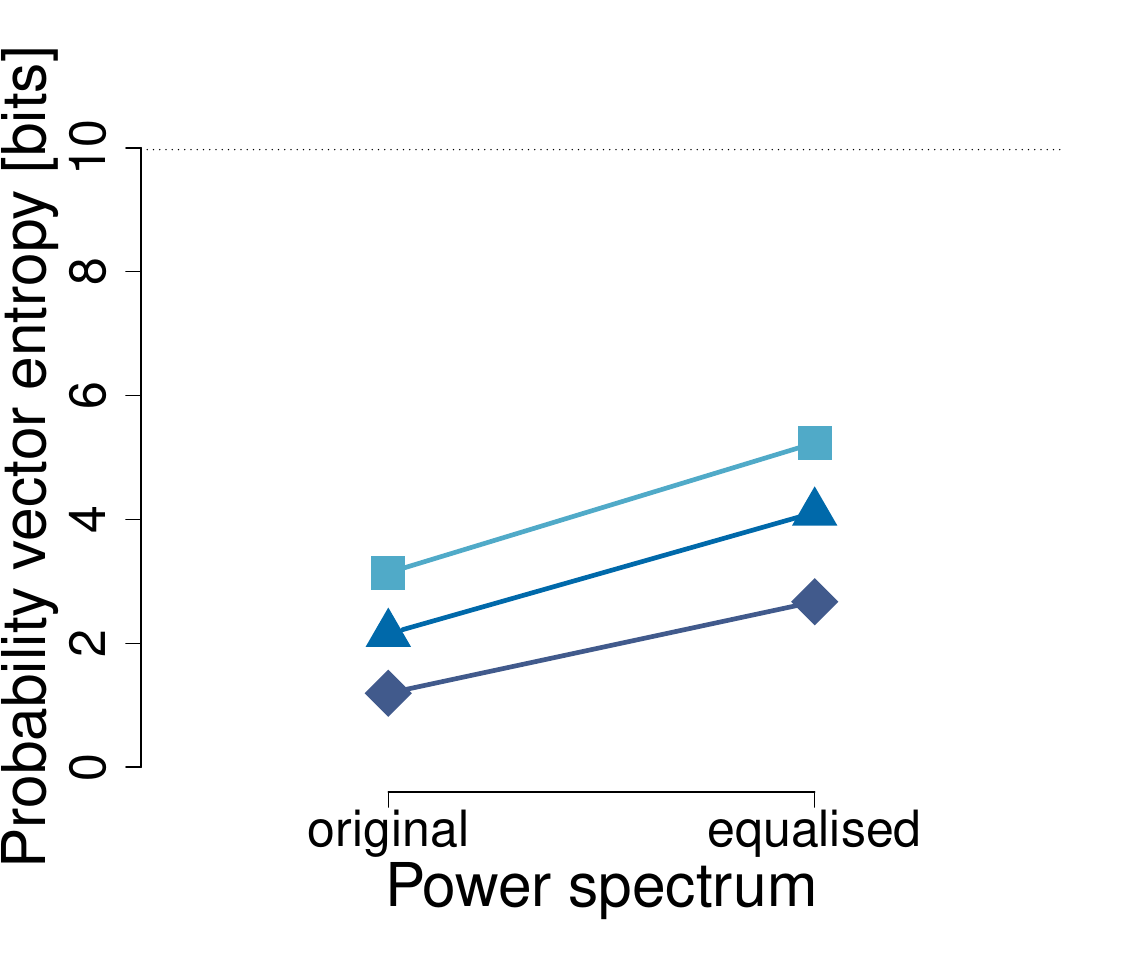}
        \vspace{\captionspace}
        \caption{Power equalisation}
	\end{subfigure}\hfill
	
	\begin{subfigure}{\figwidth}
	    \centering
	    \includegraphics[width=\linewidth]{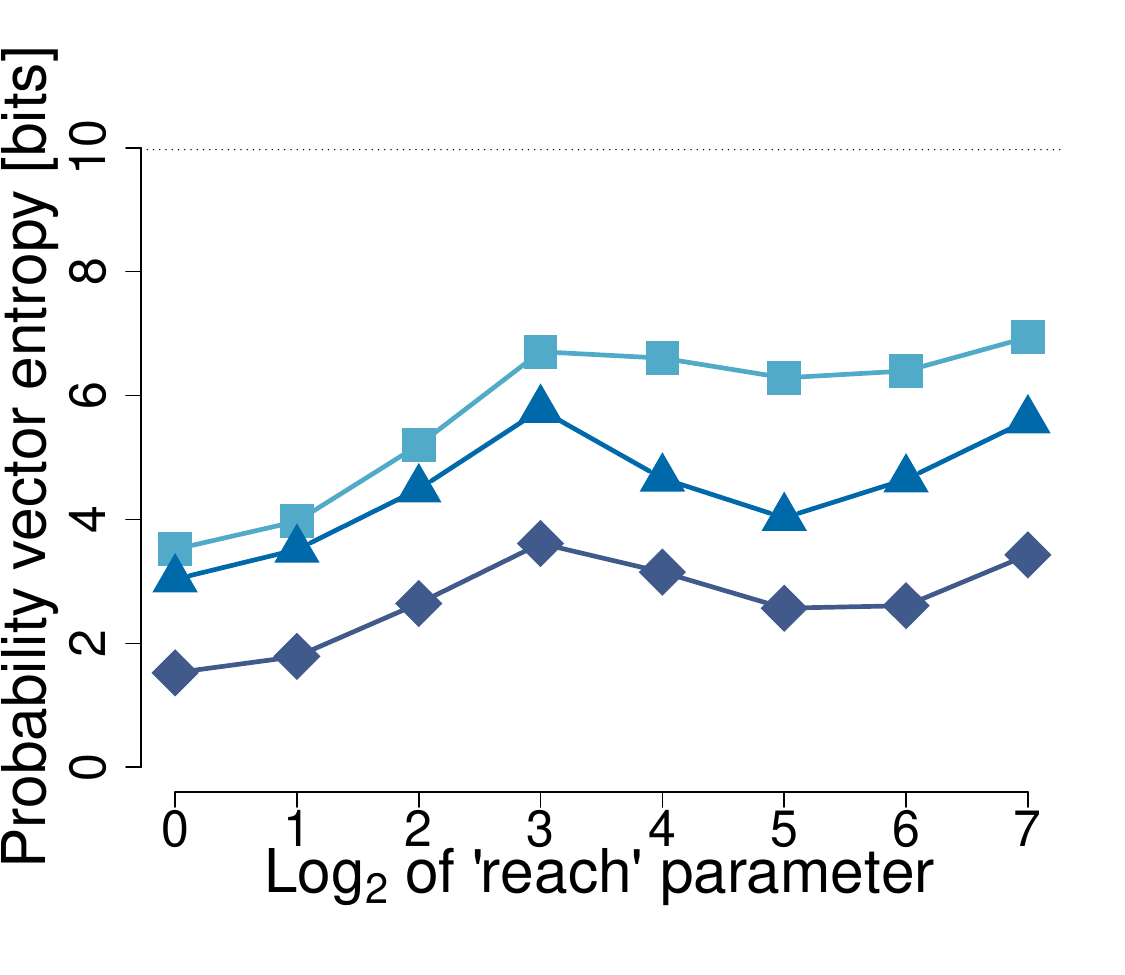}
        \vspace{\captionspace}
        \caption{Eidolon I}
	\end{subfigure}\hfill
	\begin{subfigure}{\figwidth}
	    \centering
	    \includegraphics[width=\linewidth]{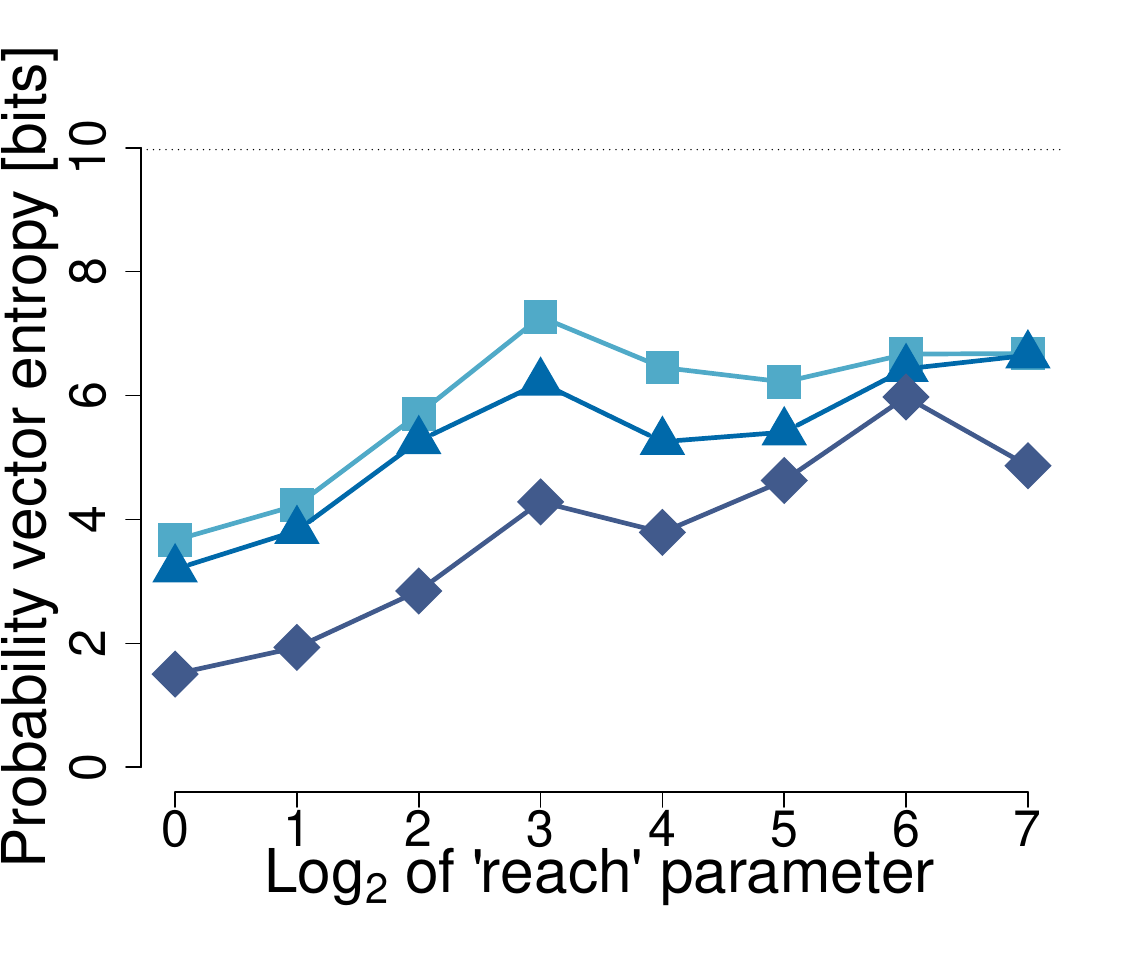}
        \vspace{\captionspace}
        \caption{Eidolon II}
	\end{subfigure}\hfill
	\begin{subfigure}{\figwidth}
	    \centering
	    \includegraphics[width=\linewidth]{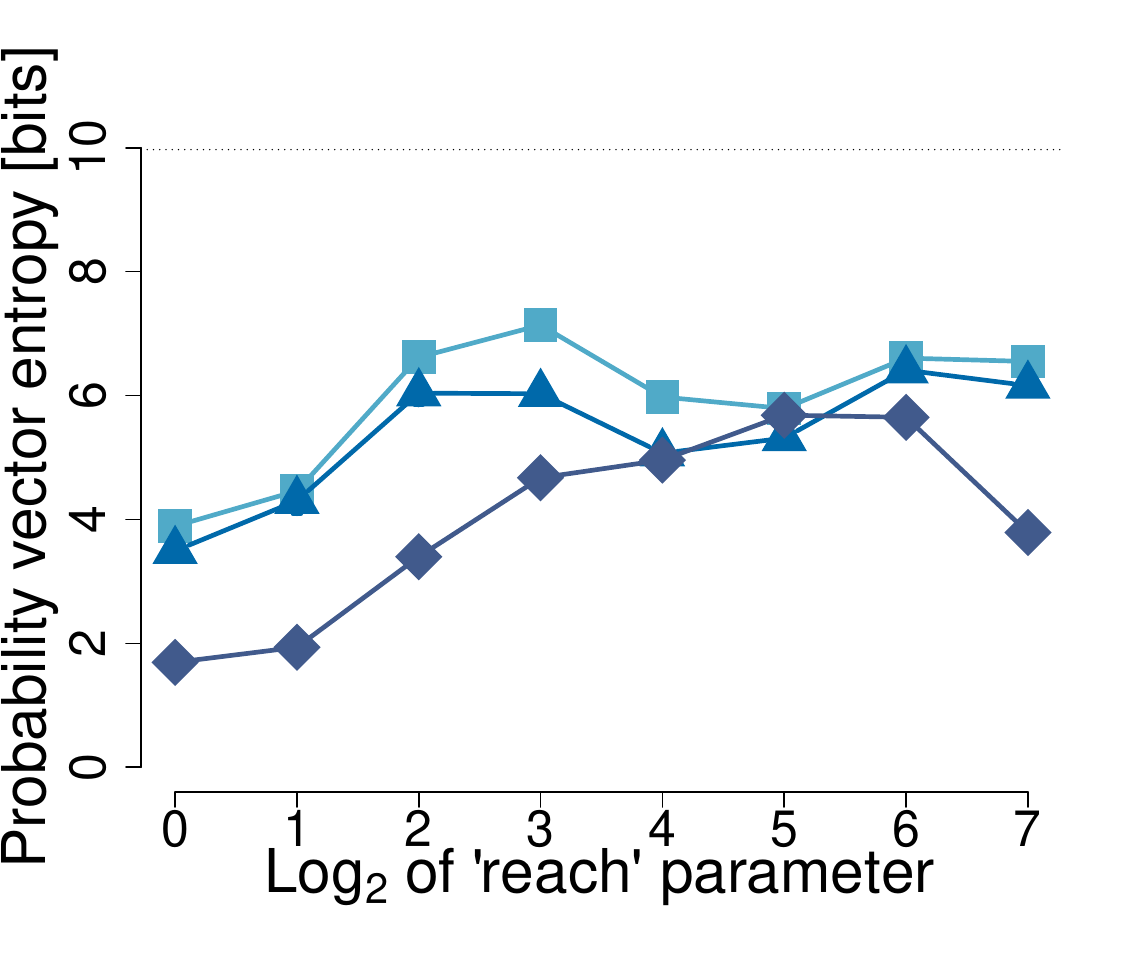}
        \vspace{\captionspace}
        \caption{Eidolon III}
	\end{subfigure}\hfill
	\begin{subfigure}{\figwidth}
	    \centering
	    \includegraphics[width=\linewidth]{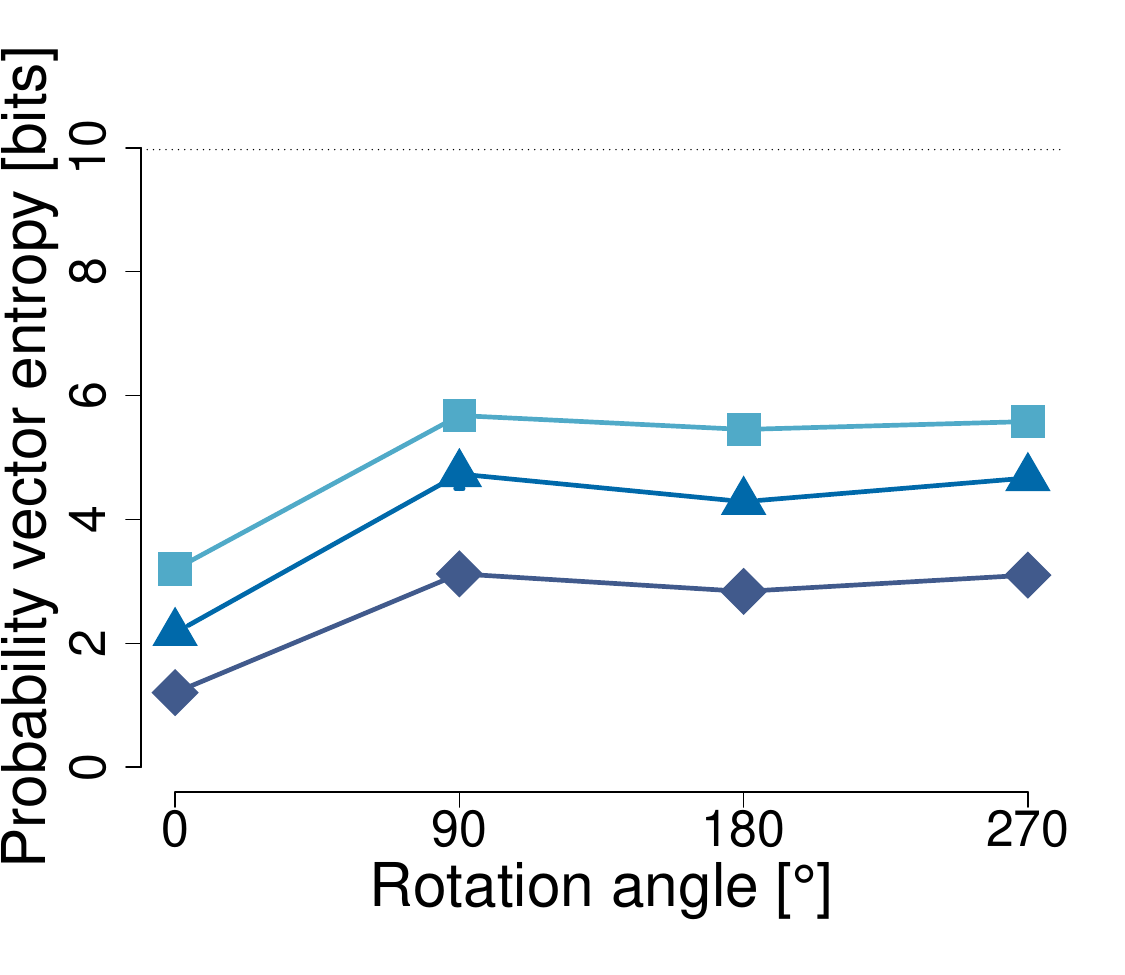}
        \vspace{\captionspace}
        \caption{Rotation}
	\end{subfigure}\hfill
	\caption{Mean entropy of the probabilities for the 1000 ILSVRC classes for \textcolor{googlenet.100}{GoogLeNet}, \textcolor{vgg.100}{VGG-19} and \textcolor{resnet.100}{ResNet-152}. Dotted line indicates the maximum possible entropy. This is a measure of network `uncertainty'.}
    \label{fig:uncertainty}
\end{figure}

\begin{figure}[h]
    \begin{subfigure}{\figwidth}
        \centering
        \textbf{Accuracy}\\
        \includegraphics[width=\linewidth]{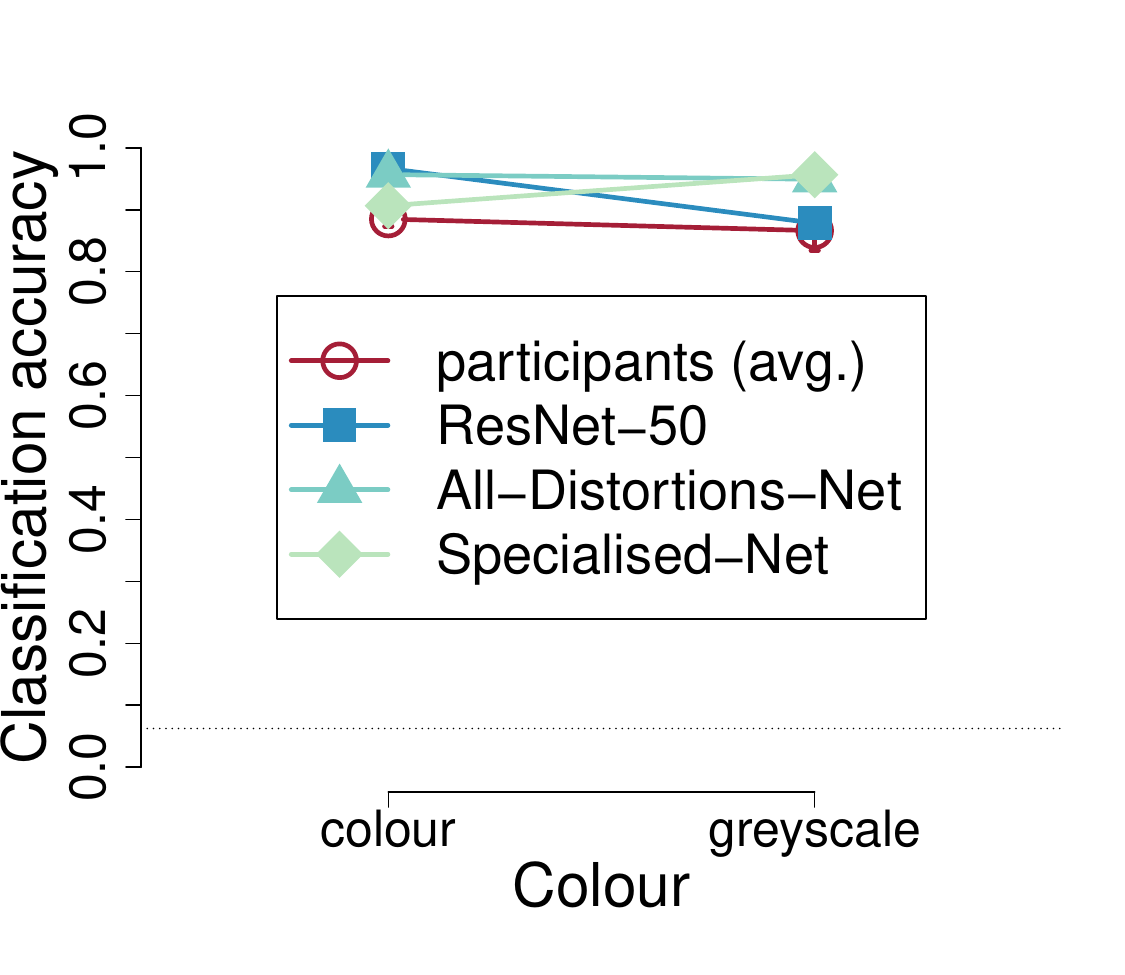}
        \vspace{\captionspace}
        \caption{Colour vs. greyscale}
    \end{subfigure}\hfill
    \begin{subfigure}{\figwidth}
        \centering
        \textbf{Entropy}\\
        \includegraphics[width=\linewidth]{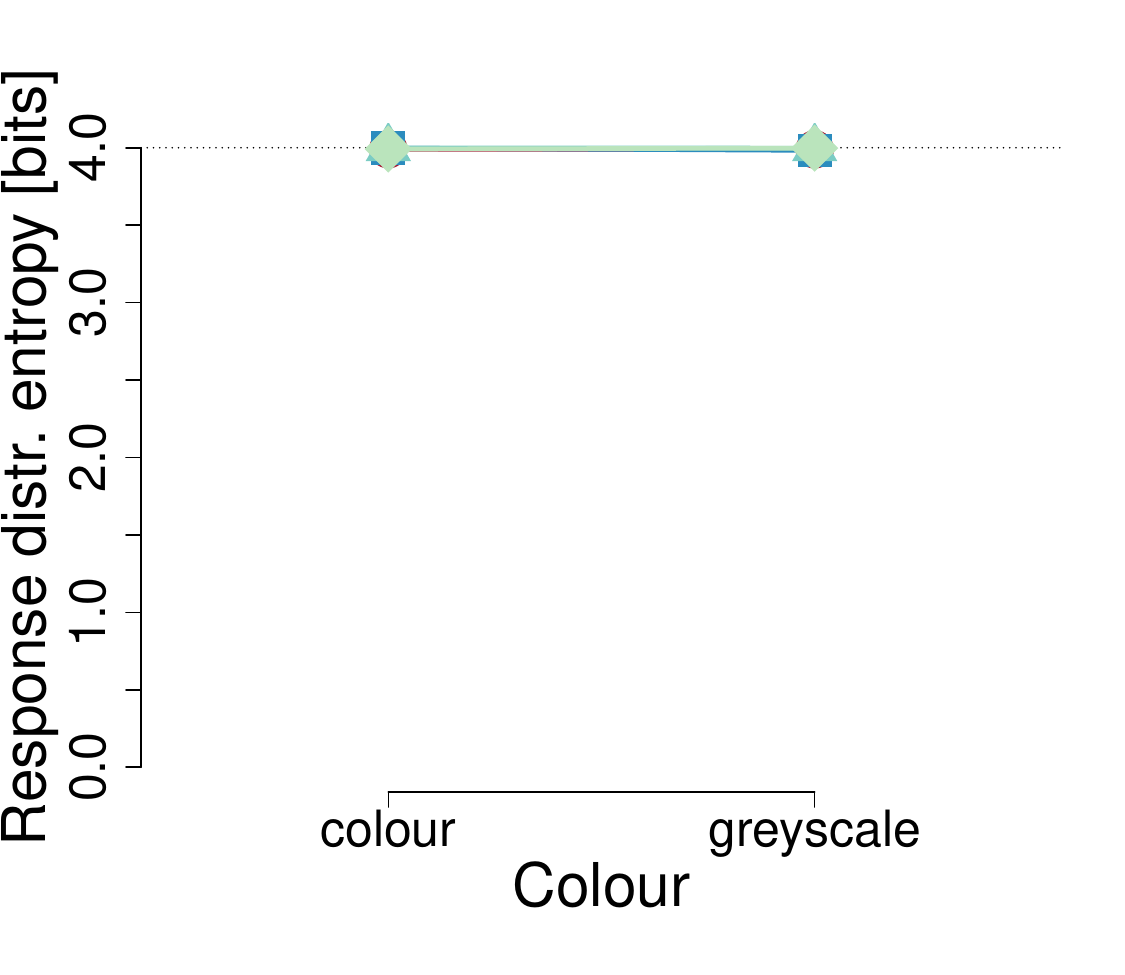}
        \vspace{\captionspace}
        \caption*{}
    \end{subfigure}\hfill
    \begin{subfigure}{\figwidth}
        \centering
        \textbf{Accuracy}\\
        \includegraphics[width=\linewidth]{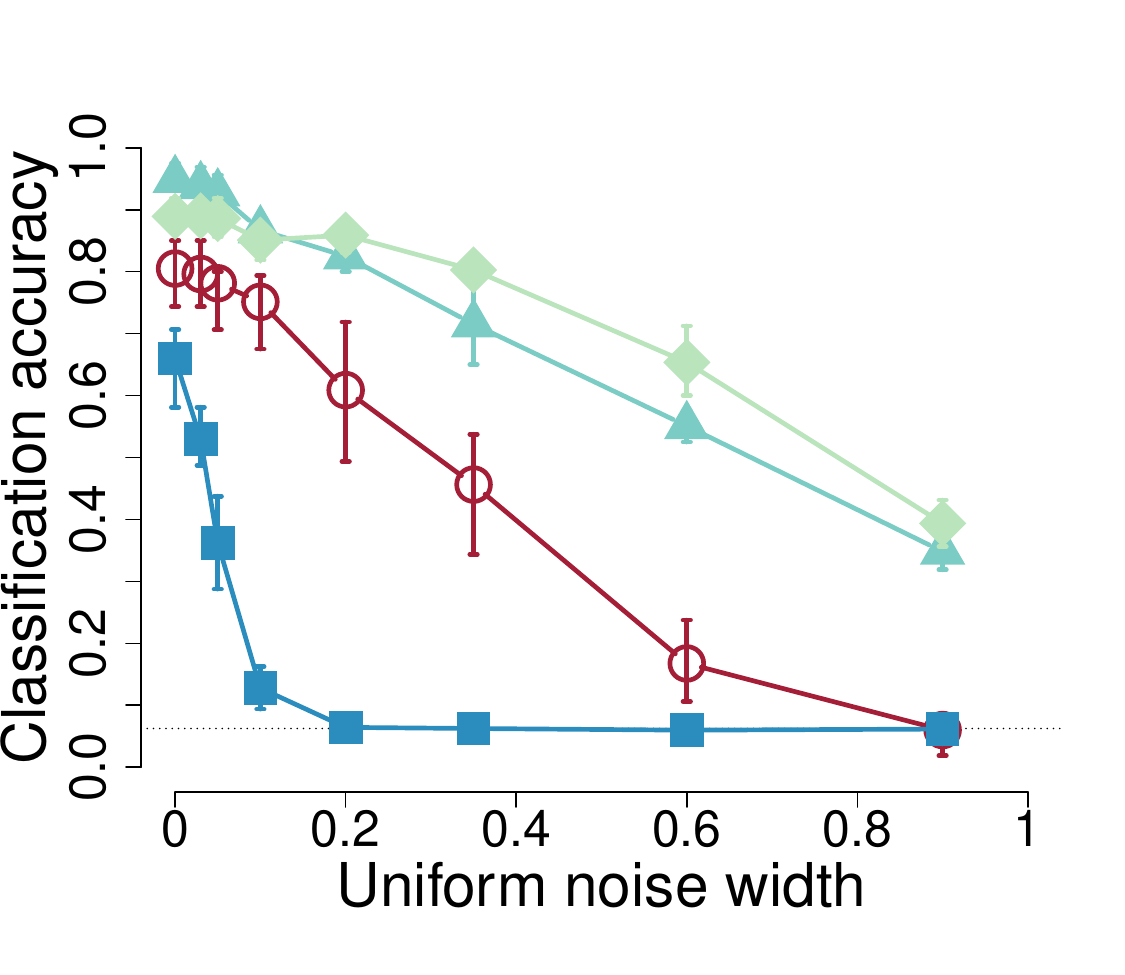}
        \vspace{\captionspace}
        \caption{Uniform noise}
        \label{sup:training_uniform_noise}
    \end{subfigure}\hfill
    \begin{subfigure}{\figwidth}
        \centering
        \textbf{Entropy}\\
        \includegraphics[width=\linewidth]{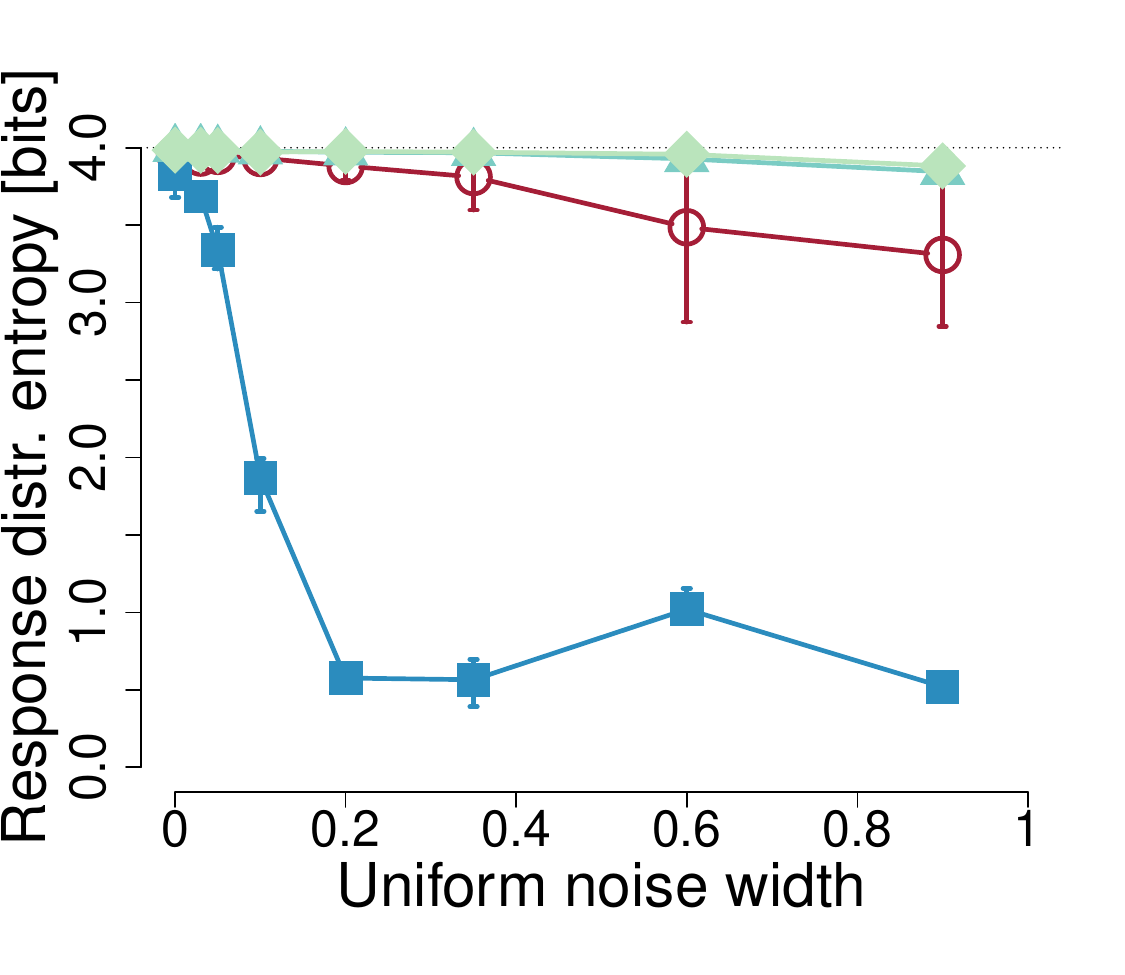}
        \vspace{\captionspace}
        \caption*{}
    \end{subfigure}\hfill

    \begin{subfigure}{\figwidth}
        \centering
        \includegraphics[width=\linewidth]{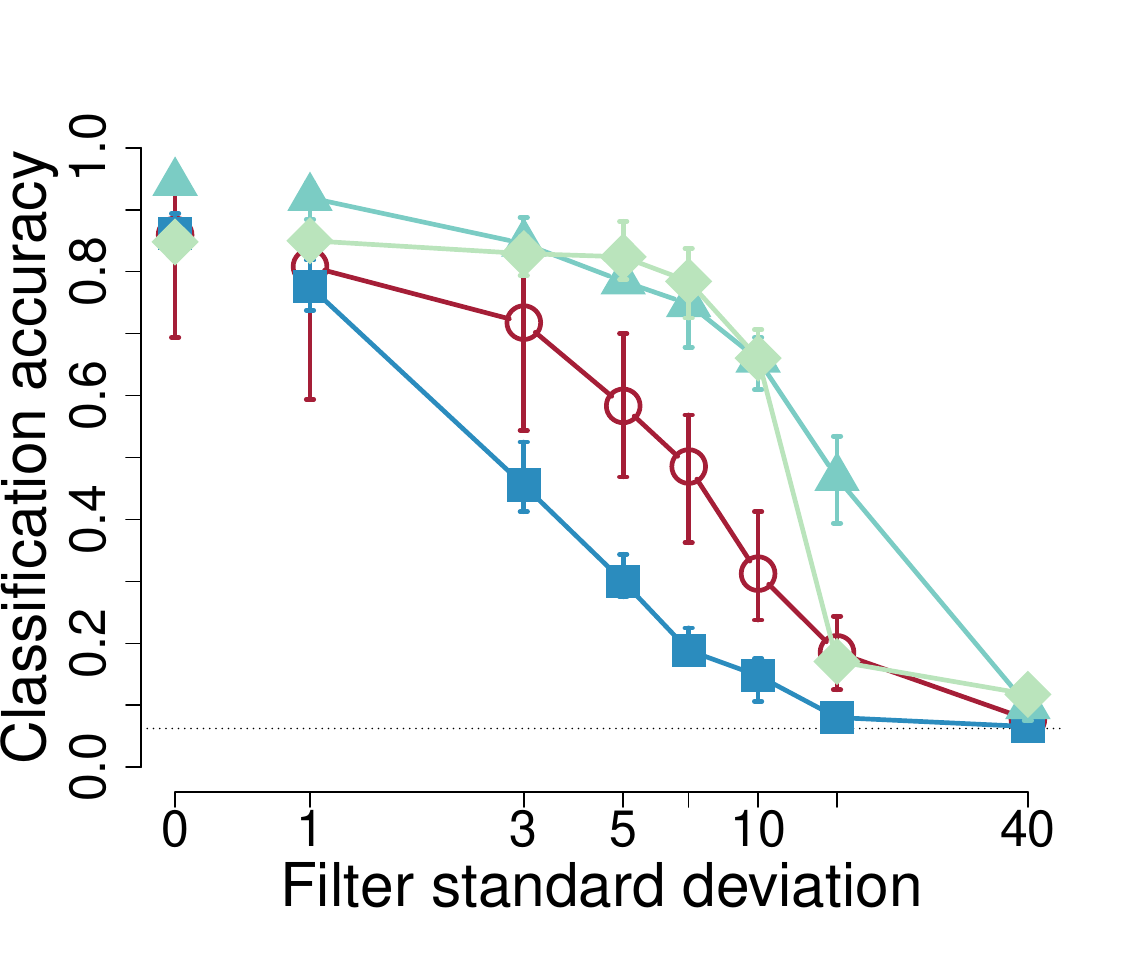}
        \vspace{\captionspace}
        \caption{Low-pass}
    \end{subfigure}\hfill
    \begin{subfigure}{\figwidth}
        \centering
        \includegraphics[width=\linewidth]{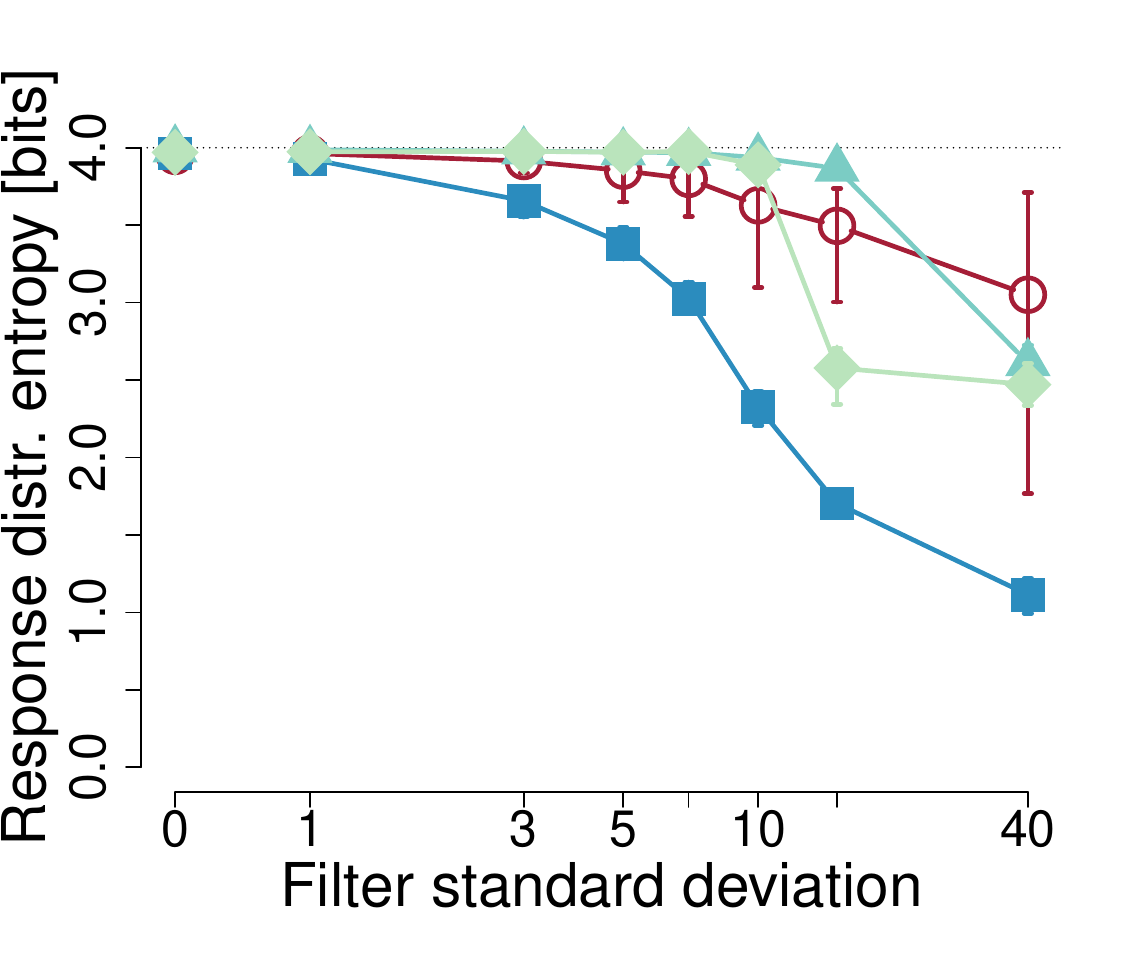}
        \vspace{\captionspace}
        \caption*{}
    \end{subfigure}\hfill
    \begin{subfigure}{\figwidth}
        \centering
        \includegraphics[width=\linewidth]{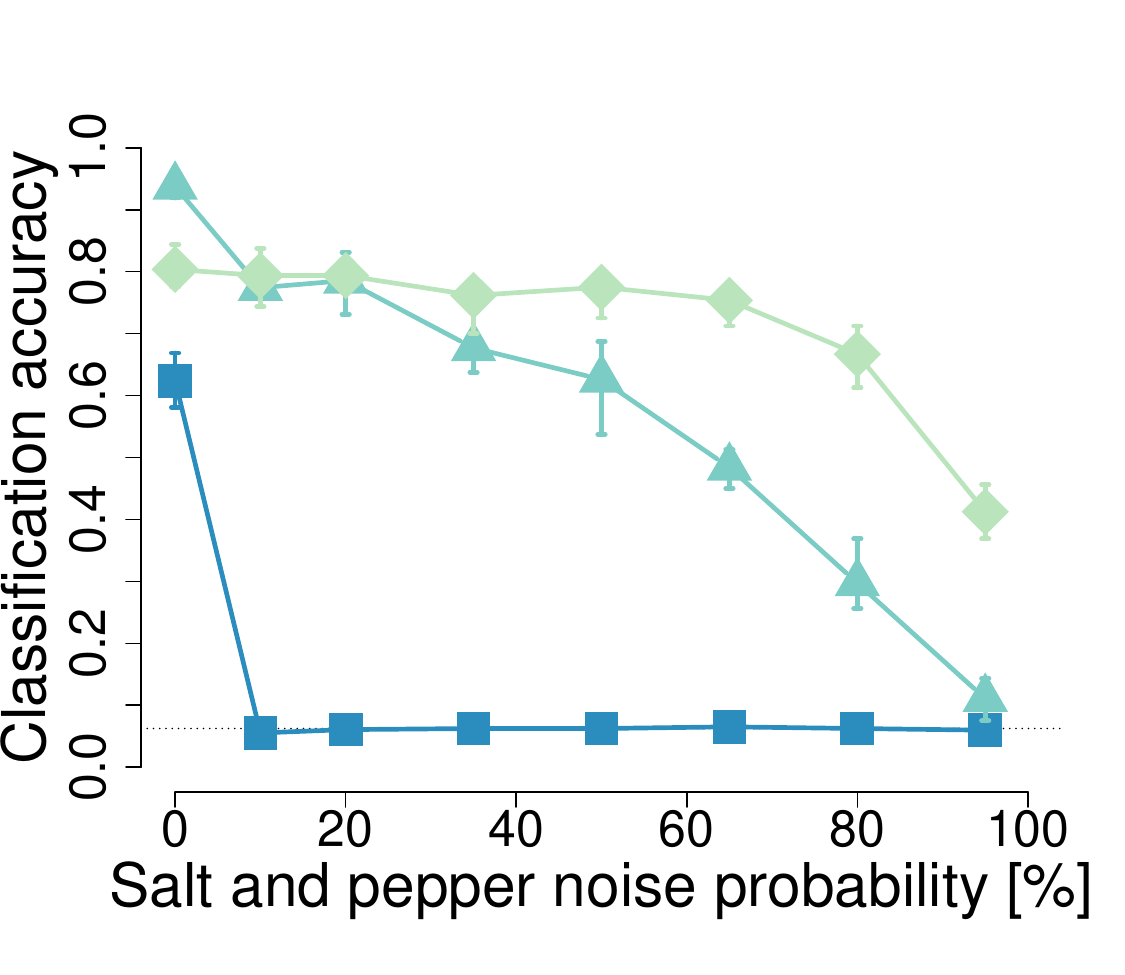}
        \vspace{\captionspace}
        \caption{Salt-and-pepper noise}
        \label{sup:training_salt_and_pepper_noise}
    \end{subfigure}\hfill
    \begin{subfigure}{\figwidth}
        \centering
        \includegraphics[width=\linewidth]{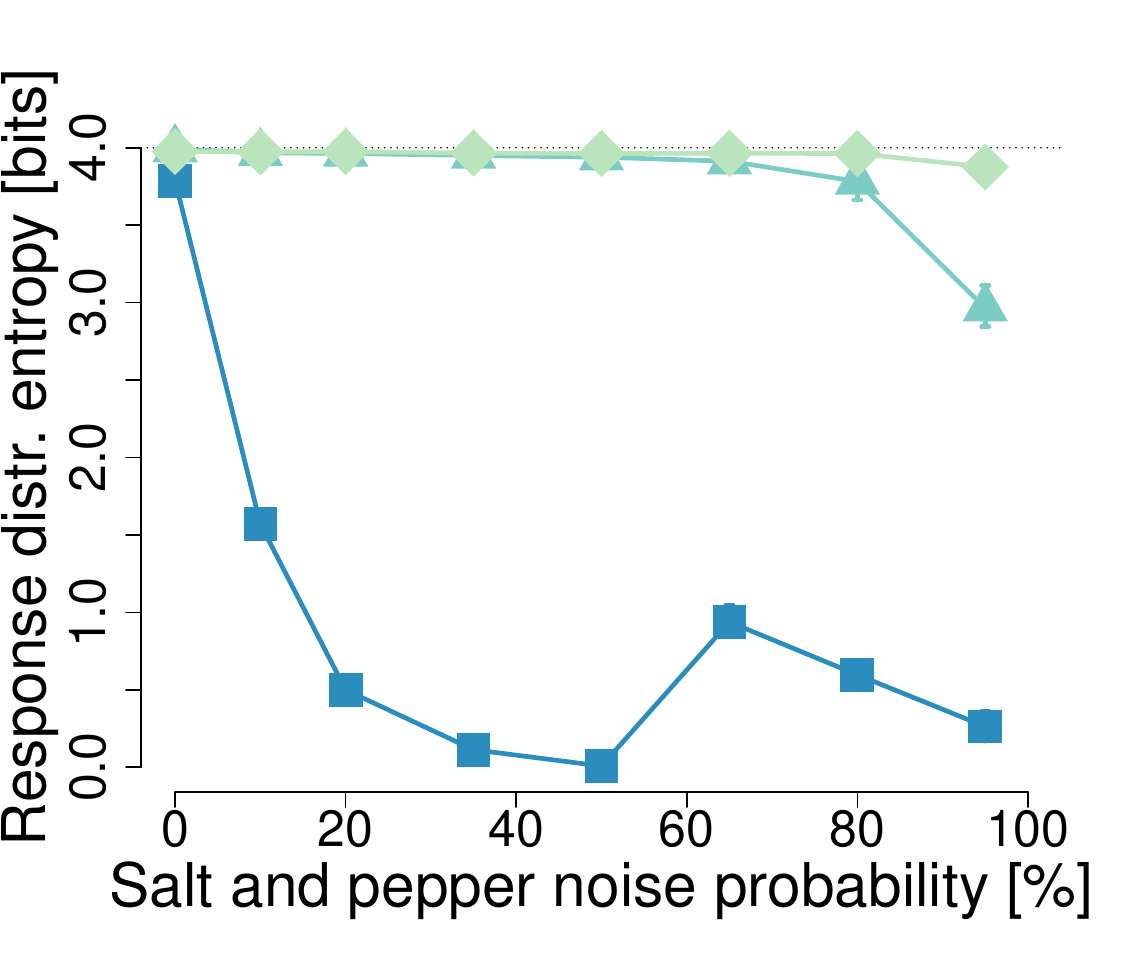}
        \vspace{\captionspace}
        \caption*{}
    \end{subfigure}\hfill

    \begin{subfigure}{\figwidth}
        \centering
        \includegraphics[width=\linewidth]{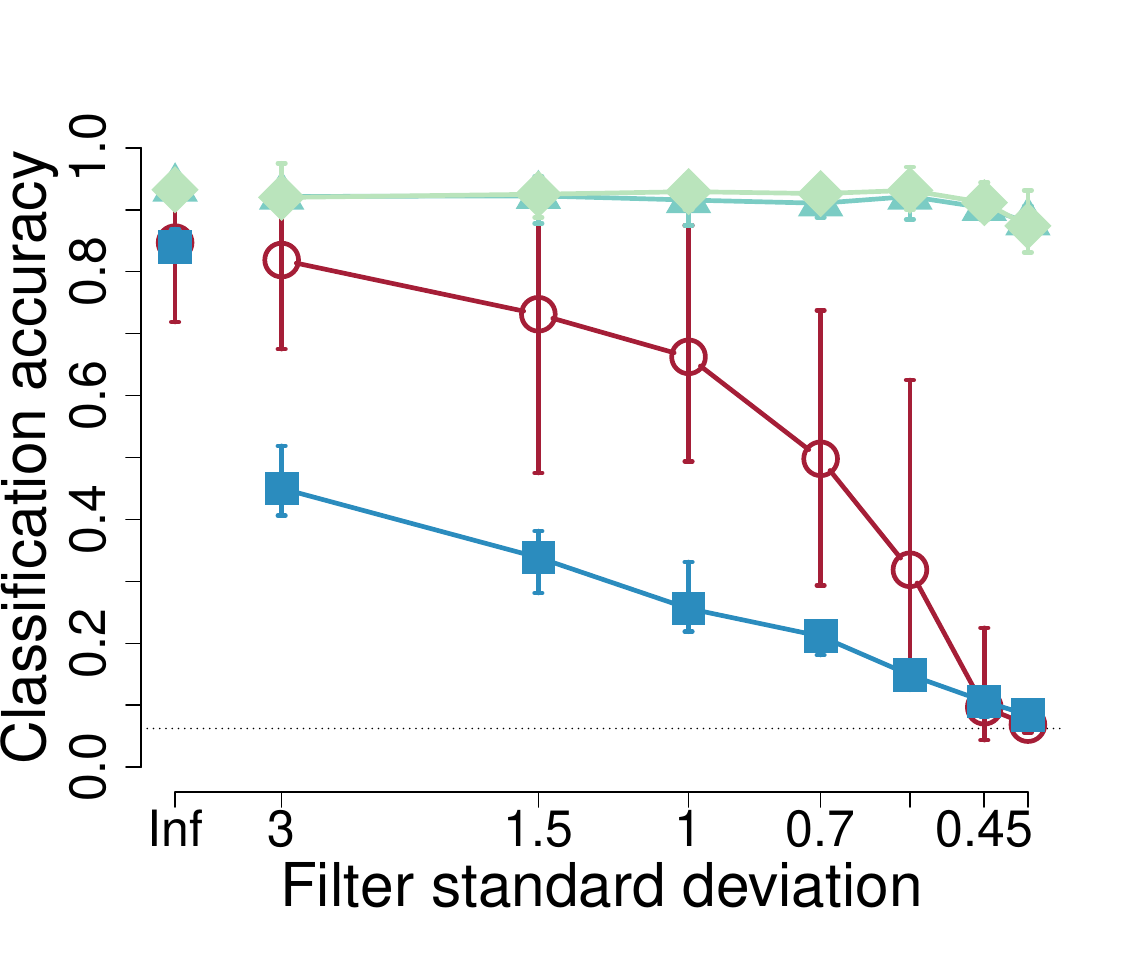}
        \vspace{\captionspace}
        \caption{High-pass}
    \end{subfigure}\hfill
    \begin{subfigure}{\figwidth}
        \centering
        \includegraphics[width=\linewidth]{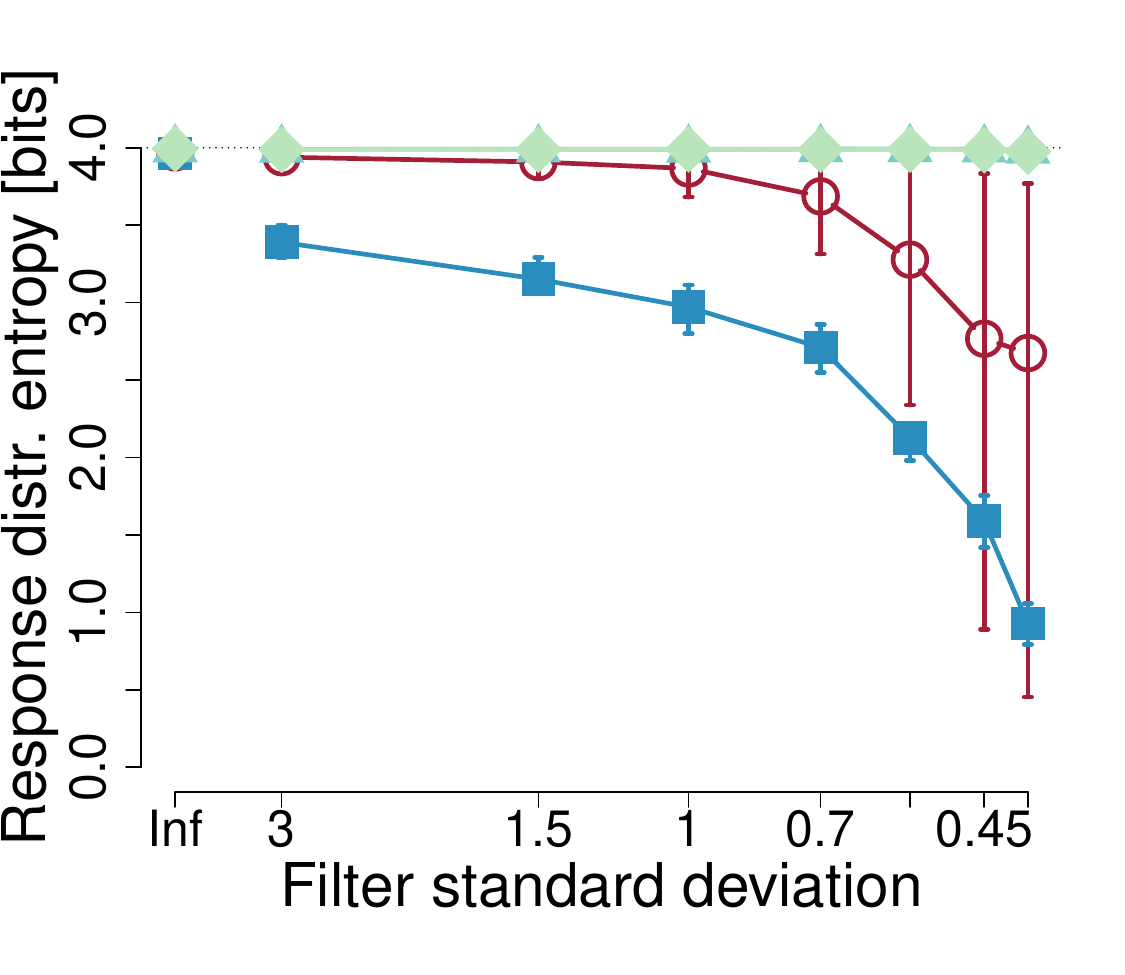}
        \vspace{\captionspace}
        \caption*{}
    \end{subfigure}\hfill
    \begin{subfigure}{\figwidth}
        \centering
        \includegraphics[width=\linewidth]{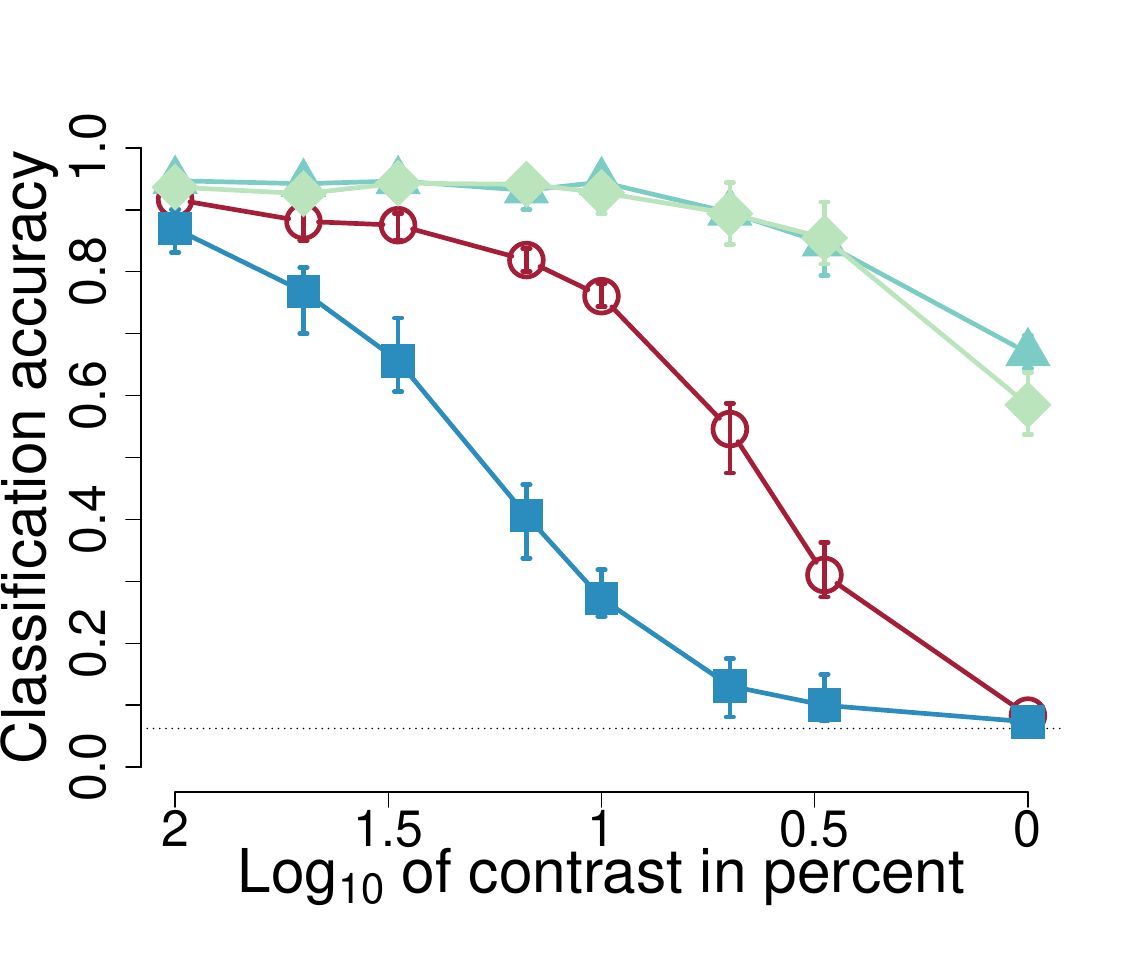}
        \vspace{\captionspace}
        \caption{Contrast}
    \end{subfigure}\hfill
    \begin{subfigure}{\figwidth}
        \centering
        \includegraphics[width=\linewidth]{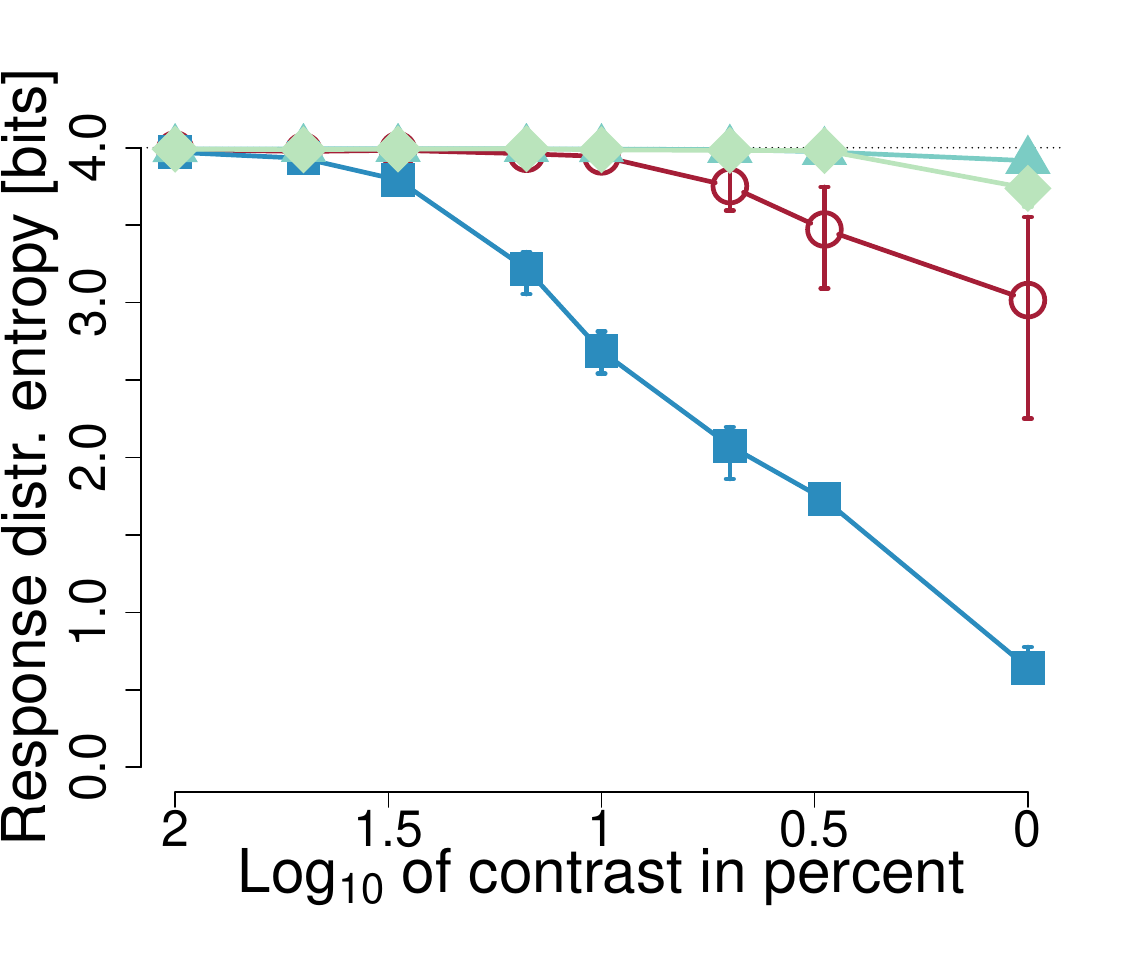}
        \vspace{\captionspace}
        \caption*{}
    \end{subfigure}\hfill

    \begin{subfigure}{\figwidth}
        \centering
        \includegraphics[width=\linewidth]{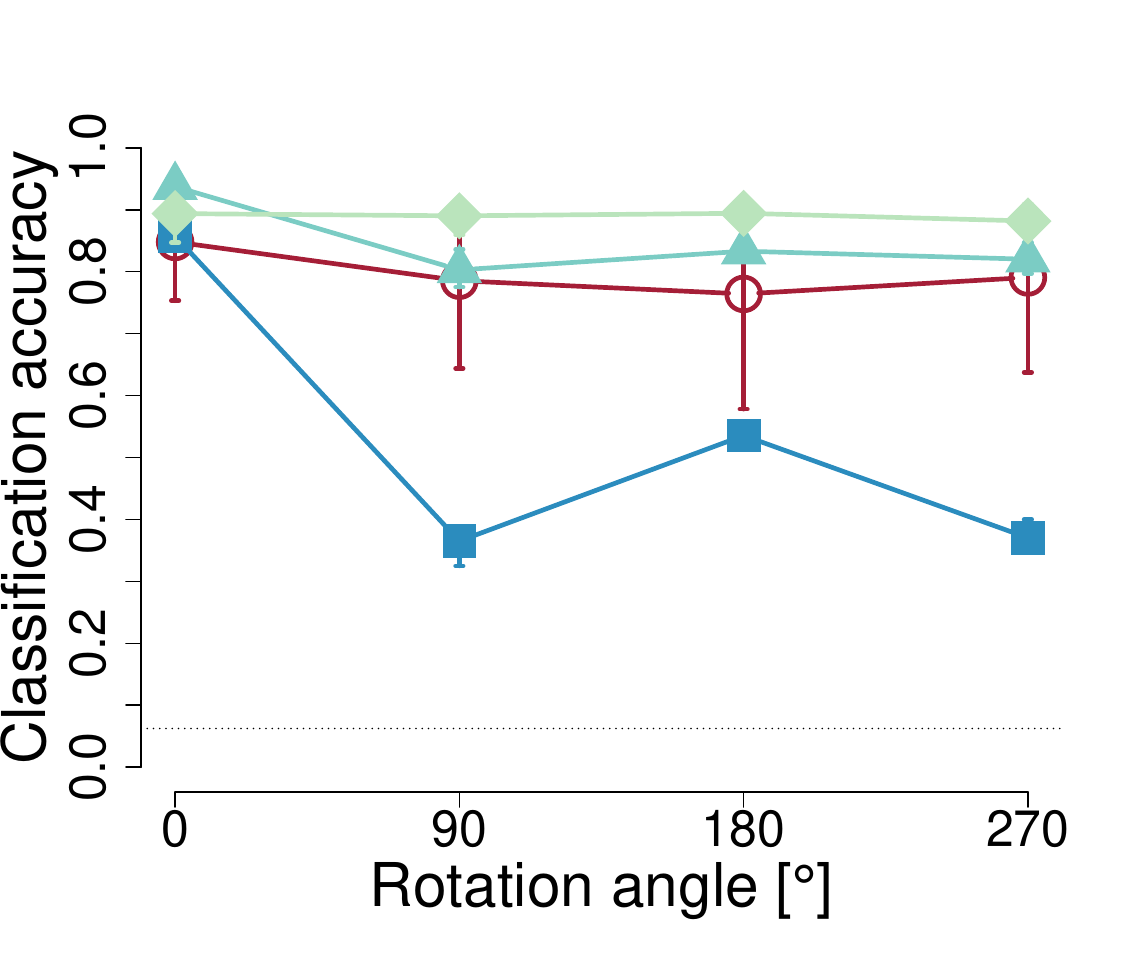}
        \vspace{\captionspace}
        \caption{Rotation}
    \end{subfigure}\hfill
    \begin{subfigure}{\figwidth}
        \centering
        \includegraphics[width=\linewidth]{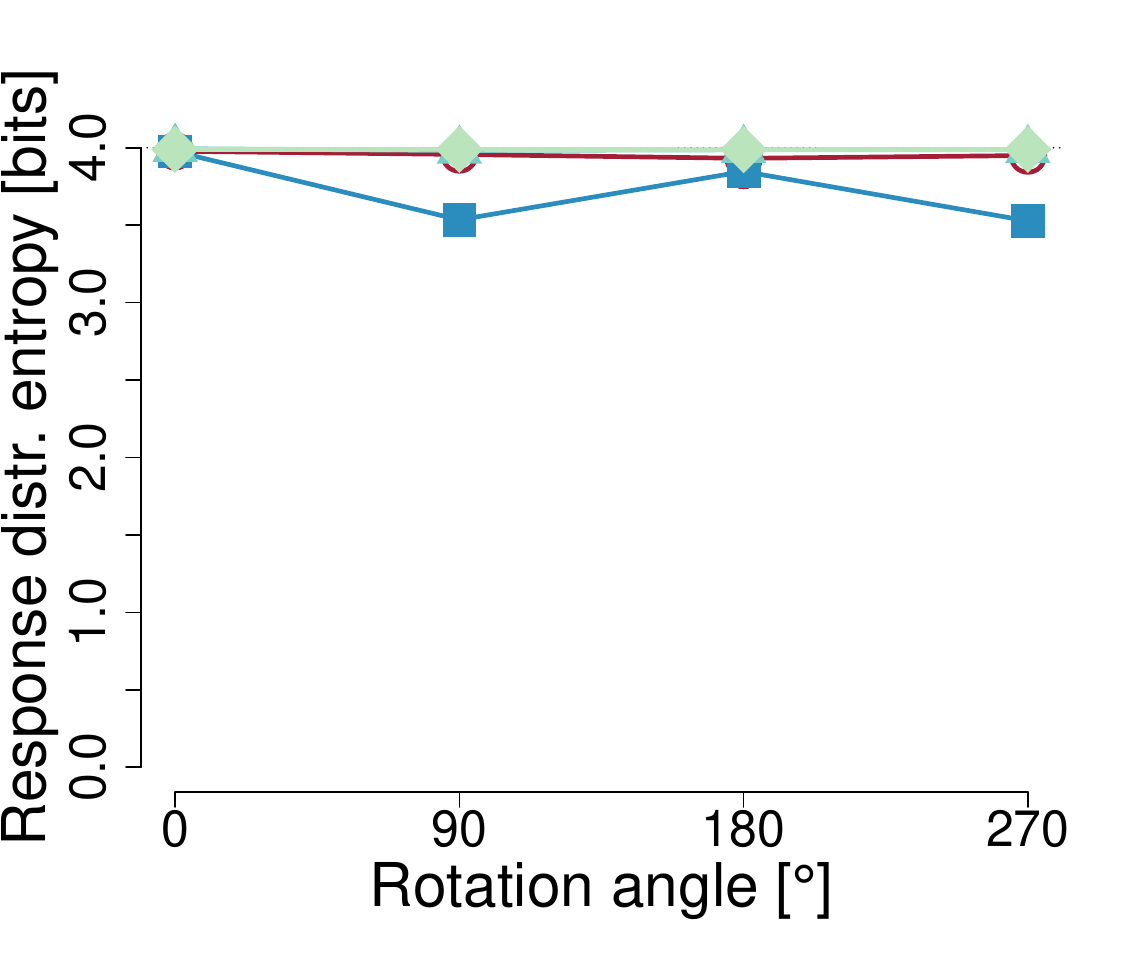}
        \vspace{\captionspace}
        \caption*{}
    \end{subfigure}\hfill
    \begin{subfigure}{\figwidth}
        \centering
        \includegraphics[width=\linewidth]{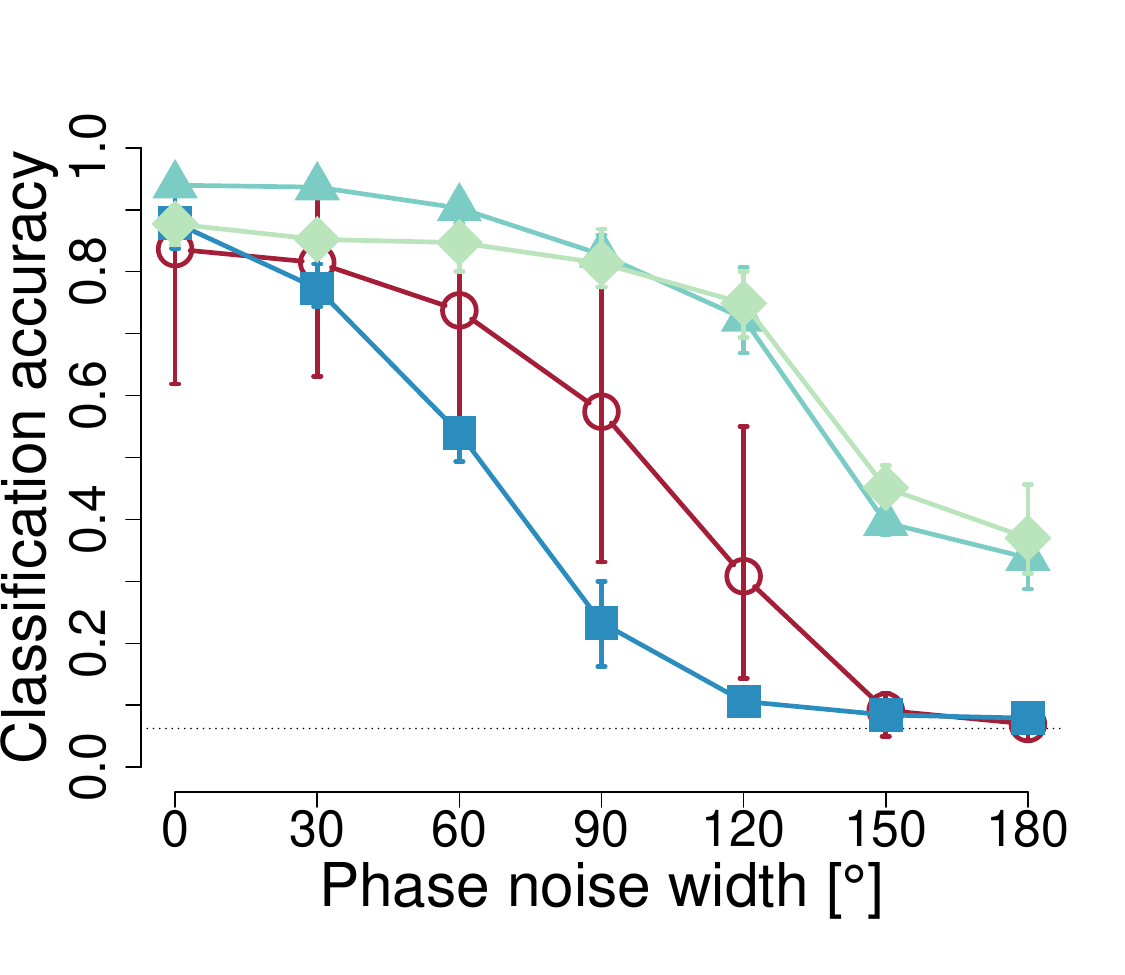}
        \vspace{\captionspace}
        \caption{Phase noise}
    \end{subfigure}\hfill
    \begin{subfigure}{\figwidth}
        \centering
        \includegraphics[width=\linewidth]{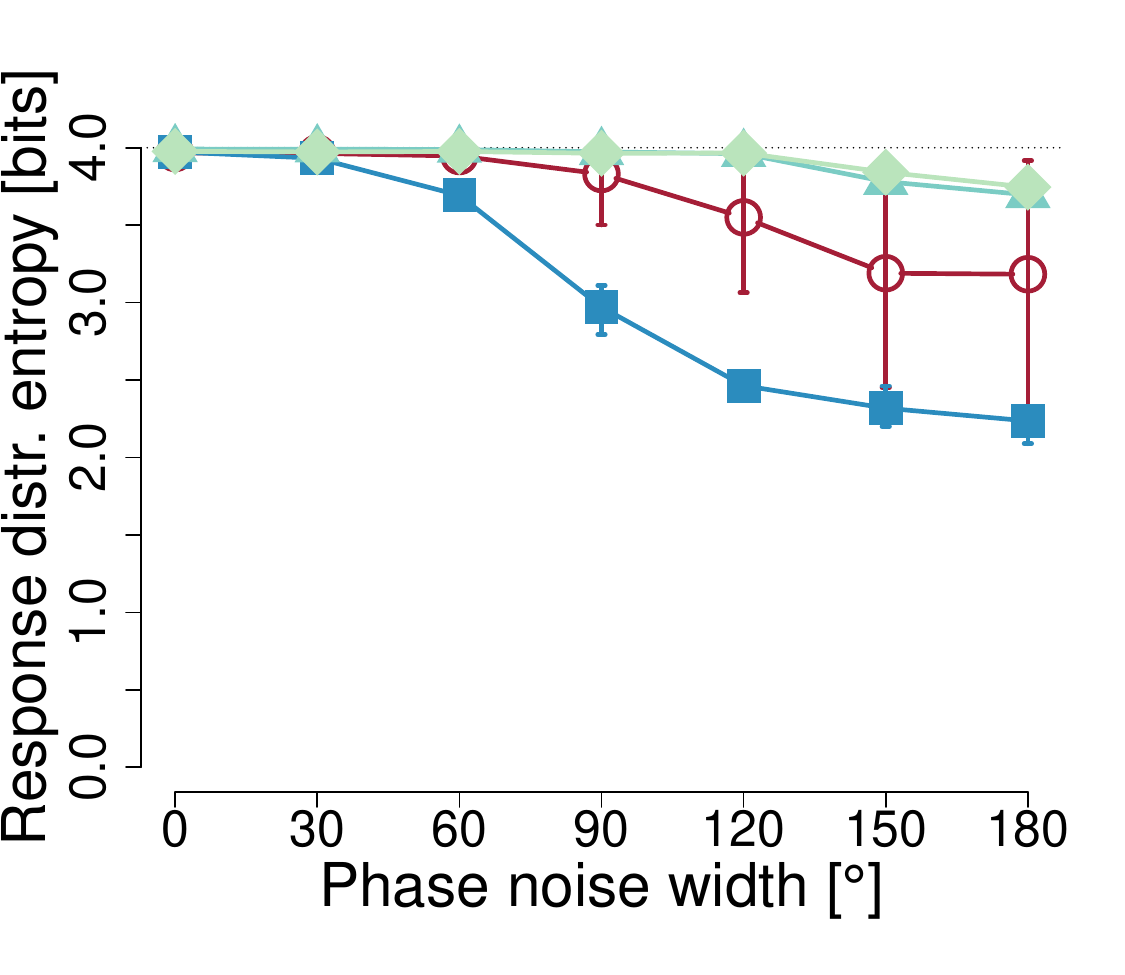}
        \vspace{\captionspace}
        \caption*{}
    \end{subfigure}\hfill
    \caption{Classification accuracy and response distribution entropy for \textcolor{human.100}{human observers}, \textcolor{resnet50.100}{ResNet-50} as well as an \textcolor{all.distortions.net.100}{All-Distortions-Net} and a \textcolor{specialised.net.100}{Specialised-Net}. All networks are trained from scratch; the Specialised-Net in every plot is trained on a single distortion (models A1 to A9 in Figure~\ref{fig:results_training}) whereas the All-Distortions-Net is trained on a number of distortions simultaneously. This corresponds to models C1 and C2 in Figure~\ref{fig:results_training}: for subplot \ref{sup:training_salt_and_pepper_noise}, salt-and-pepper noise, performance of model C1 is shown. For subplot \ref{sup:training_uniform_noise}, uniform noise, performance of C2 is shown. For all other plots, performance of the All-Distortions-Net is shown as the mean of performance for models C1 and C2.}
    \label{fig:training_accuracy_entropy}
\end{figure}

\subsection*{Error bars \& entropy}
When showing \emph{accuracy} in any of the plots, the error bars provided report the range of the data observed for different observers (not the often shown S.E. of the means, which would be much smaller). To produce a comparable measure of uncertainty for the DNNs, we computed seven runs with different subsets of the data, with each run consisting of the same number of images per category and condition that a single human observer was exposed to and report the range of accuracies observed in these runs. Seven runs are the maximum possible number of runs without ever showing an image to a DNN more than once per experiment.

For all response distribution entropy results (Figures~\ref{fig:results_accuracy_entropy} and \ref{fig:training_accuracy_entropy}), we calculated the entropy as the average of individual participants' entropies: otherwise, if the entropy was calculated over the aggregated human trials, individual differences might cancel each other out, which would lead to a higher human response distribution entropy.

\subsection*{Prediction uncertainty}
\label{sup:prediction_uncertainty}
Figure \ref{fig:uncertainty} shows the entropy of the networks' predictions over the 1000 ILSVRC12 classes as a measure of the networks' `uncertainty'. In principle, the more uncertain a network is in its predictions the more evenly it will distribute its softmax activations between classes and thus the higher the entropy will be. 
For all experiments, uncertainty roughly increases with distortion strength as reported by previous studies \cite{Dodge2016, Vasiljevic2016}. However, for uniform noise and the eidolon distortions all networks become more certain again for some higher distortion levels. Furthermore, ResNet-152 also becomes more certain for stronger distortions in the low-pass experiment and is consistently more confident in its predictions than GoogleNet and VGG-19. 
Thus there are distortions for which all networks and especially ResNet-152 fail to capture that the input signal is becoming worse. Instead they limit their predictions to only a few classes (cf. Figure \ref{fig:results_accuracy_entropy}) with high certainty. This result is in line with previous reports stating that uncertainty in standard discriminative DNNs is not well represented \cite[e.g.][]{Gal2016}.

\begin{figure}
    \begin{subfigure}{0.86\linewidth}
        \begin{subfigure}{0.493\linewidth}
            \centering
            \includegraphics[width=\linewidth]{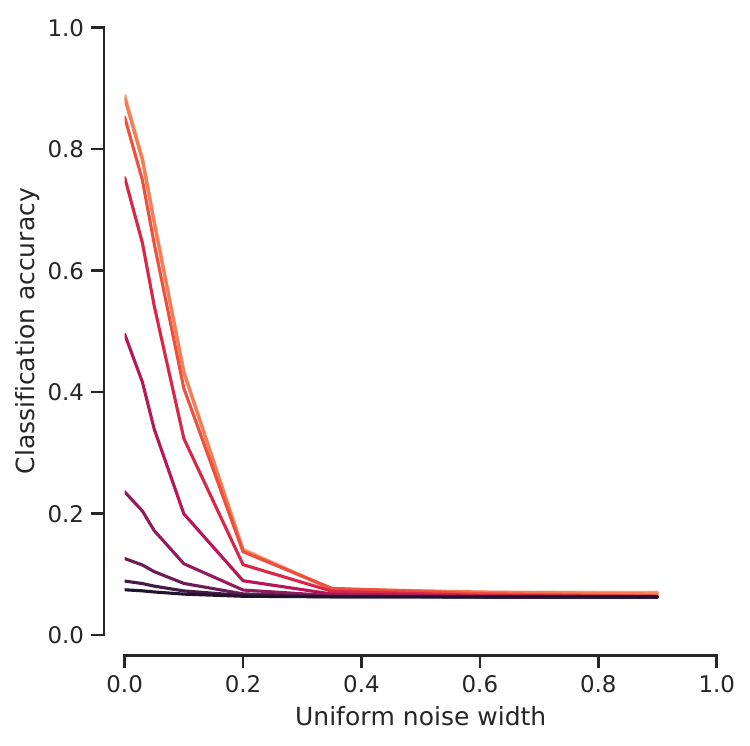}
            \vspace{\captionspaceII}
            \caption{Accuracy}
        \end{subfigure}\hfill
        \begin{subfigure}{0.493\linewidth}
            \centering
            \includegraphics[width=\linewidth]{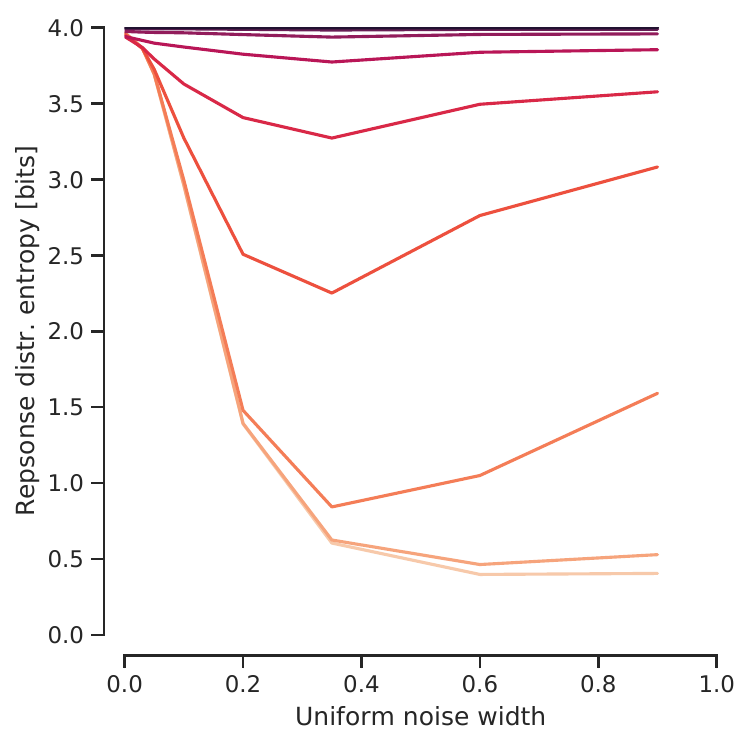}
            \vspace{\captionspaceII}
            \caption{Entropy}
        \end{subfigure}\hfill
        \vspace{1.2em}
        \begin{subfigure}{0.31\linewidth}
            \centering
            \includegraphics[width=\linewidth]{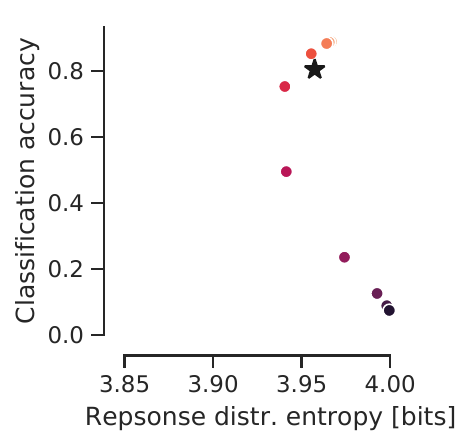}
            \vspace{\captionspaceII}
            \caption{Uniform noise width = 0.0}
        \end{subfigure}\hfill
        \begin{subfigure}{0.31\linewidth}
            \centering
            \includegraphics[width=\linewidth]{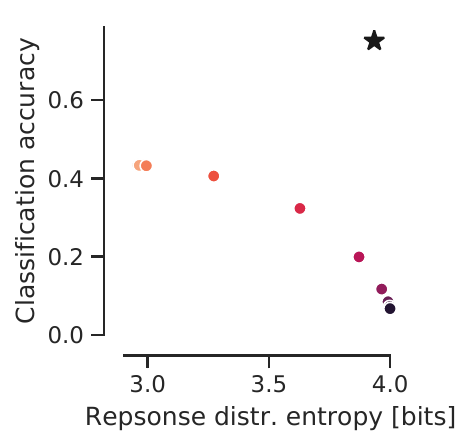}
            \vspace{\captionspaceII}
            \caption{Uniform noise width = 0.1}
        \end{subfigure}\hfill
        \begin{subfigure}{0.31\linewidth}
            \centering
            \includegraphics[width=\linewidth]{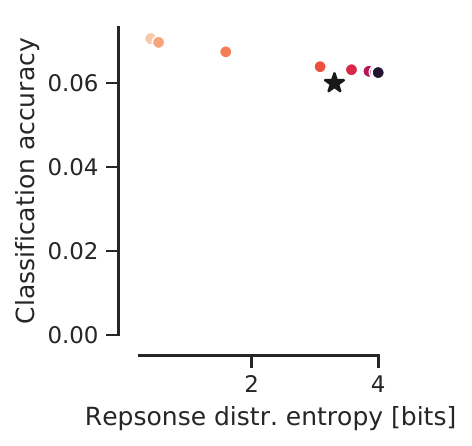}
            \vspace{\captionspaceII}
            \caption{Uniform noise width = 0.9}
        \end{subfigure}\hfill
        \vspace{1em}
    \end{subfigure}\hfill
    \begin{subfigure}{0.13\linewidth}
        \begin{subfigure}{\linewidth}
            \centering
            \includegraphics[width=\linewidth]{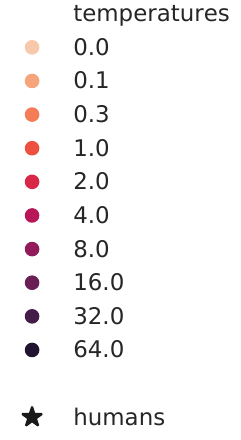}
        \end{subfigure}\hfill
    \end{subfigure}\hfill
    \caption{Classifcation accuracy \textbf{(a)} and response distribution entropy \textbf{(b)} as well as the trade-off between accuracy and entropy \textbf{(c, d, e)} for different softmax temperatures when the decision of a ResNet-50 model is sampled from its distribution over classes (softmax output) rather than taking the argmax of the distribution (which is equivalent to sampling with temperature $\to$ 0). While increasing the temperature does increase the response distribution entropy of ResNet-50, it simultaneously decreases the classification accuracy. For uniform noise with a width of 0.1 (d), increasing the temperature to match the response distribution entropy of humans reduces the accuracy of ResNet-50 below 0.1 whereas human accuracy is at 0.75.}
    \label{fig:temperature_tradeoff}
\end{figure}

\begin{figure}
	\centering
	\captionsetup[subfigure]{labelformat=empty}
	\begin{subfigure}{0.01335\linewidth}
		\caption{Noise}
		\vspace{-0.35cm}
		\vspace*{37pt}
		0.0
		\hfill
		\vspace*{47pt}
		0.03
		\hfill
		\vspace*{47pt}
		0.05
		\hfill
		\vspace*{47pt}
		0.1
		\hfill
		\vspace*{47pt}
		0.2
		\hfill
		\vspace*{47pt}
		0.35
		\hfill
		\vspace*{47pt}
		0.6
		\hfill
		\vspace*{47pt}
		0.9	
		\vspace*{27pt}
	\end{subfigure}
	\hspace{14pt}
	\begin{subfigure}{0.16\linewidth}
		\caption{Humans}
		\vspace{-0.35cm}
		\begin{center}		
		\includegraphics[width=0.98\linewidth]{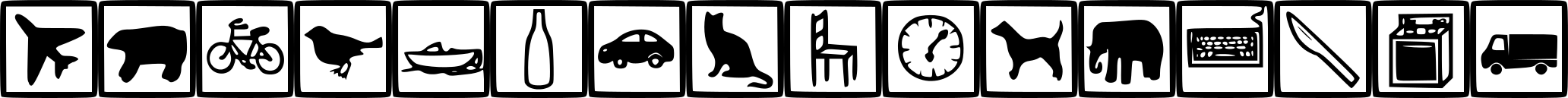}
		\hspace{10pt}
		\hfill
		\vspace{-11pt}
		\includegraphics[width=1.0\linewidth]{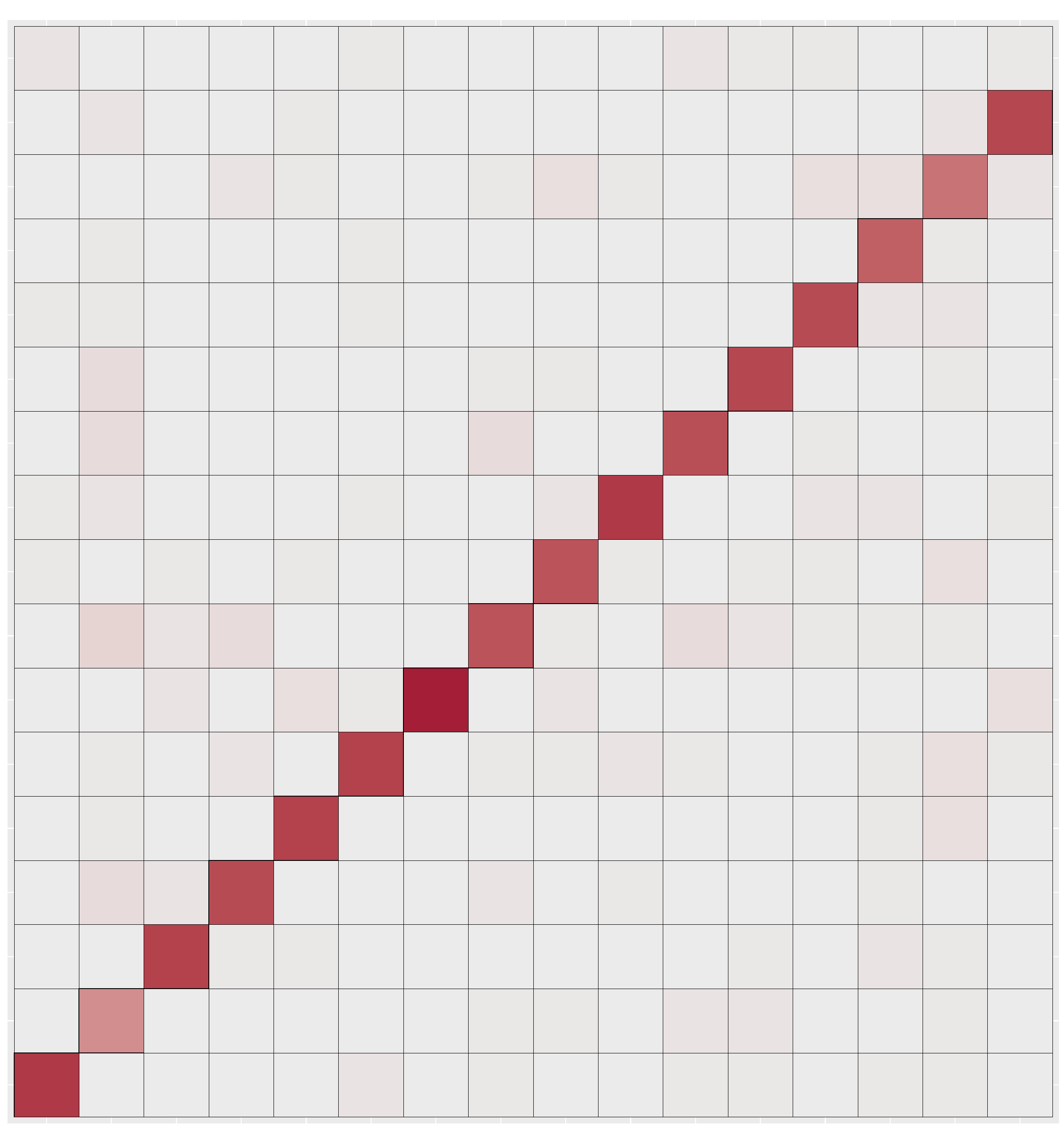}
		\hfill
		\includegraphics[width=1.0\linewidth]{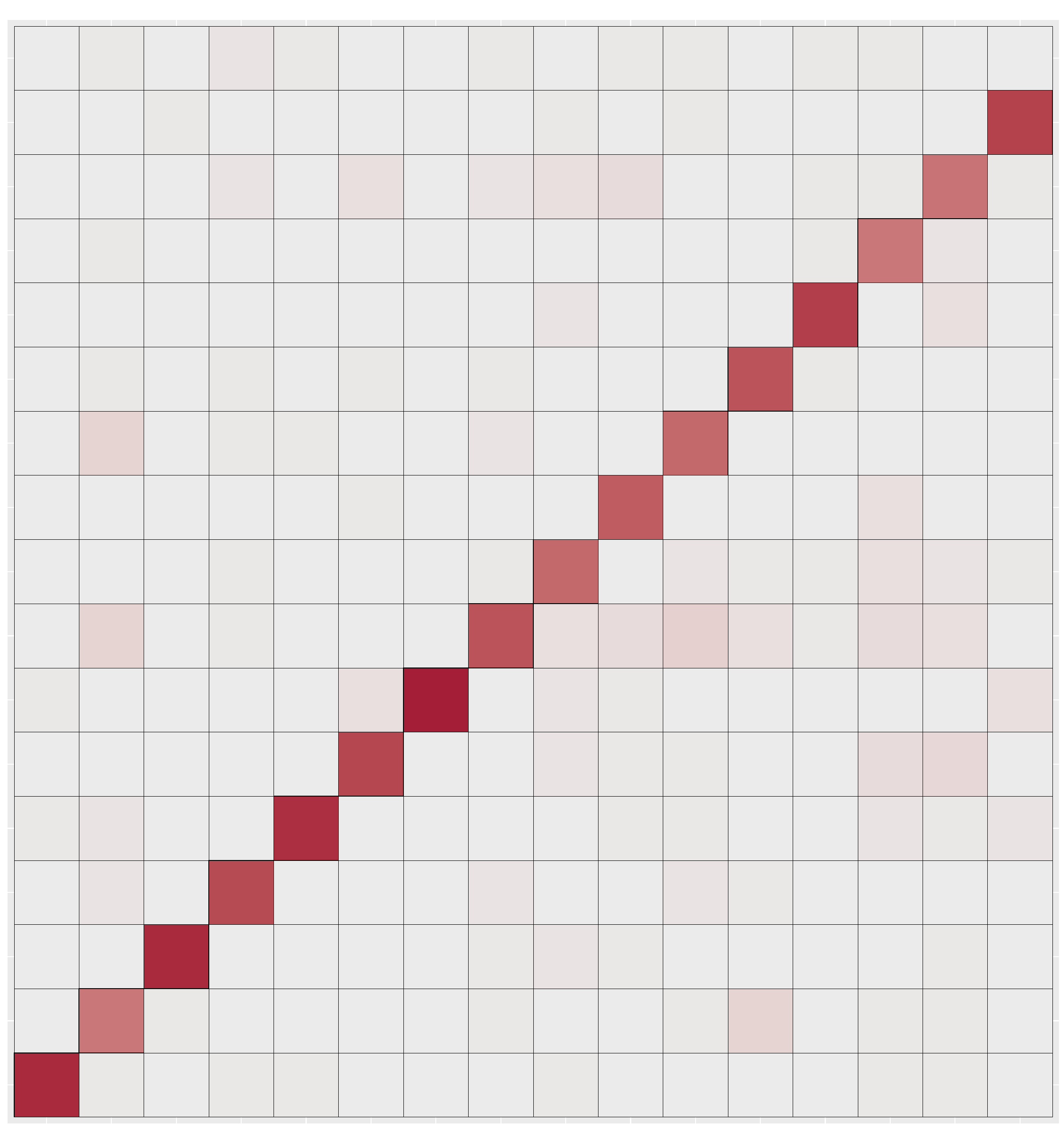}
		\hfill
		\includegraphics[width=1.0\linewidth]{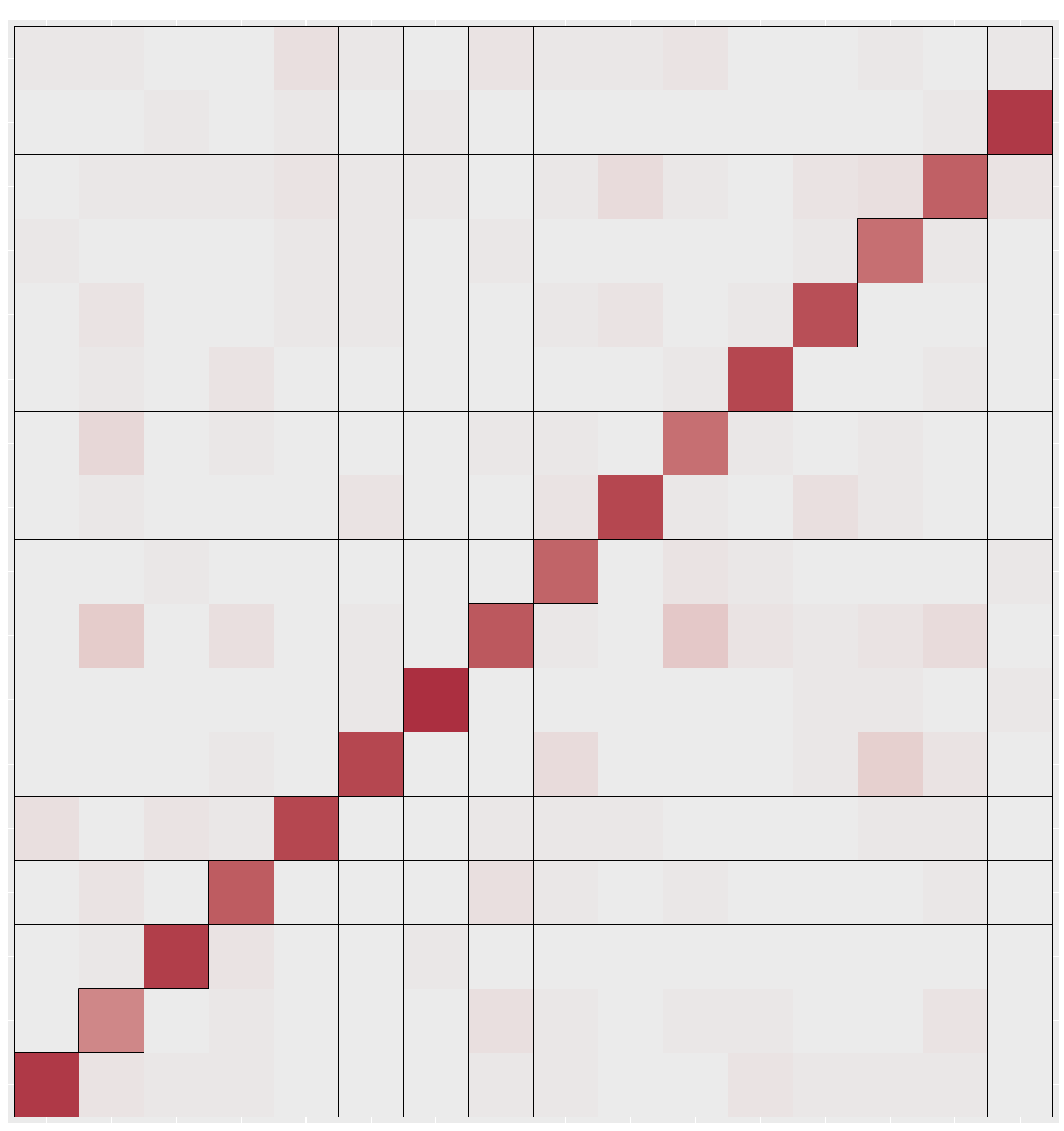}
		\hfill
		\includegraphics[width=1.0\linewidth]{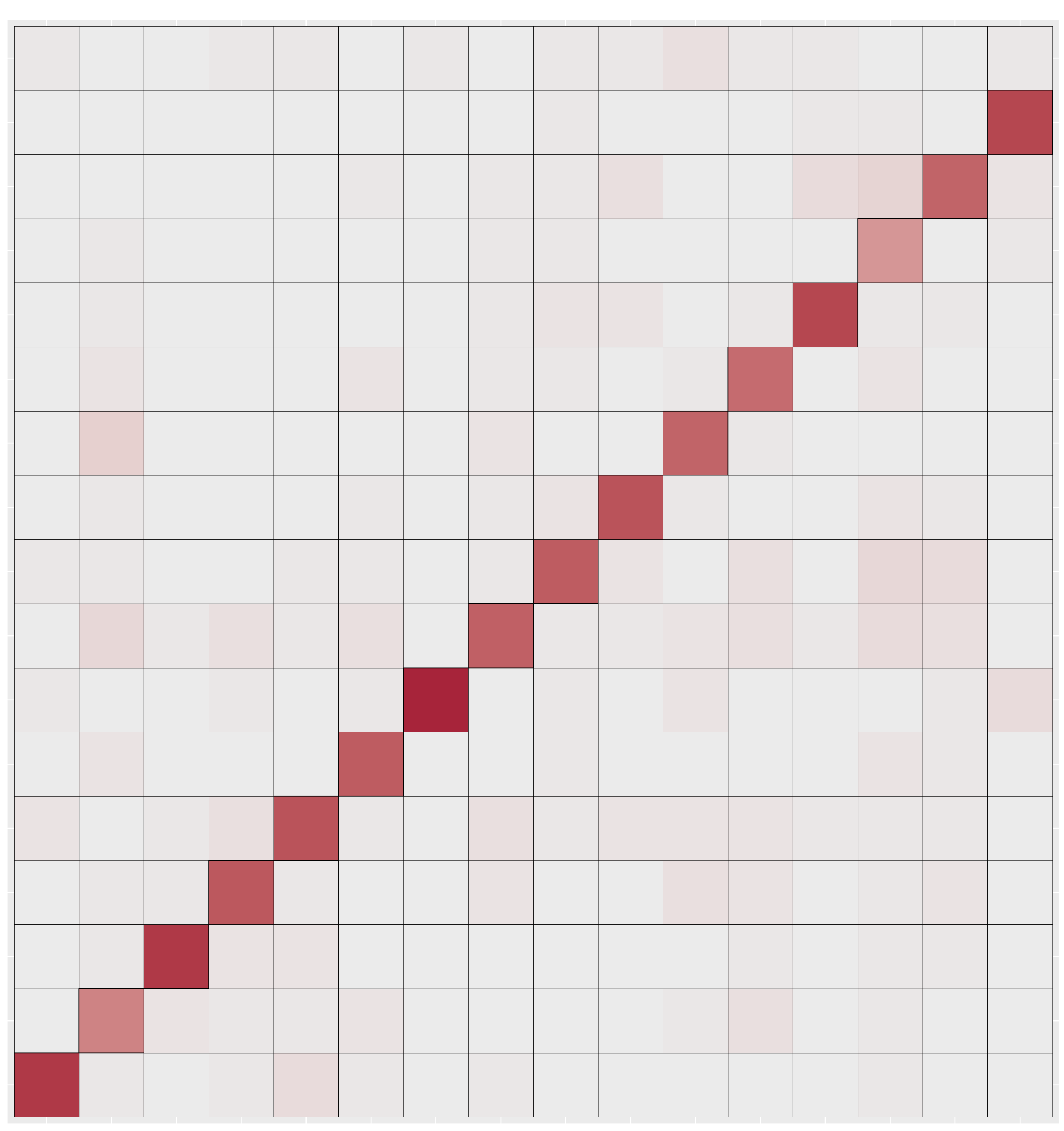}
		\hfill
		\includegraphics[width=1.0\linewidth]{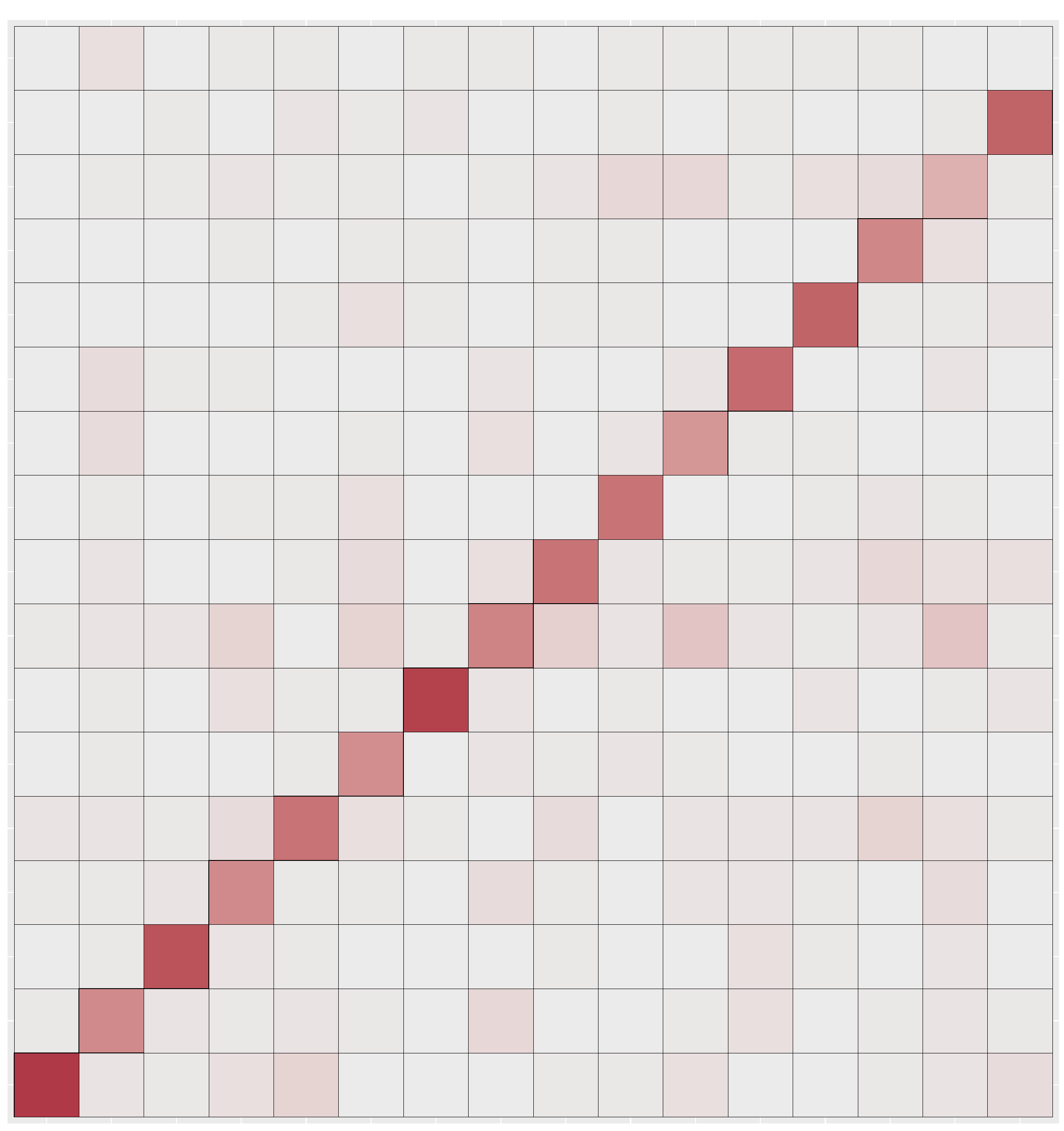}
		\hfill
		\includegraphics[width=1.0\linewidth]{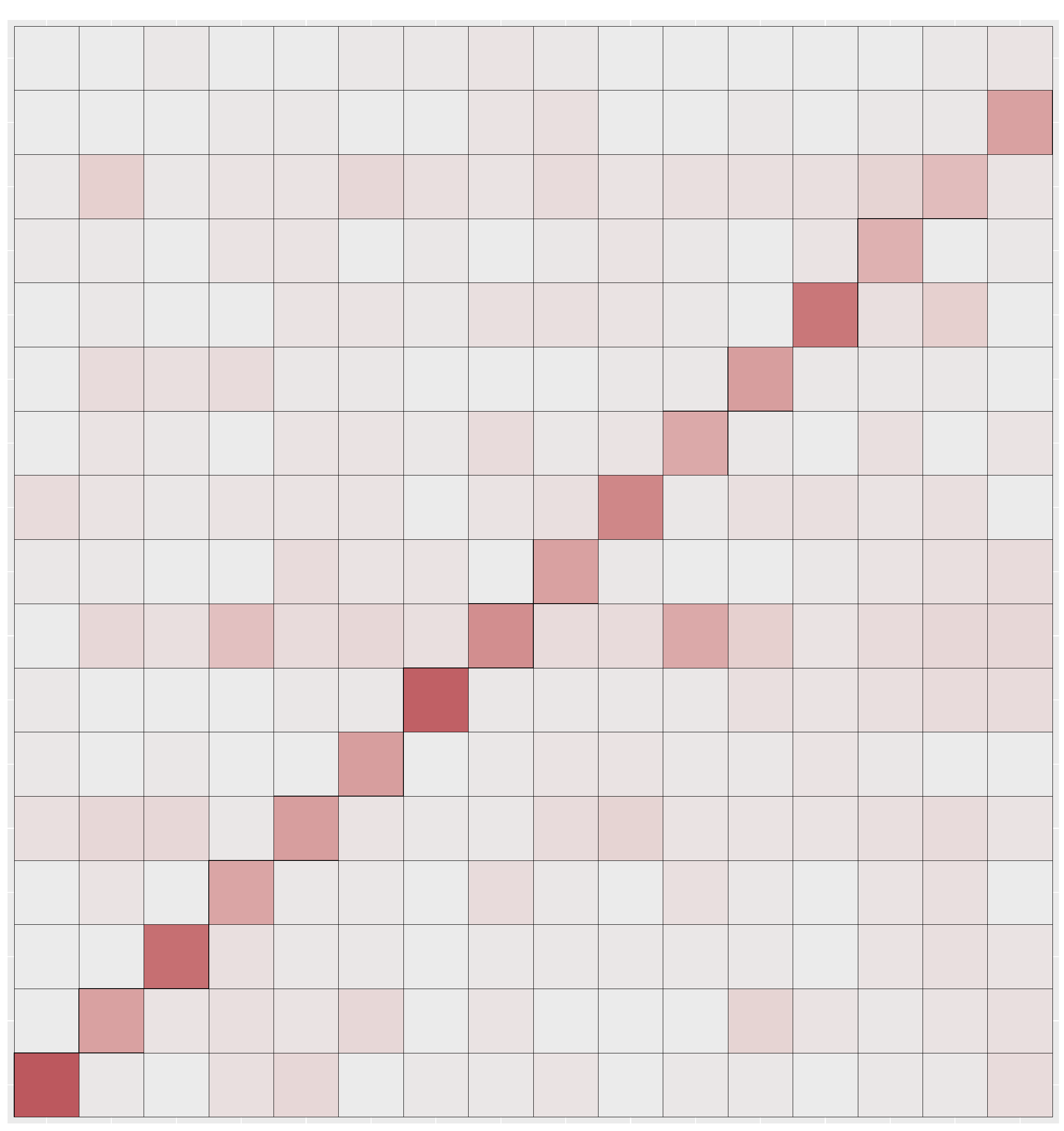}
		\hfill
		\includegraphics[width=1.0\linewidth]{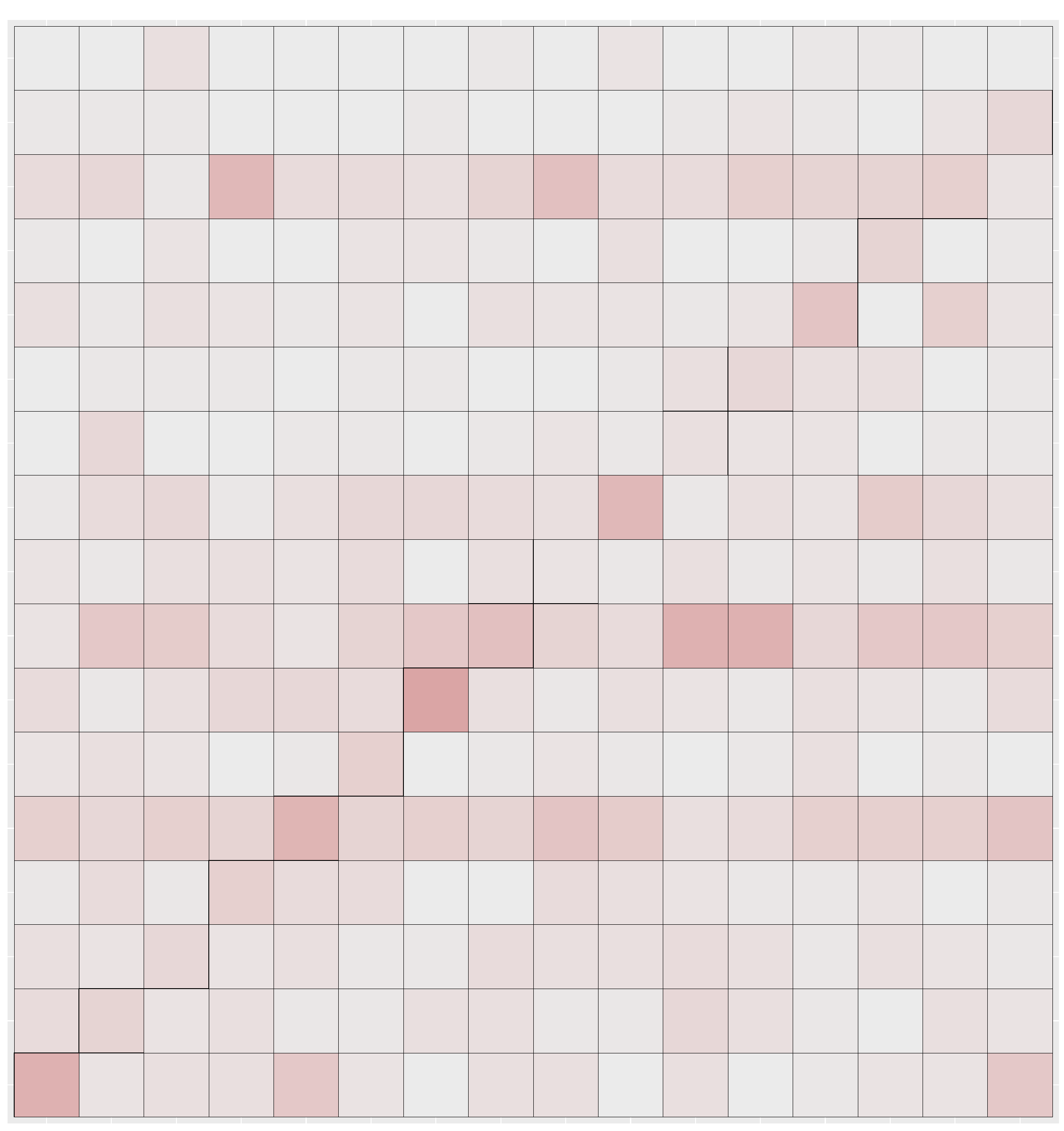}
		\hfill
		\includegraphics[width=1.0\linewidth]{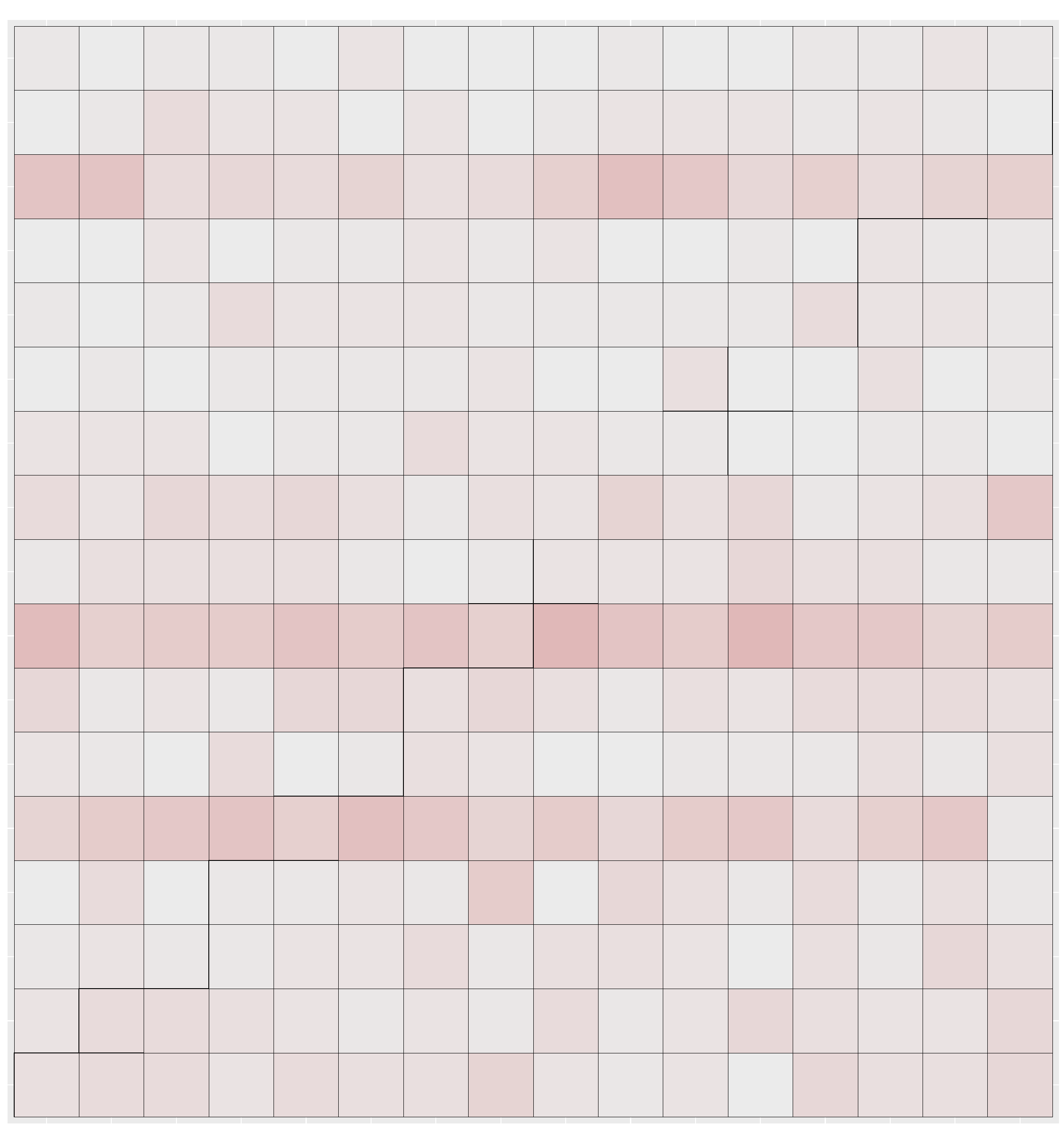}
		\hfill
		\vspace{-16pt}
		\includegraphics[width=0.98\linewidth]{confusion/response_screens_combination/response_icons_horizontal.png}
		\hfill
		\end{center}
	\end{subfigure}
	\hspace*{-5pt}
	\begin{subfigure}{0.16\linewidth}
		\caption{GoogLeNet}
		\vspace{-0.35cm}
		\begin{center}		
		\includegraphics[width=0.98\linewidth]{confusion/response_screens_combination/response_icons_horizontal.png}
		\hspace{2pt}
		\hfill
		\vspace{-11pt}
		\includegraphics[width=1.0\linewidth]{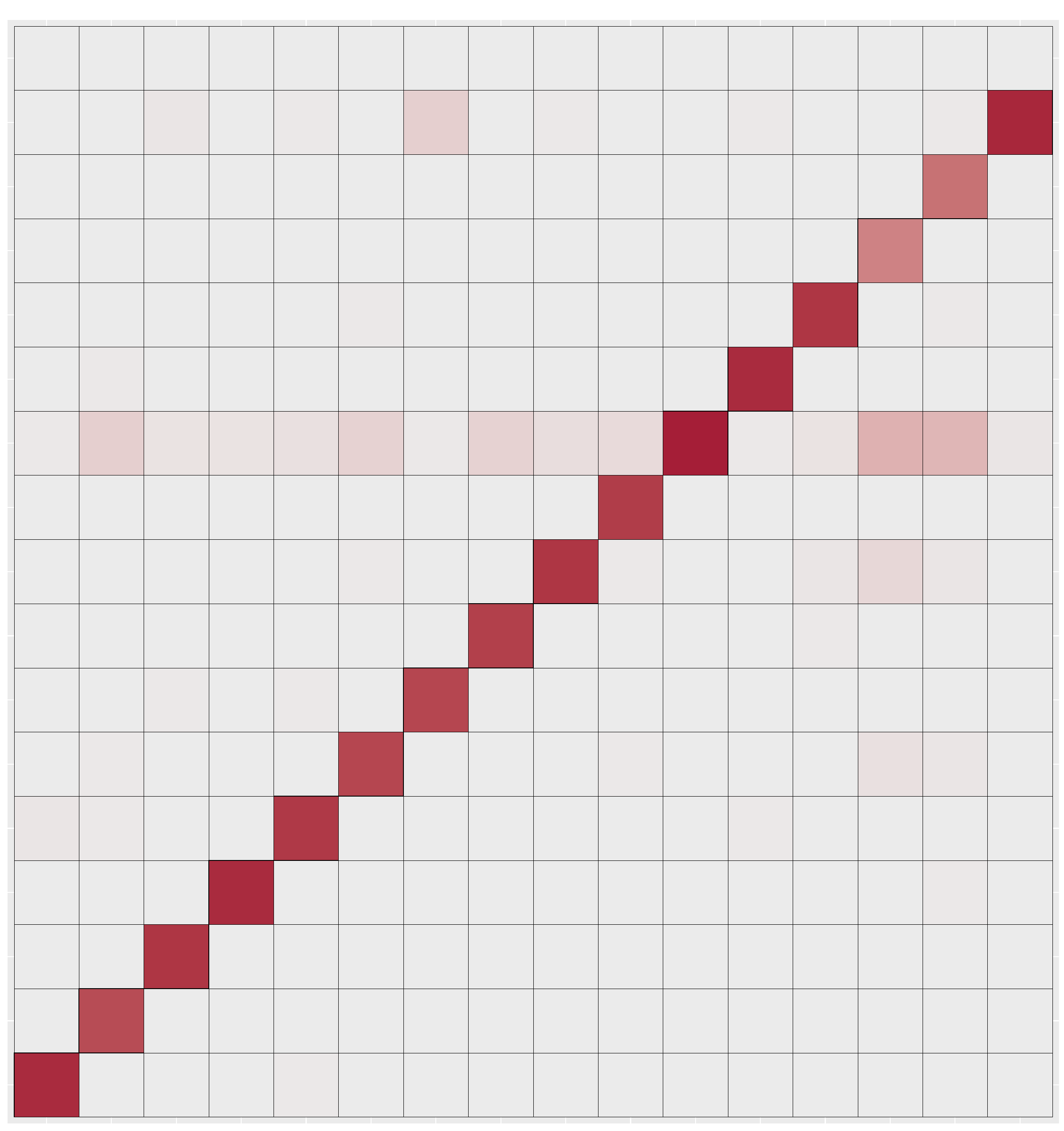}
		\hfill
		\includegraphics[width=1.0\linewidth]{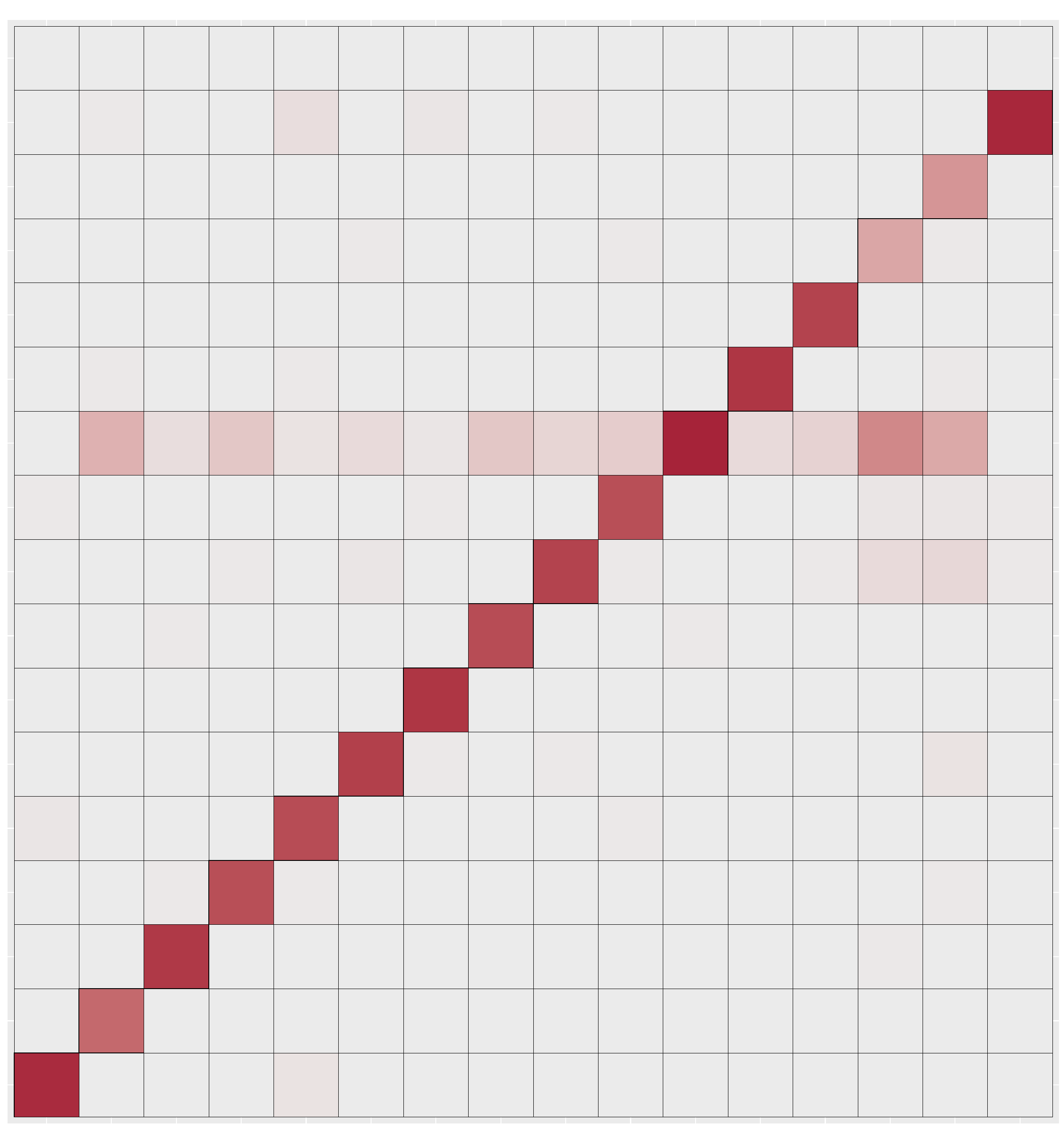}
		\hfill
		\includegraphics[width=1.0\linewidth]{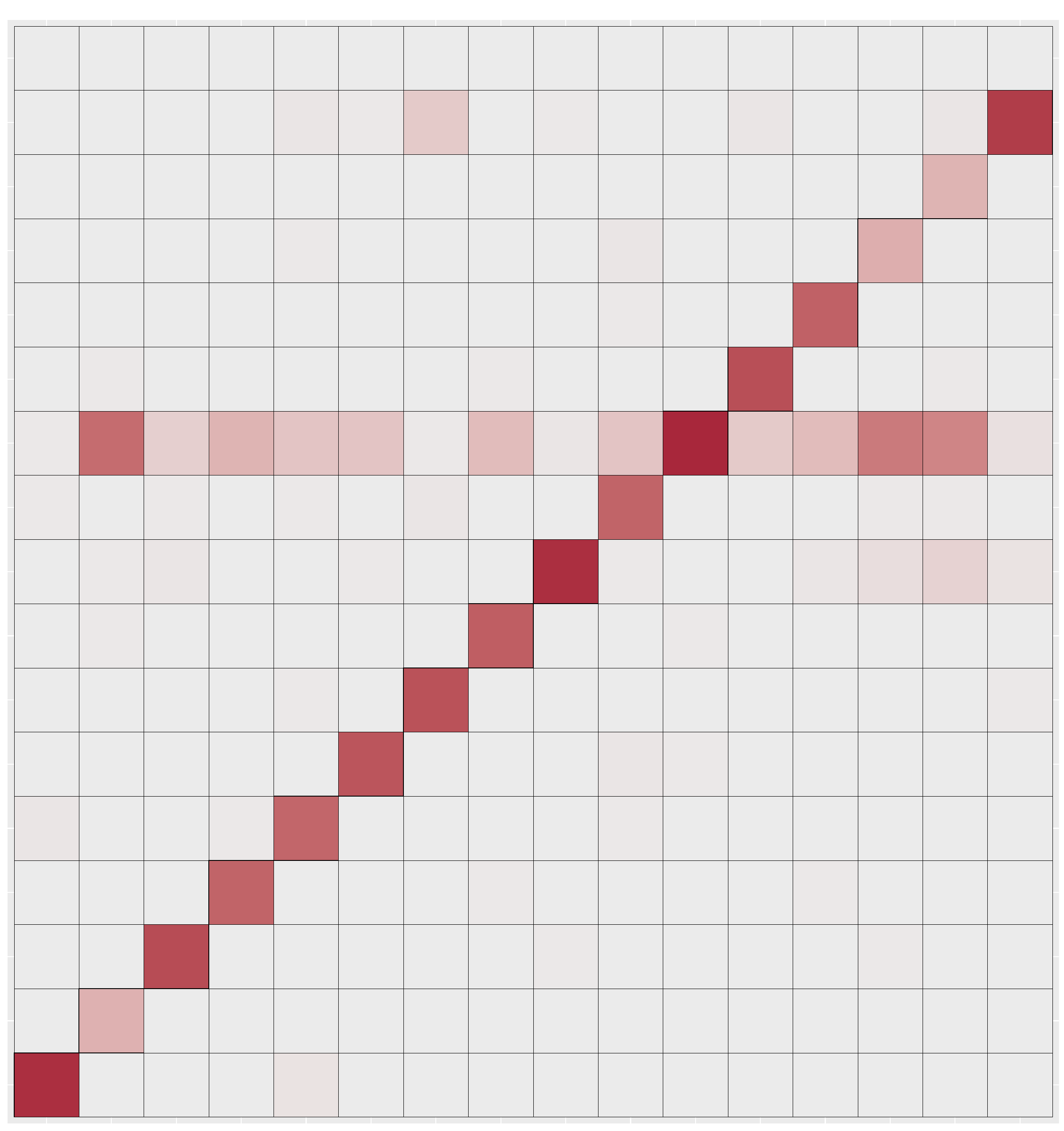}
		\hfill
		\includegraphics[width=1.0\linewidth]{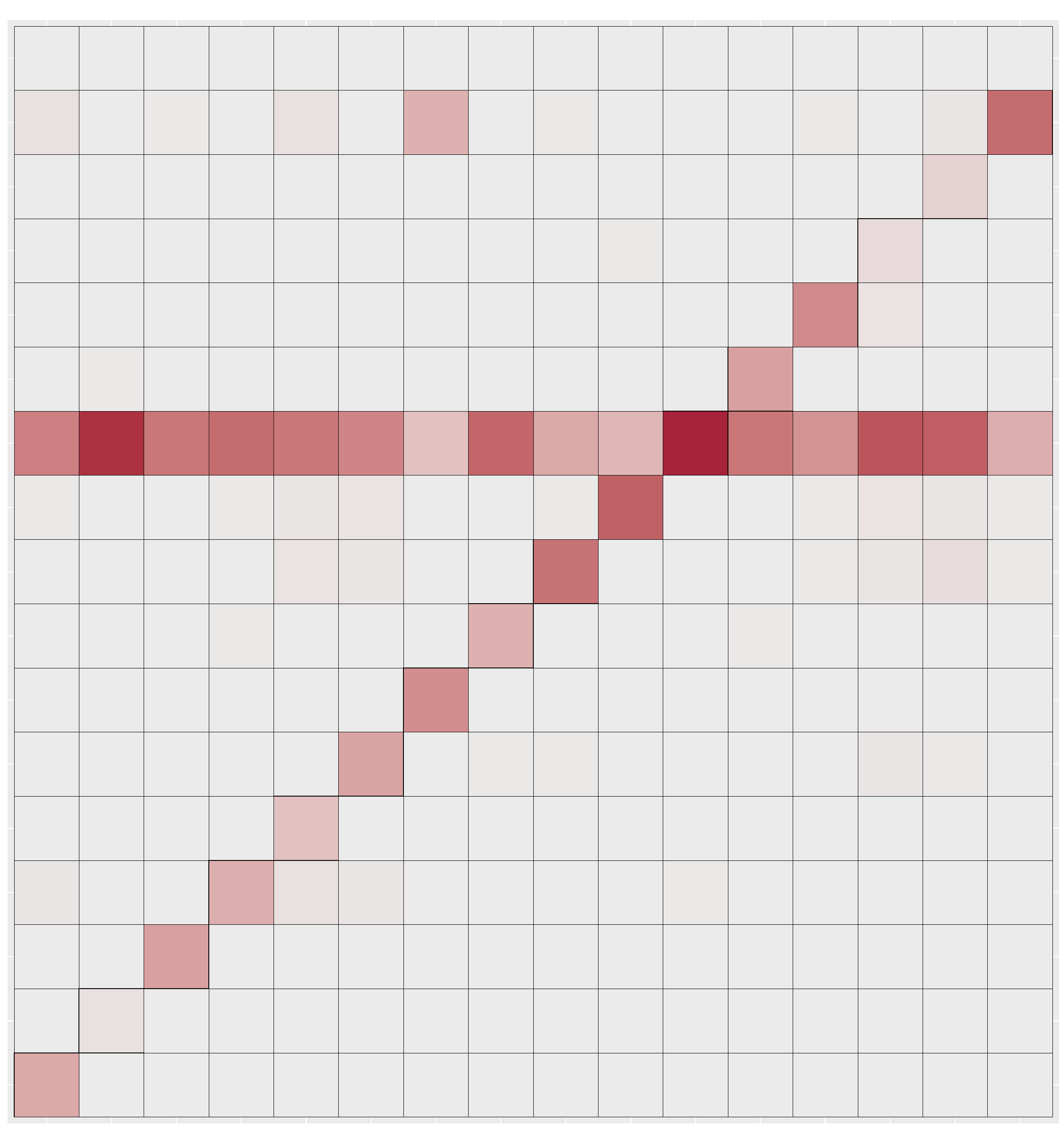}
		\hfill
		\includegraphics[width=1.0\linewidth]{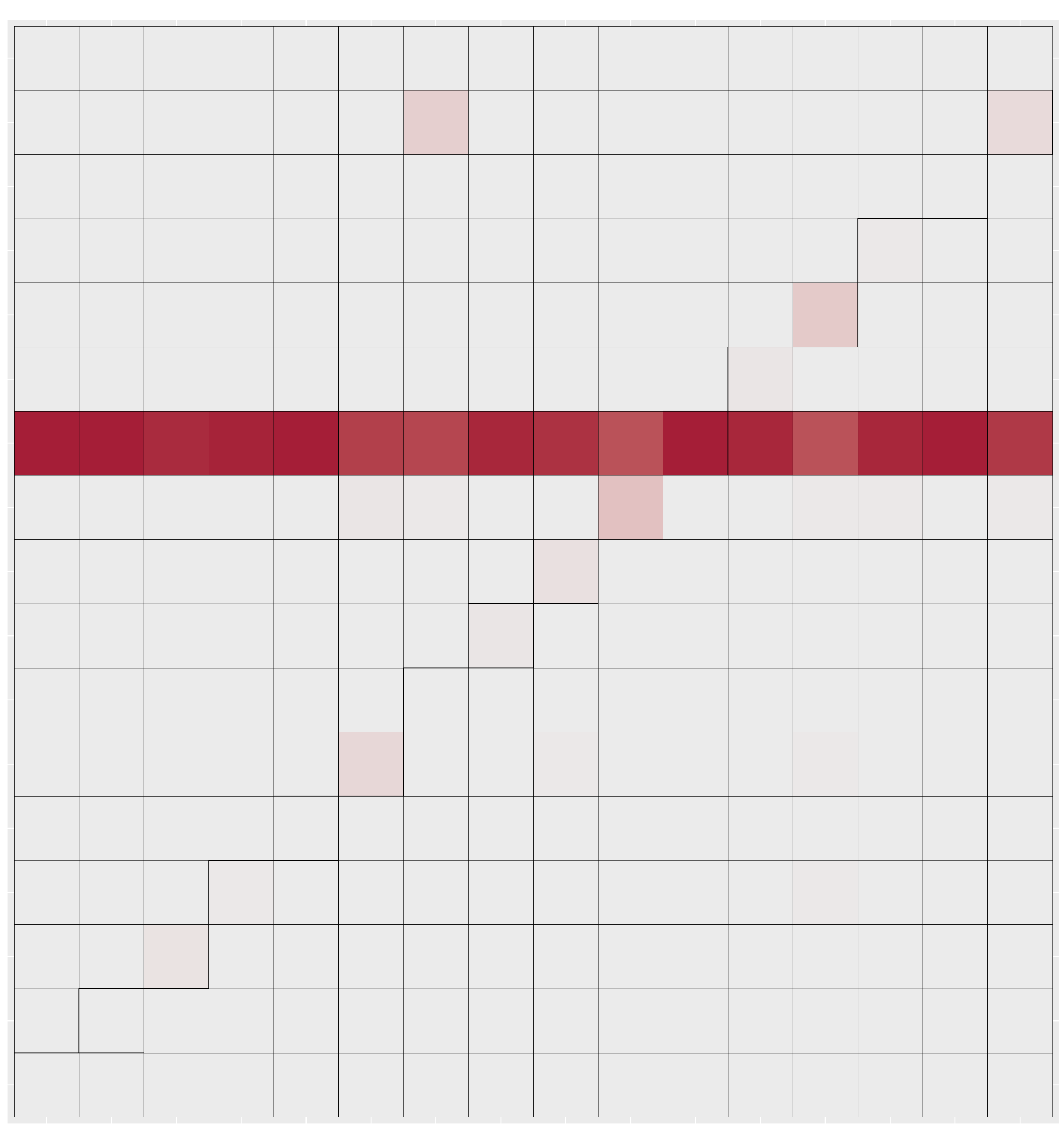}
		\hfill
		\includegraphics[width=1.0\linewidth]{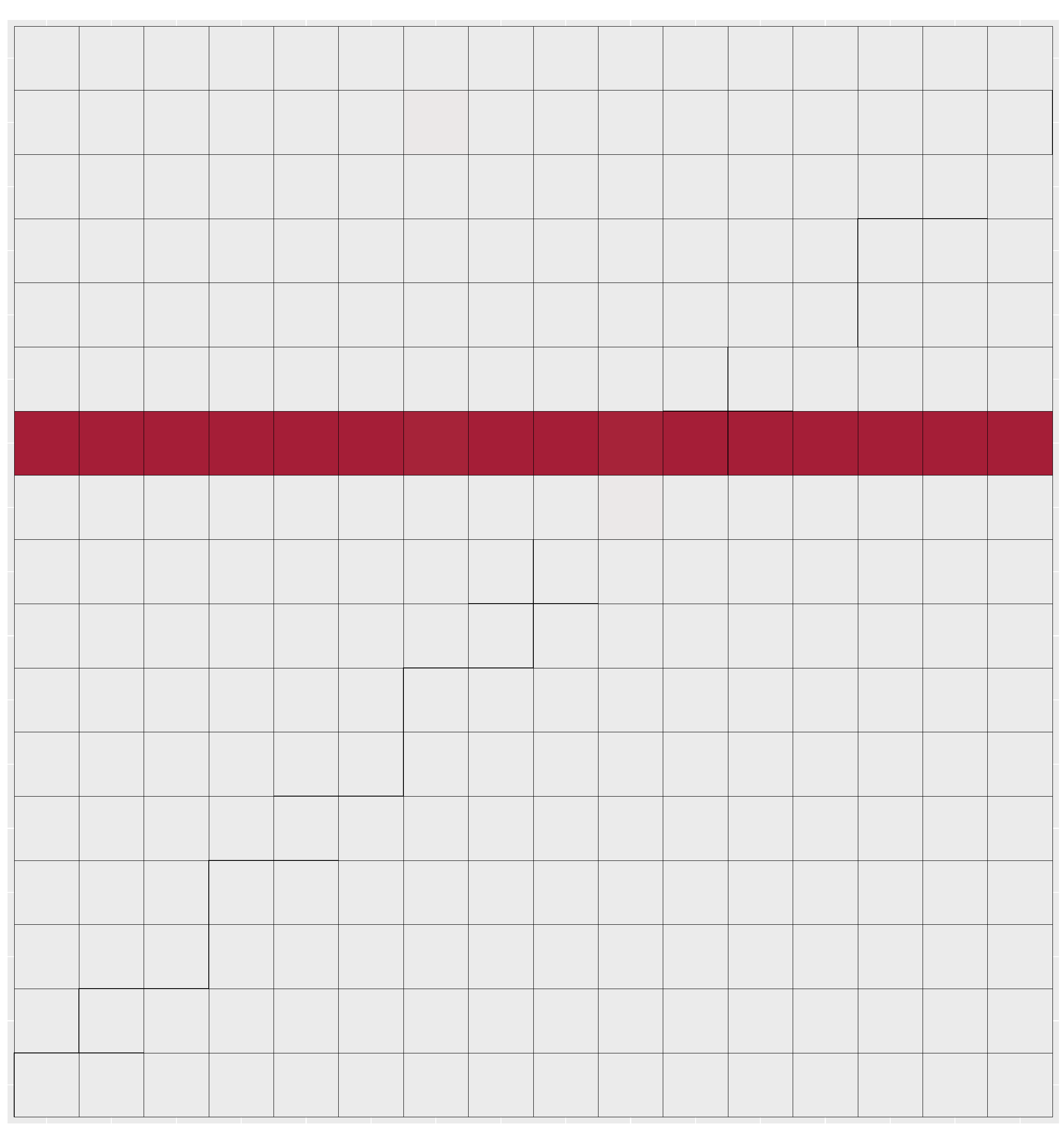}
		\hfill
		\includegraphics[width=1.0\linewidth]{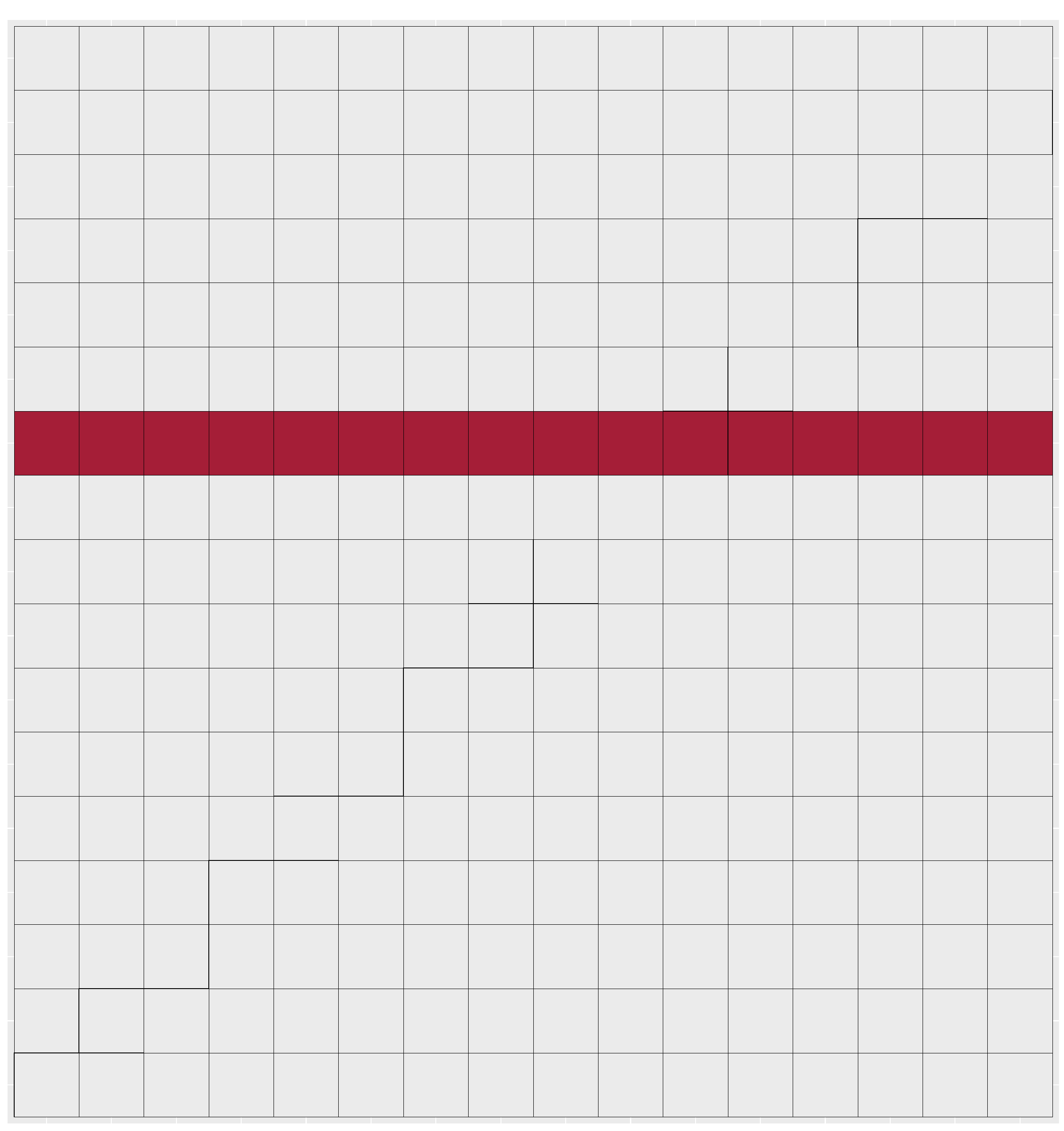}
		\hfill
		\includegraphics[width=1.0\linewidth]{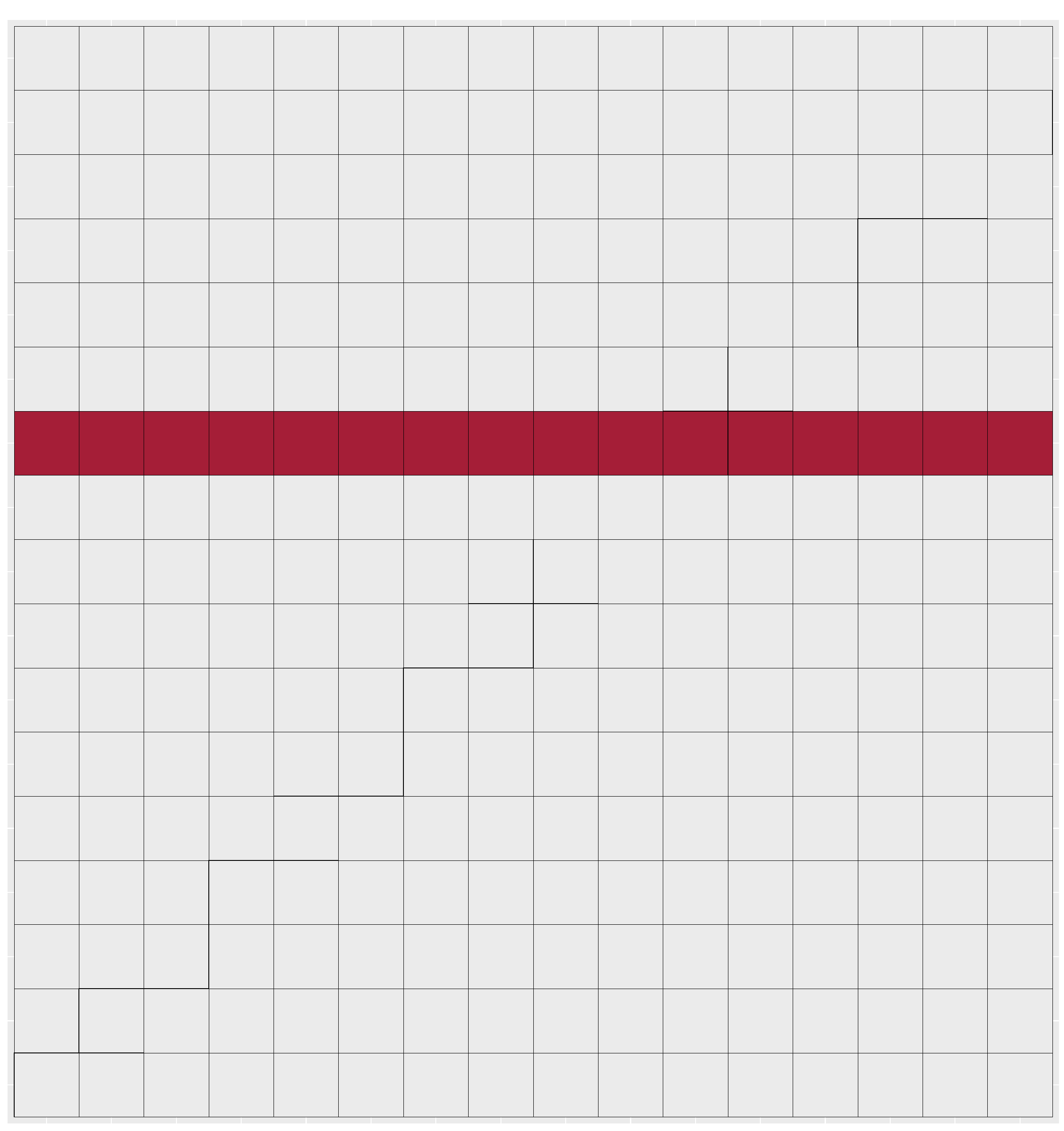}
		\hfill
		\vspace{-16pt}
		\includegraphics[width=0.98\linewidth]{confusion/response_screens_combination/response_icons_horizontal.png}
		\hfill
		\end{center}
	\end{subfigure}
	\hspace*{-5pt}
	\begin{subfigure}{0.16\linewidth}
		\caption{VGG-19}
		\vspace{-0.35cm}
		\begin{center}		
		\includegraphics[width=0.98\linewidth]{confusion/response_screens_combination/response_icons_horizontal.png}
		\hspace{2pt}
		\hfill
		\vspace{-11pt}
		\includegraphics[width=1.0\linewidth]{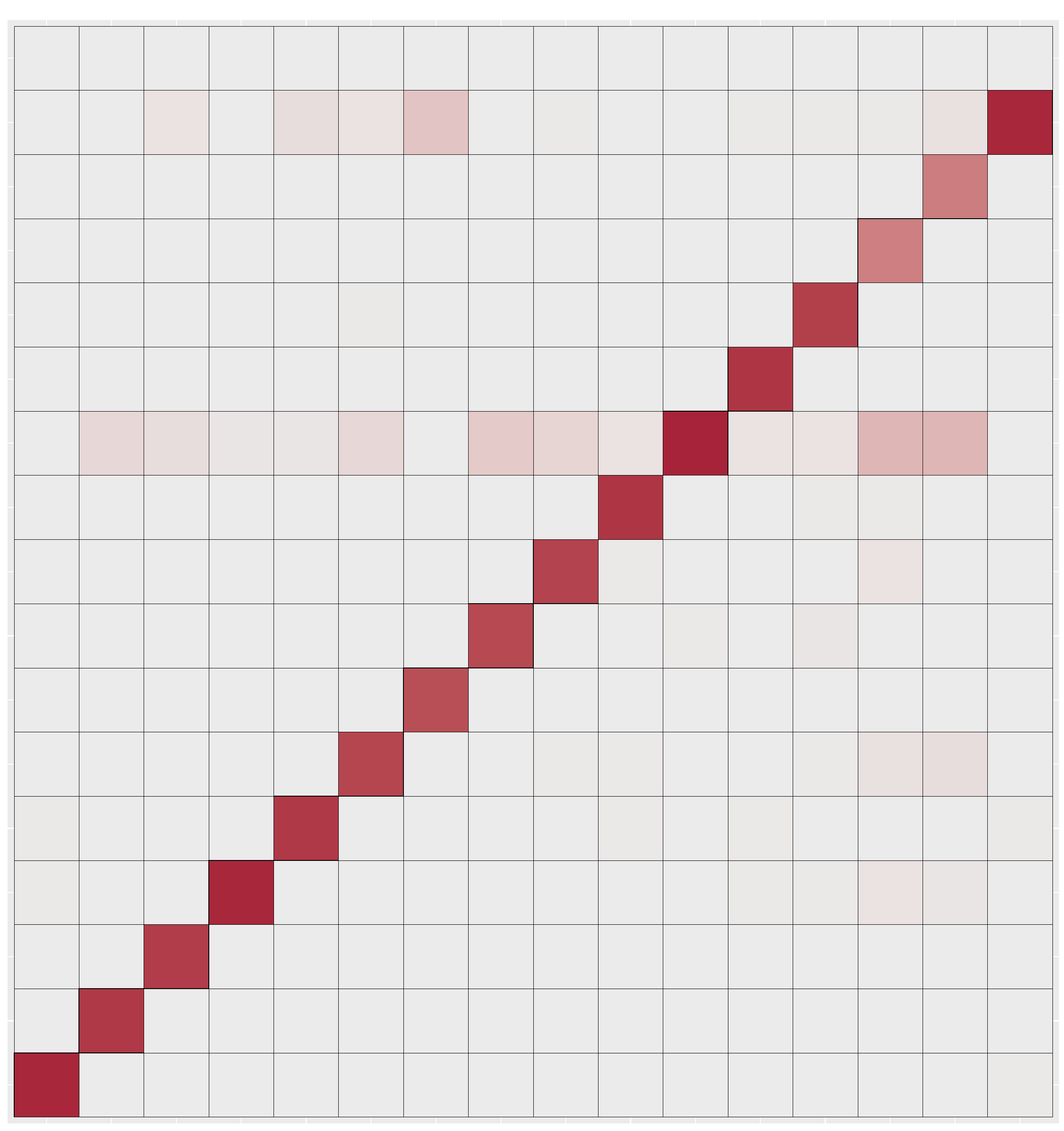}
		\hfill
		\includegraphics[width=1.0\linewidth]{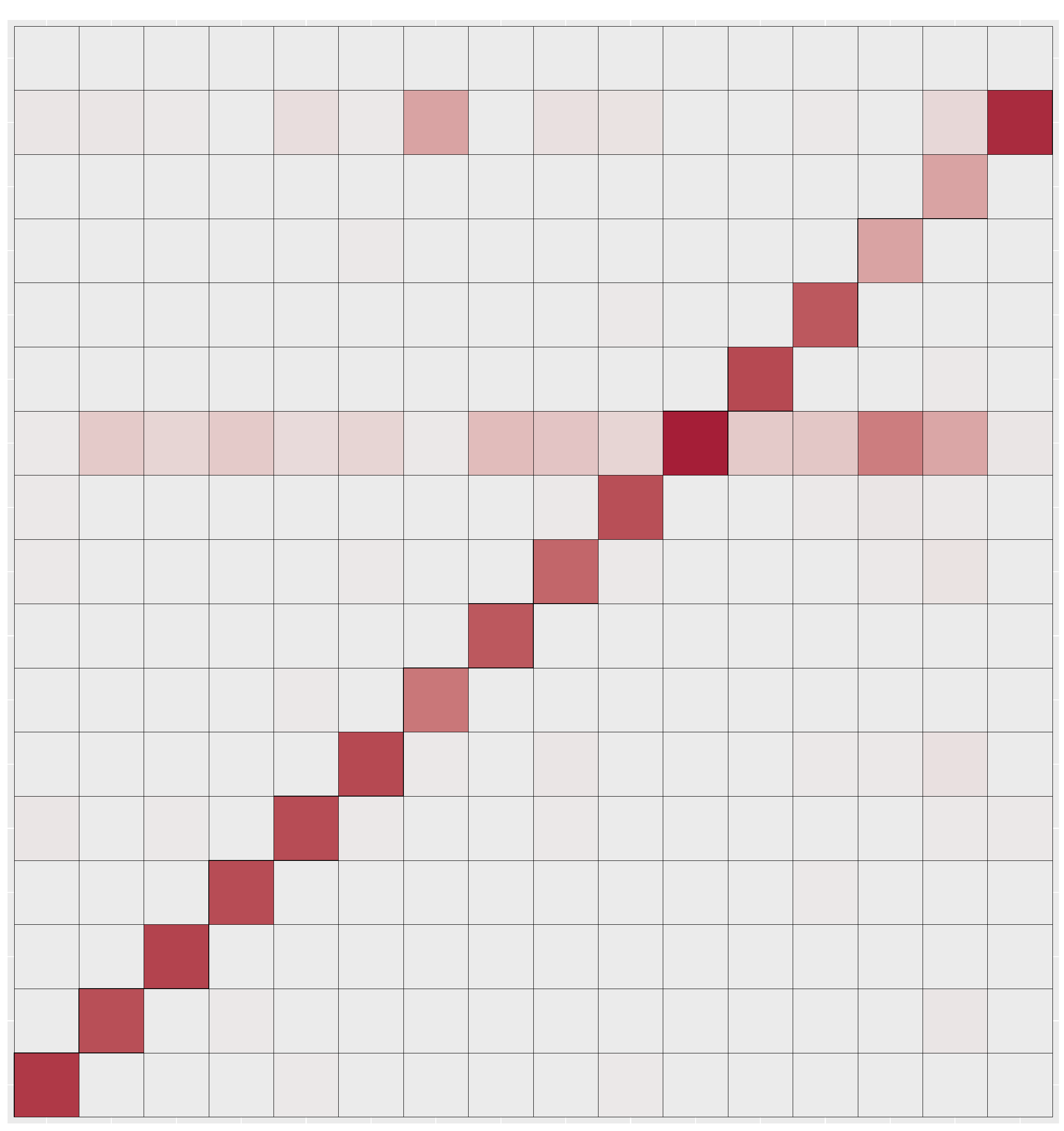}
		\hfill
		\includegraphics[width=1.0\linewidth]{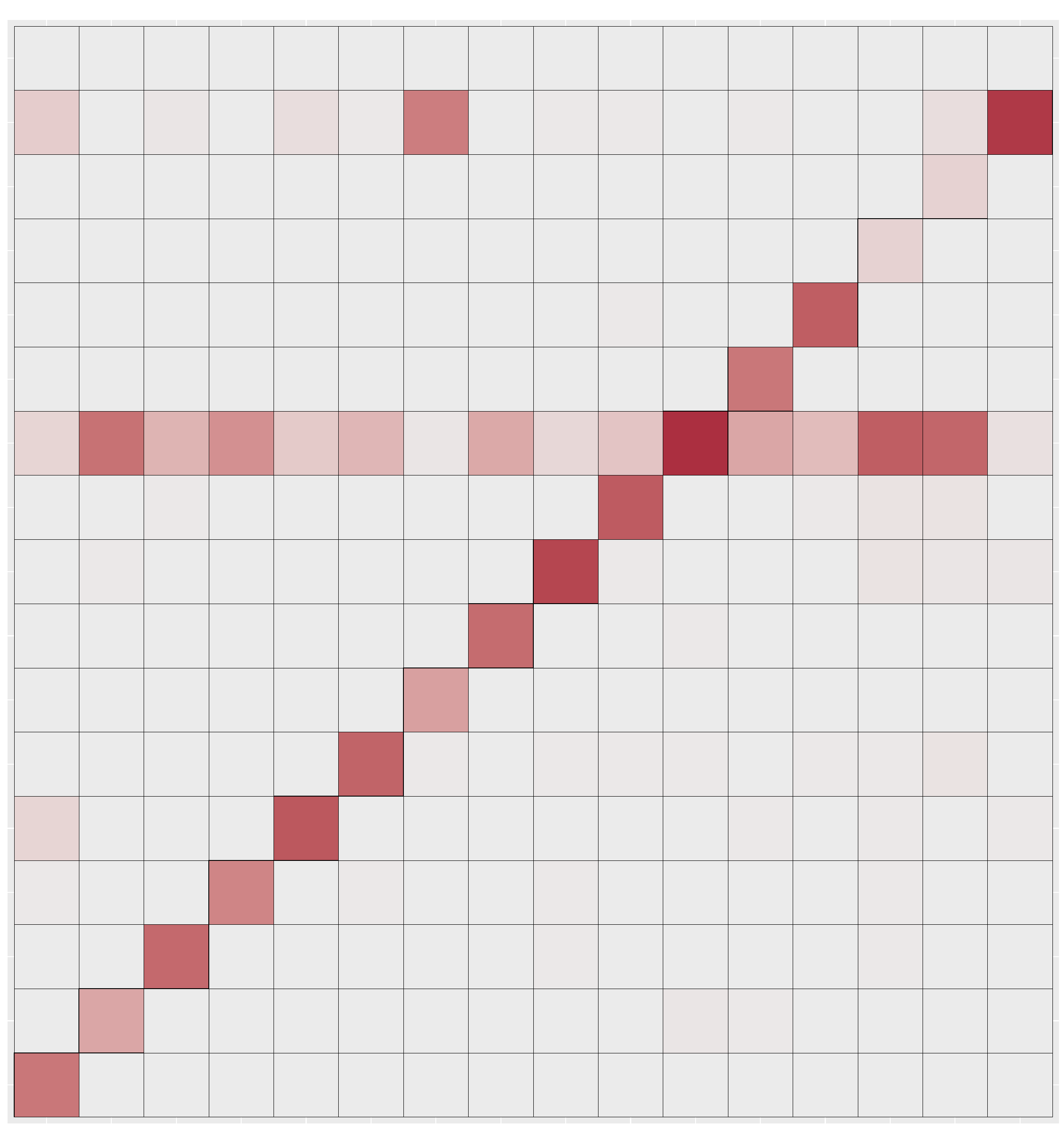}
		\hfill
		\includegraphics[width=1.0\linewidth]{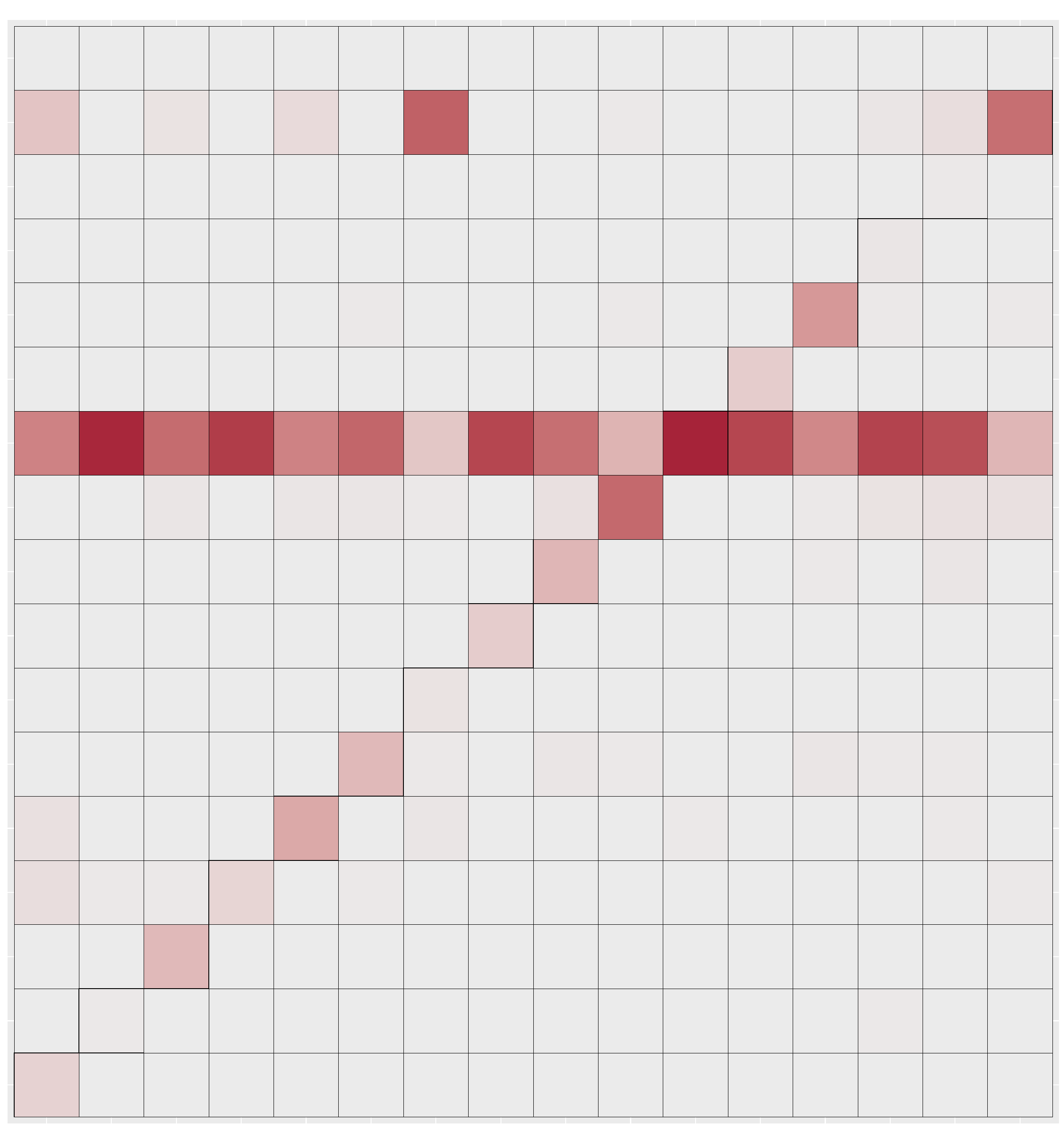}
		\hfill
		\includegraphics[width=1.0\linewidth]{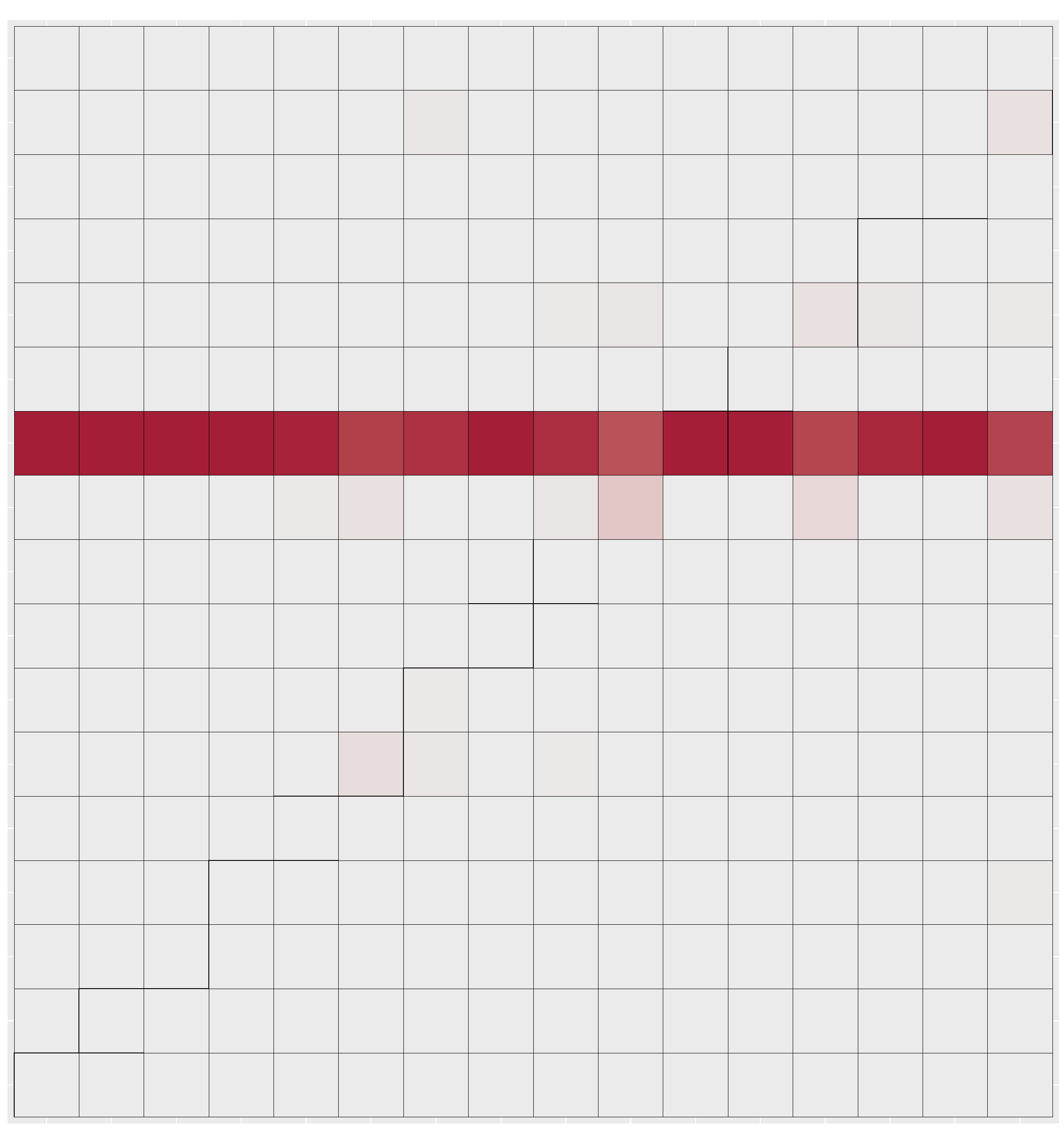}
		\hfill
		\includegraphics[width=1.0\linewidth]{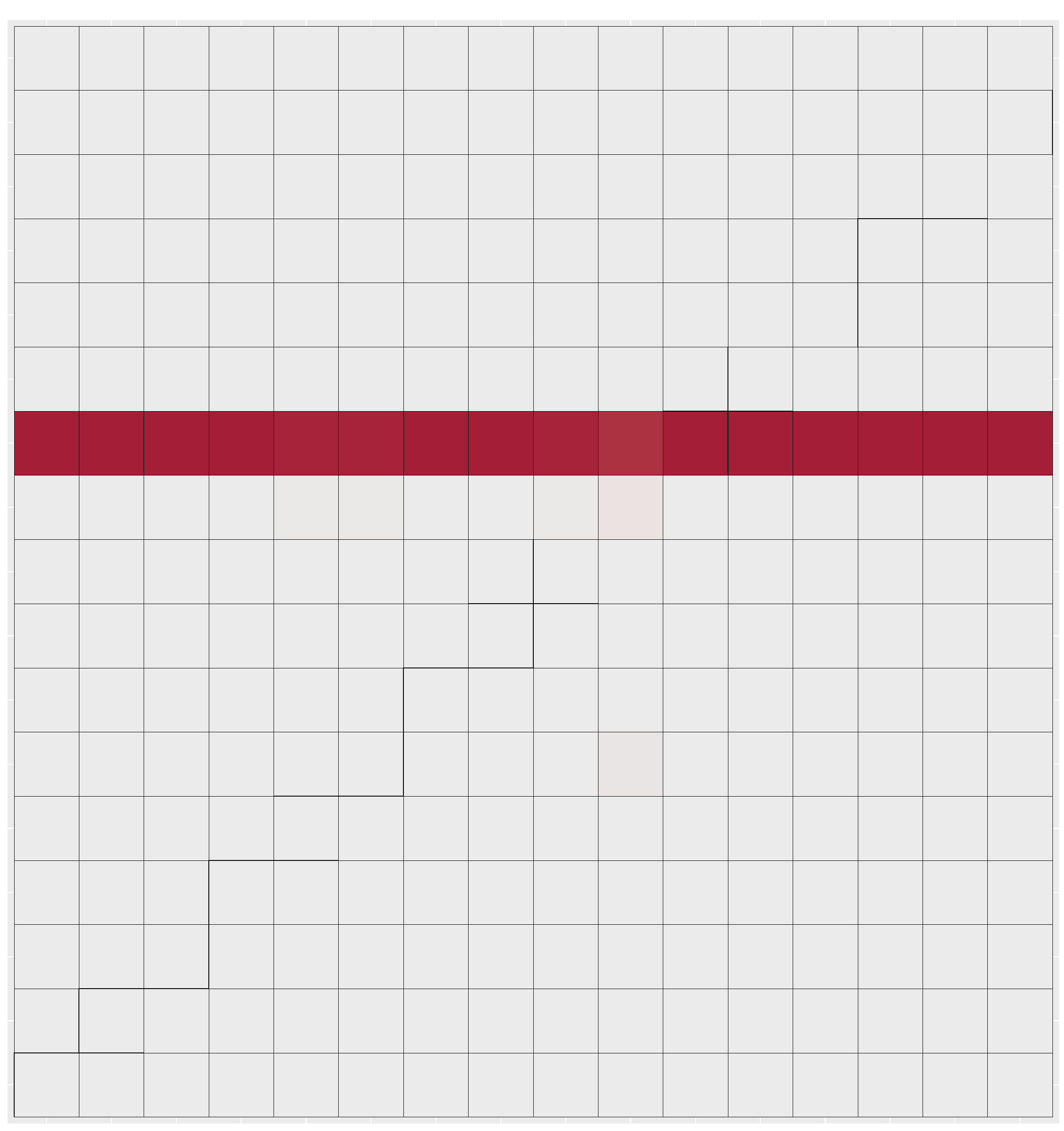}
		\hfill
		\includegraphics[width=1.0\linewidth]{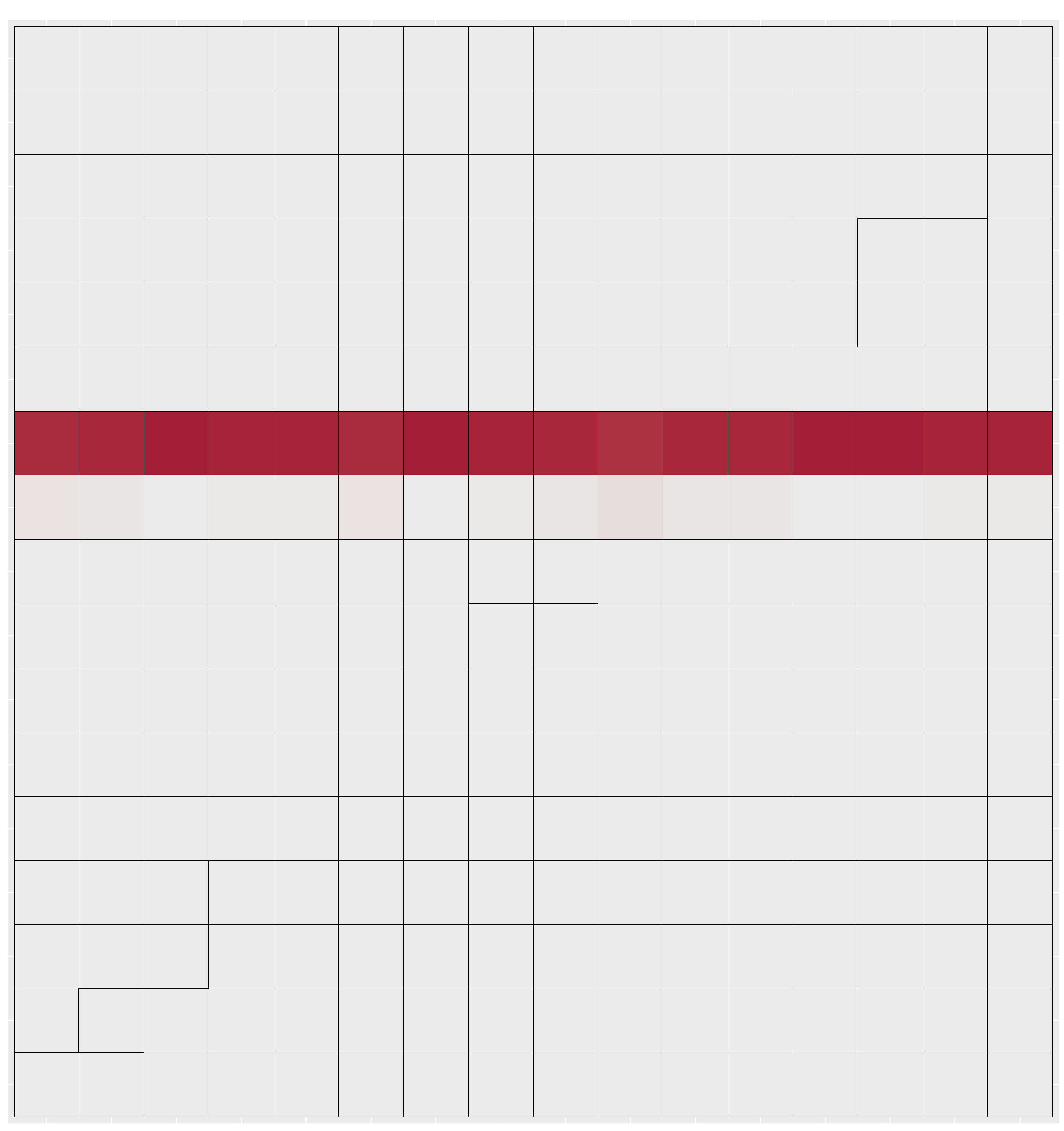}
		\hfill
		\includegraphics[width=1.0\linewidth]{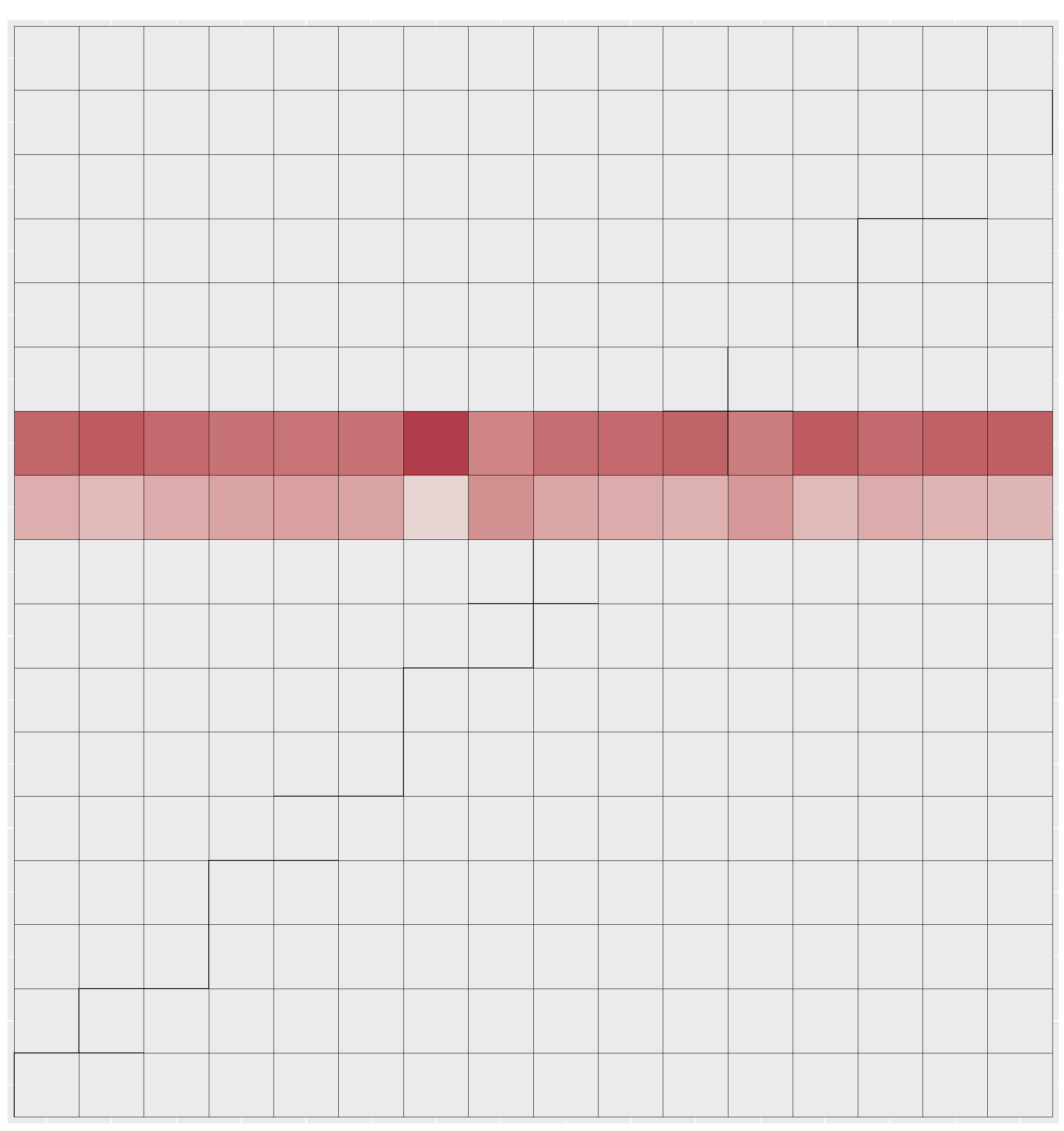}
		\hfill
		\vspace{-16pt}
		\includegraphics[width=0.98\linewidth]{confusion/response_screens_combination/response_icons_horizontal.png}
		\hfill
		\end{center}
		\end{subfigure}
	\hspace*{-5pt}
	\begin{subfigure}{0.16\linewidth}
		\caption{ResNet-152}
		\vspace{-0.35cm}
		\begin{center}		
		\includegraphics[width=0.98\linewidth]{confusion/response_screens_combination/response_icons_horizontal.png}
		\hspace{2pt}
		\hfill
		\vspace{-11pt}
		\includegraphics[width=1.0\linewidth]{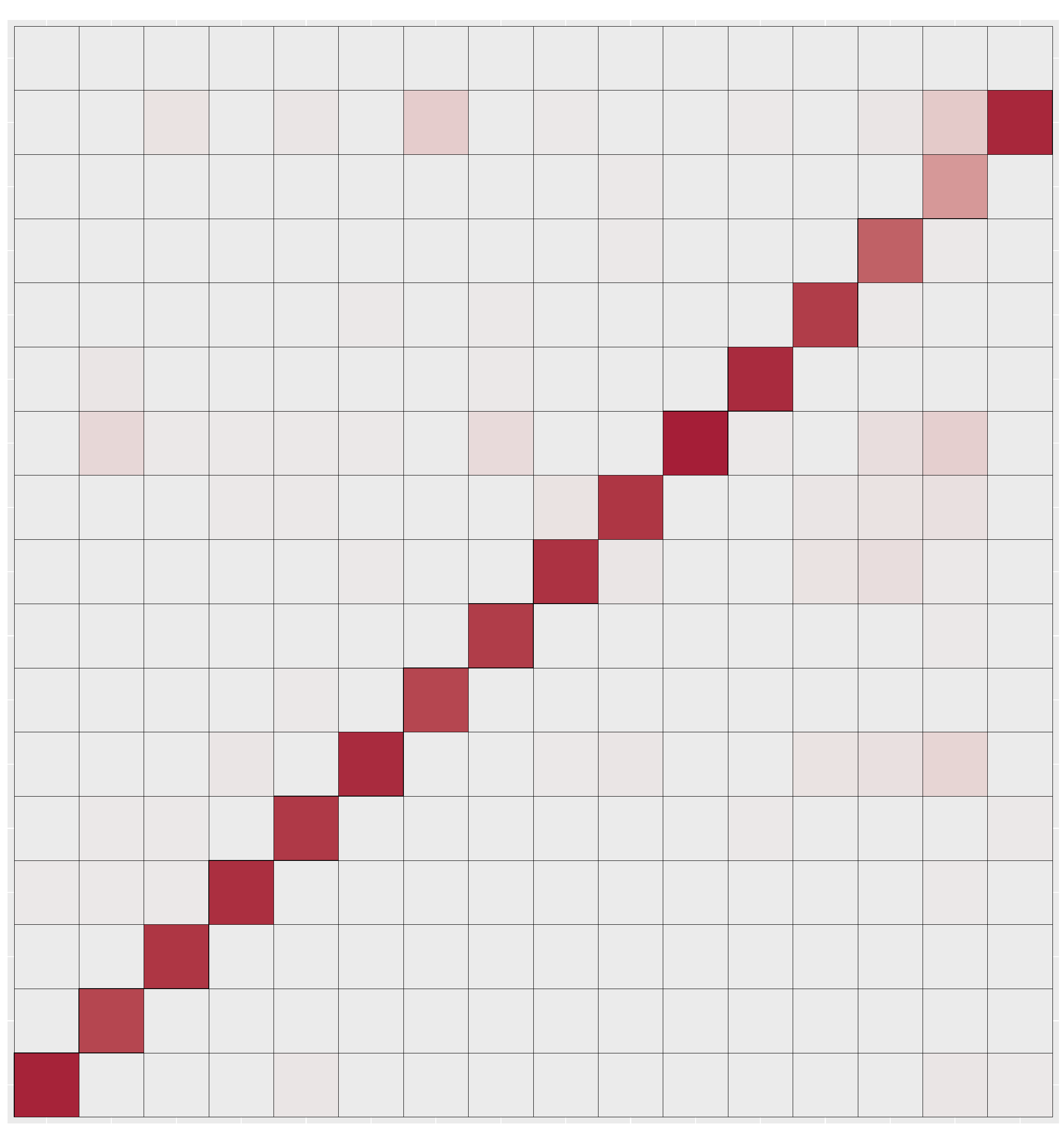}
		\hfill
		\includegraphics[width=1.0\linewidth]{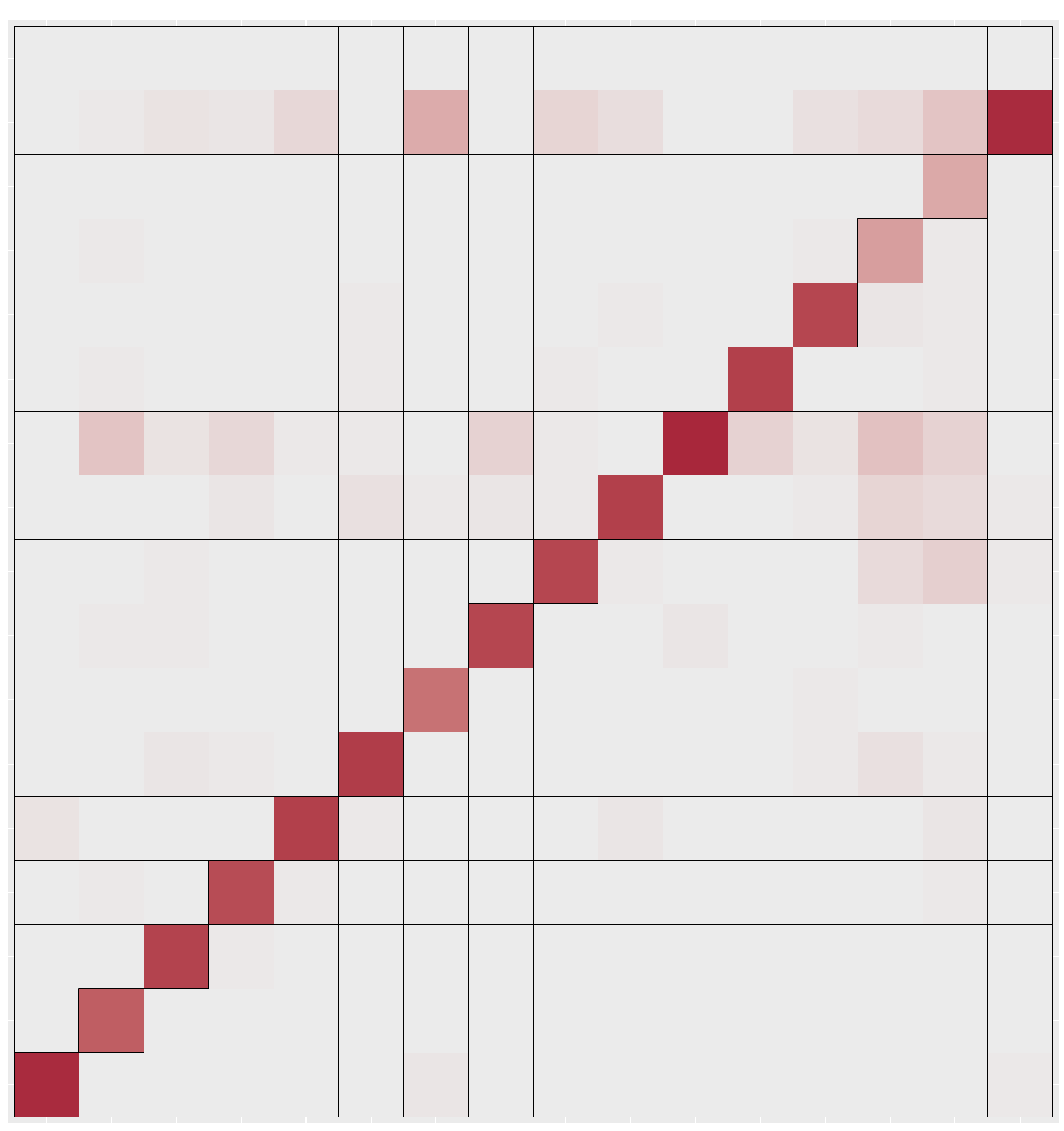}
		\hfill
		\includegraphics[width=1.0\linewidth]{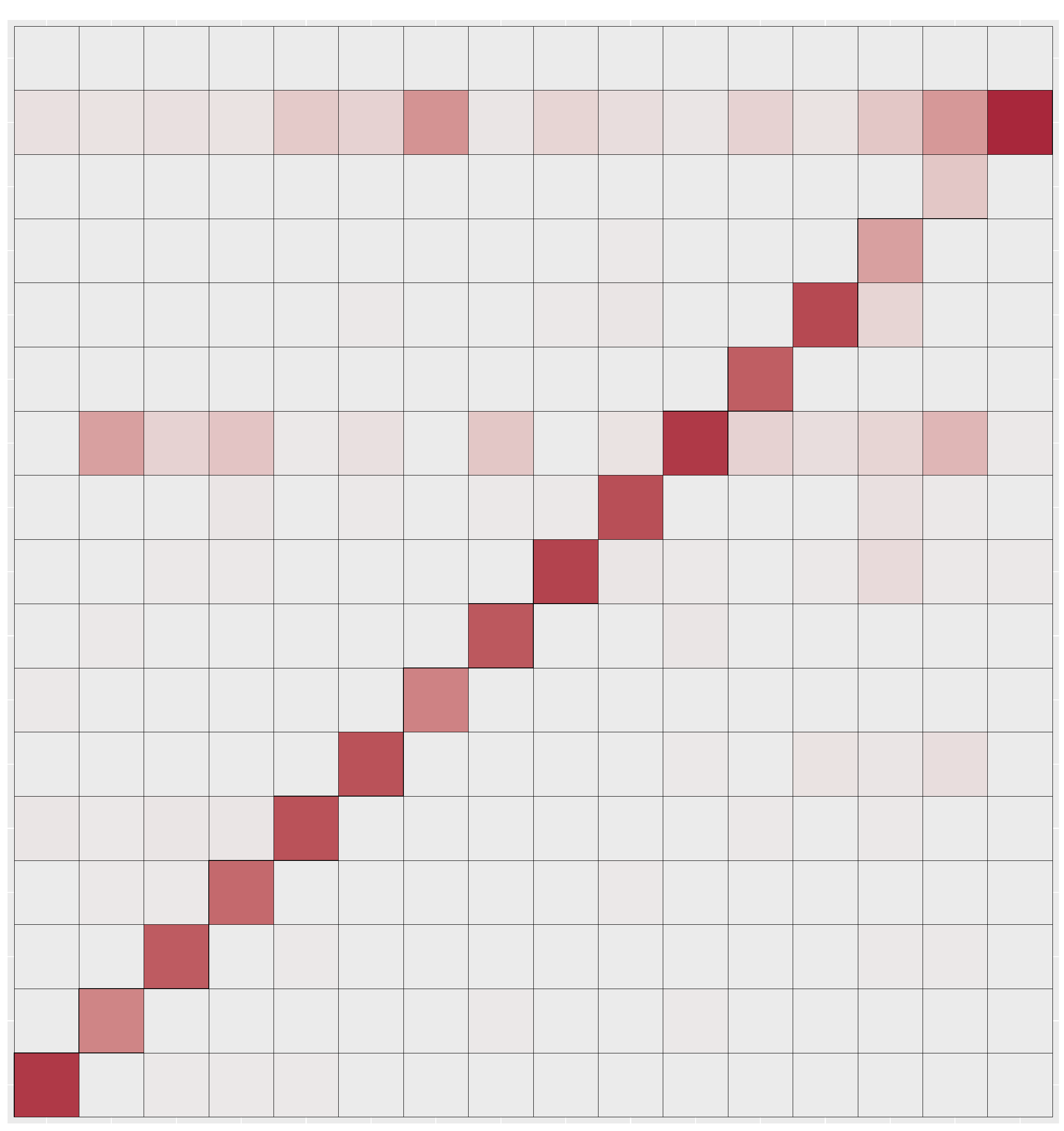}
		\hfill
		\includegraphics[width=1.0\linewidth]{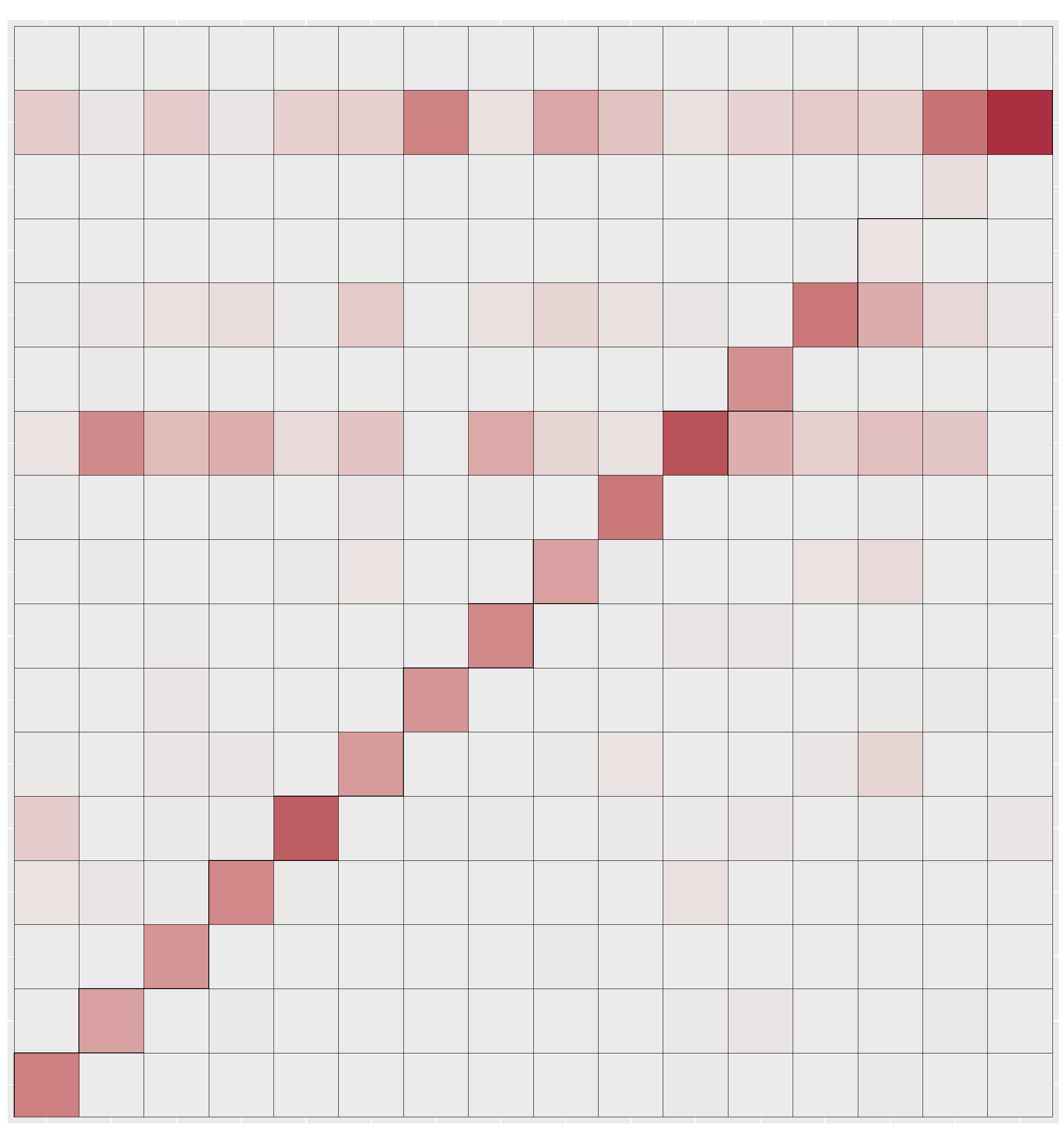}
		\hfill
		\includegraphics[width=1.0\linewidth]{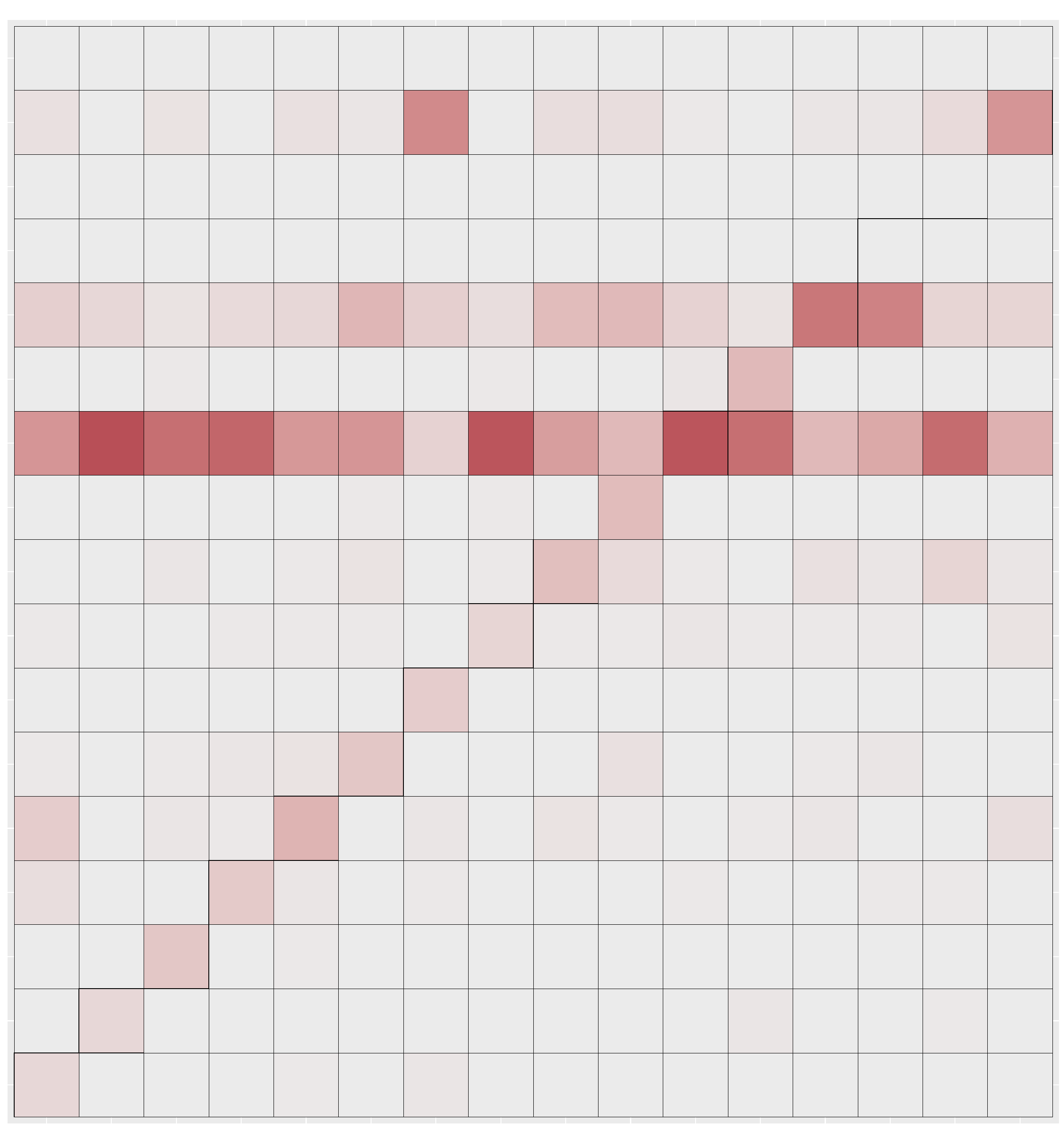}
		\hfill
		\includegraphics[width=1.0\linewidth]{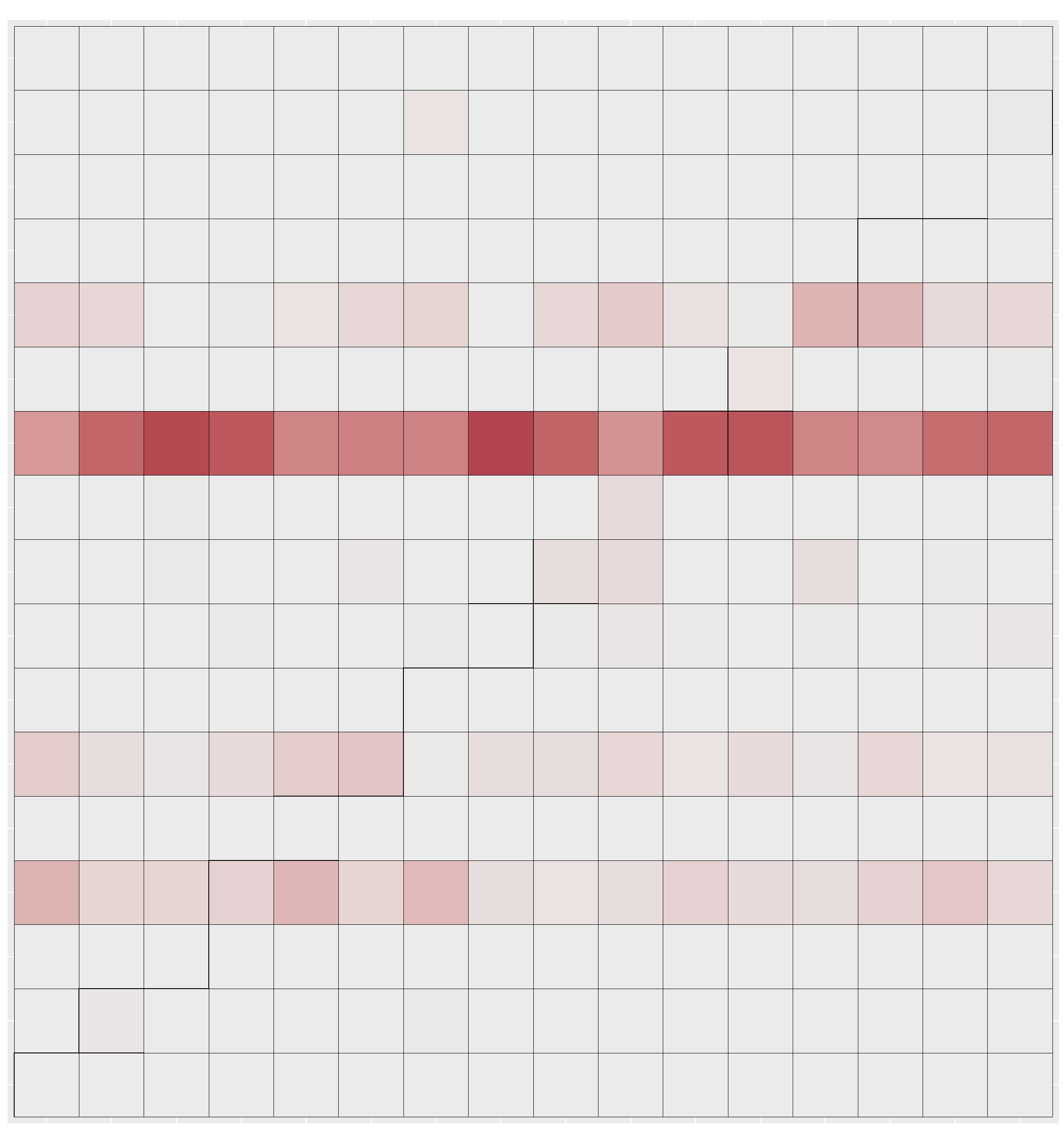}
		\hfill
		\includegraphics[width=1.0\linewidth]{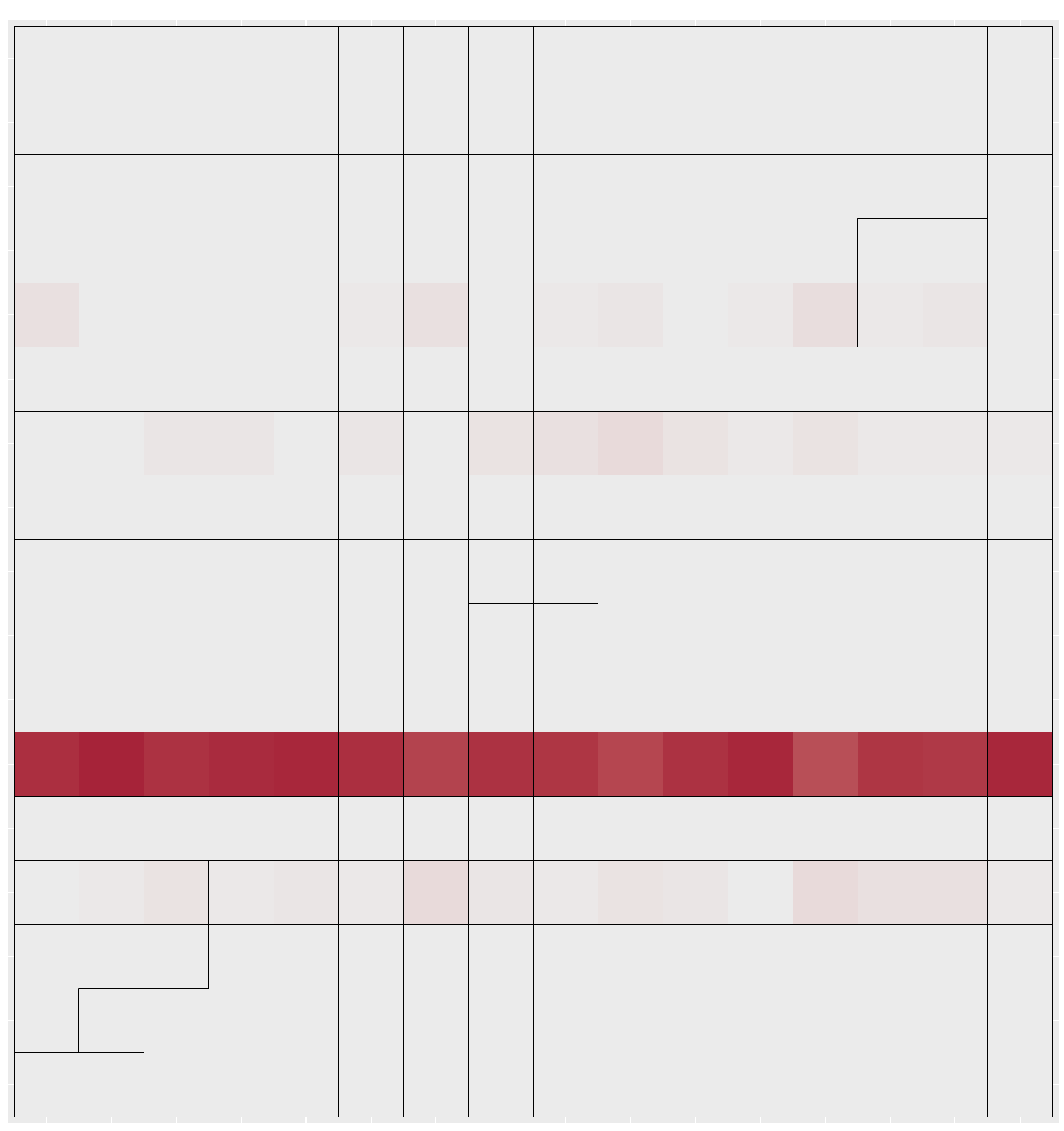}
		\hfill
		\includegraphics[width=1.0\linewidth]{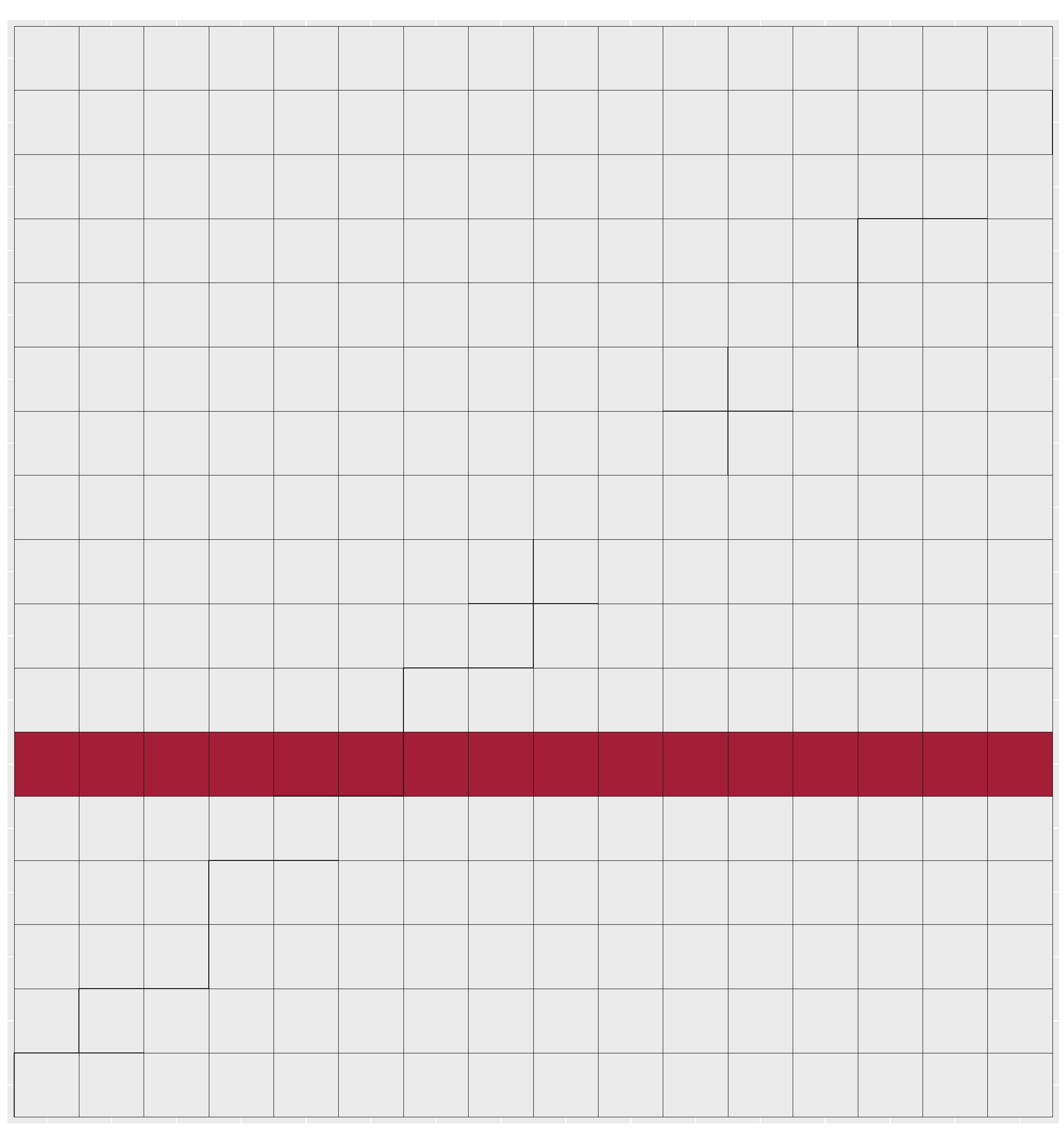}
		\hfill
		\vspace{-16pt}
		\includegraphics[width=0.98\linewidth]{confusion/response_screens_combination/response_icons_horizontal.png}
		\hfill
		\end{center}
	\end{subfigure}
	\hspace*{-2.7pt}
	\begin{subfigure}{0.01\linewidth}
		\vspace{-0.35cm}
		\begin{center}
		\vspace*{24.3pt}
		\includegraphics[height=0.0945\textheight]{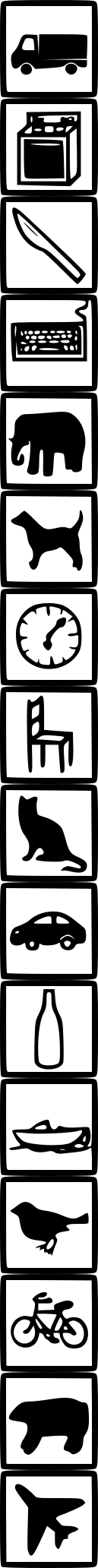}
		\hfill
		\vspace{-4.8pt}
		\includegraphics[height=0.0945\textheight]{confusion/response_screens_combination/response_icons_vertical.png}
		\hfill
		\vspace{-4.9pt}
		\includegraphics[height=0.0945\textheight]{confusion/response_screens_combination/response_icons_vertical.png}
		\hfill
		\vspace{-4.8pt}
		\includegraphics[height=0.0945\textheight]{confusion/response_screens_combination/response_icons_vertical.png}
		\hfill
		\vspace{-4.9pt}
		\includegraphics[height=0.0945\textheight]{confusion/response_screens_combination/response_icons_vertical.png}
		\hfill
		\vspace{-4.8pt}
		\includegraphics[height=0.0945\textheight]{confusion/response_screens_combination/response_icons_vertical.png}
		\hfill
		\vspace{-4.9pt}
		\includegraphics[height=0.0945\textheight]{confusion/response_screens_combination/response_icons_vertical.png}
		\hfill
		\vspace{-4.9pt}
		\includegraphics[height=0.0945\textheight]{confusion/response_screens_combination/response_icons_vertical.png}
		\end{center}
	\end{subfigure}
	\hspace{10pt}
	\begin{subfigure}{0.05\linewidth}
		\begin{center}
		\vspace*{19pt}
		\includegraphics[height=0.4\textheight]{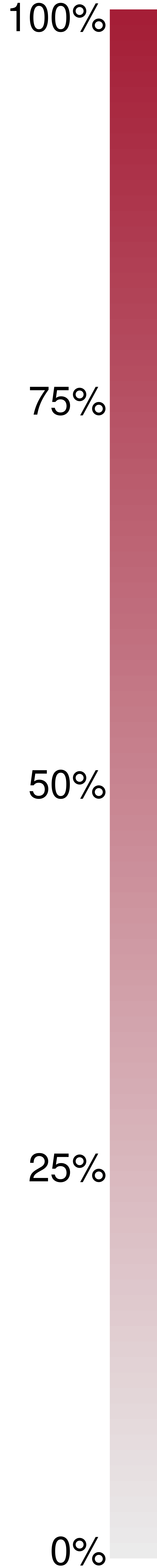}
		\end{center}
	\end{subfigure}

	\caption{Confusion matrices for additive uniform noise. Columns indicate correct category, rows the given classification decision. The top row of each confusion matrix indicates the fraction of failures to respond (i.e. if human observers failed to click on a category).}
	\label{fig:confusion_noise}
\end{figure}

\subsection*{Network training results across all distortion conditions}
\label{sup:network_training_results}
While Figure~\ref{fig:results_training} shows performance of networks trained on distortions for a single distortion level per manipulation, we here report the performance across all stimulus levels. In Figure~\ref{fig:training_accuracy_entropy}, the performance of a vanilla ResNet-50 is compared against a network with the same architecture that is trained on a distortion directly (refered to as `Specialised-Net'), as well as to a network that is trained on all distortions simultaneously (named `All-Distortions-Net'). Object recognition accuracy shows a relatively consistent pattern across experiments: human performance is better than the performance of a vanilla ResNet-50. However, both an All-Distortions-Net and a Specialised-Net reach extremely high accuracies, with the Specialised-Net being either on par with or slightly better than the All-Distortions-Net. Interestingly, the response distribution entropy of those two networks is largely human-like, i.e. close to 4 bits of entropy (or no bias towards a certain category), even for conditions where the overall accuracy is low (e.g. for the difficult conditions of uniform and salt-and-pepper noise).

\section*{Aggregating probabilities from coarse classes}
\label{app:optimal_aggregation}
\subsection*{Motivation}
What is the optimal way to aggregate probabilities from many fine-grained classes to a few coarse classes? Throughout this paper, we used a sensible ad hoc choice for aggregation: summing the probabilities of the fine-grained classes corresponding to a coarse class. All plots in this paper are based on this aggregation method. However, we later discovered that this way of aggregating probabilities is suboptimal and that there indeed exists a principled aggregation choice. Below, we outline how this optimal choice can be derived. We observe that the choice of aggregation method has an influence on experimental results (both accuracy and response distribution entropy). While this influence is certainly not negligible (and we thus recommend that future experiments use the optimal aggregation choice), none of the conclusions from our paper are affected by this choice.

\subsection*{Approach}
Unlike generative models that predict the likelihood of an input given a class and thus can be directly combined with a new prior over classes, deep neural networks for classification are typically trained discriminatively to directly predict the posterior distribution over classes given an input. They assume a fixed prior distribution that is the same at train and test time. Changing this prior distribution over classes at test time without retraining is, however, desirable for many applications.

Here we show that it is actually possible to change the prior distribution over classes at test time even when the network is trained discriminatively to predict the posterior under a different prior. We derive the new posterior distribution as a function of the old posterior distribution given by the network, the old prior distribution given by the training data, and our new, arbitrarily chosen prior distribution. In addition, this allows us to derive the decision of a DNN on a set of coarse classes, each representing a different subset of the fine-grained classes the DNN was trained on. The resulting formula for the DNN's decision on coarse classes differs from the ad hoc choice used in this paper.

Deep neural networks for classification are trained with a fixed prior distribution given by the prior distribution of the training data (and possibly additional class-weighting factors). Given this prior distribution \( p(c) \) over classes \( c \), a network learns to predict the posterior distribution \( p(c | x) \) over classes given an input. We are interested in choosing a new prior distribution \( q(c) \) at test time and obtaining the new posterior distribution \( q(c | x) \). Choosing a new prior at \emph{test time} requires us to work with the posterior distribution \( p(c | x) \) that we already have without training the network on the new prior distribution \( q(c) \).

Note that we are ultimately interested in the network's decision for one of \( k < N \) coarse classes, each of which represents a different set of the \( N \) fine-grained classes the network was trained on. The number of fine-grained classes belonging to a coarse class can differ between coarse classes, and each fine-grained class can belong to zero or one coarse class, i.e., the coarse classes might not cover all fine-grained classes, but no fine-grained class belongs to more than one coarse class. All of this can be represented by a new, non-uniform prior distribution that includes setting the prior of certain fine-grained classes to \( 0 \). Therefore, we will first derive the new posterior distribution for an arbitrary new prior distribution and then use this result to derive a formula for the aggregation of fine-grained probabilities to coarse probabilities.

\subsection*{Derivation of the posterior}

We start by noting that the likelihood \( p(x | c) \) is fixed and does not depend on the prior, i.e.\ \( p(x | c) = q(x | c) \). This is in contrast not only to the prior distribution \( p(c) \ne q(c) \) and the posterior distribution \( p(c | x) \ne q(c | x) \) but also the marginal likelihood \( p(x) \ne q(x) \) and the joint distribution \( p(c, x) \ne q(c, x) \). By applying Bayes' rule twice, once to \( q(c | x) \) and once to \( p(x | c) \), we can derive the new posterior \( q(c | x) \) as a function of the known posterior \( p(c | x) \), the known priors \( p(c) \) and \( q(c) \) and the \emph{unknown} marginal likelihoods \( p(x) \) and \( q(x) \):

\begin{equation}
\begin{split}
	q(c | x) &= \frac{q(x | c)q(c)}{q(x)} \\
	&= p(x | c) \frac{q(c)}{q(x)} \\
	&= \frac{p(c | x)p(x)}{p(c)} \frac{q(c)}{q(x)} \\
	&= p(c | x) \frac{q(c)}{p(c)} \frac{p(x)}{q(x)} \\
	&= p(c | x) \frac{q(c)}{p(c)} \frac{1}{\gamma_x}
\end{split}
\end{equation}

Note that \( p(x) \ne q(x) \), but both are fixed constants given \( x \), thus \( \frac{q(x)}{p(x)} \) can be seen as the normalization constant~\( \gamma_x \) of \( q(c | x) \) that we do not need to calculate to obtain \( q(c | x) \). Instead, the new posterior is simply given by the old posterior multiplied by the ratio of the new prior and the old prior, and renormalized such that \( q(c | x) \) is a probability distribution.

\subsection*{Derivation of the decision}

Let us consider \( k \) coarse classes \( C_1, \dots, C_k \), each of which represents a set of the fine-grained classes \( c_1, \dots, c_N \), i.e.\ \( C_i = \{ c_{i_1}, \dots, c_{i_{N_i}} \} \), with \( N_i = \vert C_i \vert \). Each fine-grained class belongs to at most one coarse class, i.e.\ \( C_i \cap C_j = \emptyset \) for \( i \ne j \) and the sets can have varying sizes, i.e.\ in general \( N_i \ne N_j \) for \( i \ne j \).

This is exactly the setting introduced in our paper: sixteen coarse classes, e.g.\ dogs and cars, each covering between \( 1 \) and \( 109 \) fine-grained ImageNet classes. In total, the sixteen coarse classes make up for \( 207 \) fine-grained classes. The remaining \( 793 \) fine-grained classes are ignored.

The new prior \( q(c) \in [0, 1]^{1000} \) is then defined by

\begin{equation}
	[q(c)]_j =
	\begin{cases}
		\frac{1}{16} \frac{1}{\vert C_i \vert} \text{ if } c_j \in C_i \\
 		0 \text{ if } c_j \notin C_1 \cup \dots \cup C_{16}
	\end{cases}
\end{equation}

Assuming our network was trained on a uniform prior \( p(c) = \frac{1}{1000} \) over the \( 1000 \) ImageNet classes, we can now calculate the new posterior distribution \( q(c | x) \) over fine-grained classes \( c \):

\begin{equation}
	q(c | x) = p(c | x) q(c) \cdot 1000 \cdot \frac{1}{\gamma_x}
\end{equation}

Finally, using this new posterior distribution over fine-grained classes \( c \) we can easily derive the posterior distribution over the coarse classes \( C \):

\begin{equation}
	q(C | x) = \sum_{c \in C} q(c | x) = \frac{1}{\gamma_x} \frac{1000}{16} \sum_{c \in C} \frac{1}{\vert C \vert} p(c | x)
\end{equation}

The probability of a coarse class is thus given, up to normalization, as the \textbf{average} of the probabilities of the corresponding fine-grained classes and the network's decision \( C|x \) is the coarse class with the largest average over fine-grained probabilities:

\begin{equation}
	C|x = \underset{C}{\arg\max}\, q(C | x) = \underset{C}{\arg\max} \sum_{c \in C} \frac{1}{\vert C \vert} p(c | x)
	\label{eq:decision}
\end{equation}

\subsection*{Experiments}
We performed a series of experiments to evaluate the accuracy and entropy of a ResNet-50 model on the 16-class-ImageNet task using our new aggregation method (equation~\ref{eq:decision}). On clean, unperturbed images, we achieve a 0.5\% higher accuracy than before (98.8\% instead of 98.3\%). On noisy images using the same levels and perturbation types as throughout our paper, our method reveals an underestimation of accuracy and entropy of up to 10\% and 1.8 bits (of 4 bits max) in certain conditions of as well as an overestimation of up to 4\% and 1.2 bits in others.

\subsection*{Conclusion}
We showed that it is possible to derive the correct posterior distribution of a discriminatively trained classification network under a new prior chosen at test time and used this to derive the formula for calculating the network's decision on coarse classes. While our derived formula (equation~\ref{eq:decision}) might seem unintuitive and differs from our previous ad hoc choice (summation), our experiments indicate that it does indeed lead to overall higher accuracy and entropy. Nonetheless, none of the conclusions of the paper are affected by this choice. In future experiments, we recommend using the optimal aggregation method (equation~\ref{eq:decision}).

\end{document}